\PassOptionsToPackage{unicode}{hyperref}
\PassOptionsToPackage{hyphens}{url}
\PassOptionsToPackage{dvipsnames,svgnames,x11names}{xcolor}
\documentclass[
  10pt,
  a4paper,
]{article}
\usepackage{amsmath,amssymb}
\usepackage{iftex}
\usepackage[T1]{fontenc}
\usepackage[utf8]{inputenc}
\usepackage{textcomp} % provide euro and other symbols
\usepackage{lmodern}
\IfFileExists{upquote.sty}{\usepackage{upquote}}{}
\IfFileExists{microtype.sty}{% use microtype if available
  \usepackage[]{microtype}
  \UseMicrotypeSet[protrusion]{basicmath} % disable protrusion for tt fonts
}{}
\makeatletter
\@ifundefined{KOMAClassName}{% if non-KOMA class
  \IfFileExists{parskip.sty}{%
    \usepackage{parskip}
  }{% else
    \setlength{\parindent}{0pt}
    \setlength{\parskip}{6pt plus 2pt minus 1pt}}
}{% if KOMA class
  \KOMAoptions{parskip=half}}
\makeatother
\usepackage{xcolor}
\usepackage[margin=1in]{geometry}
\usepackage{longtable,booktabs,array}
\usepackage{calc} % for calculating minipage widths
\usepackage{etoolbox}
\makeatletter
\patchcmd\longtable{\par}{\if@noskipsec\mbox{}\fi\par}{}{}
\makeatother
\IfFileExists{footnotehyper.sty}{\usepackage{footnotehyper}}{\usepackage{footnote}}
\makesavenoteenv{longtable}
\providecommand{\tightlist}{%
  \setlength{\itemsep}{0pt}\setlength{\parskip}{0pt}}
\IfFileExists{newtxtext.sty}{\usepackage{newtxtext,newtxmath}}{\usepackage{mathptmx}}
\DeclareUnicodeCharacter{00B0}{\ensuremath{^\circ}}
\DeclareUnicodeCharacter{00B1}{\ensuremath{\pm}}
\DeclareUnicodeCharacter{00B5}{\ensuremath{\mu}}
\DeclareUnicodeCharacter{00D7}{\ensuremath{\times}}
\DeclareUnicodeCharacter{03BA}{\ensuremath{\kappa}}
\DeclareUnicodeCharacter{03BB}{\ensuremath{\lambda}}
\usepackage{microtype}
\usepackage{graphicx}
\usepackage{booktabs}
\usepackage{longtable}
\usepackage{pdflscape}
\usepackage{array}
\usepackage[numbers,sort&compress,super]{natbib}
\usepackage{caption}
\usepackage{float}
\usepackage{enumitem}
\usepackage{etoolbox}
\AtBeginEnvironment{longtable}{\small}
\setlist{nosep}
\providecommand{\ul}[1]{\underline{#1}}
\ifLuaTeX
  \usepackage{selnolig}  % disable illegal ligatures
\fi
\usepackage{bookmark}
\IfFileExists{xurl.sty}{\usepackage{xurl}}{} % add URL line breaks if available
\hypersetup{
  colorlinks=true,
  linkcolor={blue},
  filecolor={Maroon},
  citecolor={Blue},
  urlcolor={blue},
  pdfcreator={LaTeX via pandoc}}

\author{}
\date{}

\begin{document}

\begin{center}
{\LARGE\bfseries Information-seeking failures of large language models in agentic clinical reasoning\par}
\vspace{0.85em}
{\large Krischan Braitsch\textsuperscript{1}, Laura K. Schmalbrock\textsuperscript{2,3}, Theresa Weltermann\textsuperscript{4}, Andrew F. Berdel\textsuperscript{5}, Isabella Miller\textsuperscript{6}, Kai Tran\textsuperscript{7}, Michael Heider\textsuperscript{8}, Sabrina Kraus\textsuperscript{9}, Florian Bassermann\textsuperscript{1,10-12}, Jacqueline Lammert\textsuperscript{13-15}, Sebastian Ziegelmayer\textsuperscript{16}, Marcus Makowski\textsuperscript{16}, Lisa C Adams\textsuperscript{16}\textsuperscript{*}, Keno K Bressem\textsuperscript{16,17}\textsuperscript{*}\par}
\vspace{0.8em}
\begin{minipage}{0.96\linewidth}
\footnotesize
\setlength{\parskip}{0.2em}
\setlength{\parindent}{0pt}
\textsuperscript{1} Technical University of Munich, School of Medicine and Health, TUM University Hospital, Klinikum rechts der Isar, Department of Medicine III, Munich, Germany\\
\textsuperscript{2} Department of Hematology, Oncology and Cancer Immunology, Charité - Universitätsmedizin Berlin, Corporate Member of Freie Universität Berlin and Humboldt-Universität zu Berlin, Berlin, Germany.\\
\textsuperscript{3} German Cancer Research Center (DKFZ), Heidelberg, Germany\\
\textsuperscript{4} Department of Medicine III, LMU University Hospital, Munich, Germany\\
\textsuperscript{5} Department of Medicine A, Hematology/Oncology, University Hospital Münster, Münster, Germany\\
\textsuperscript{6} Onkologie/Hämatologie im Elisenhof, Munich, Germany\\
\textsuperscript{7} Department of Hematology, Oncology and Palliative Care, Klinikum Traunstein, Traunstein, Germany\\
\textsuperscript{8} Department of Molecular and Cell Biology, University of California at Berkeley, Berkeley, CA, USA\\
\textsuperscript{9} Department of Internal Medicine II, University Hospital Würzburg, Würzburg, Germany\\
\textsuperscript{10} TranslaTUM, Center for Translational Cancer Research, Technical University of Munich, Munich, Germany\\
\textsuperscript{11} Deutsches Konsortium für Translationale Krebsforschung, Heidelberg, Germany\\
\textsuperscript{12} Bavarian Cancer Research Center, Munich, Germany\\
\textsuperscript{13} Department of Gynecology and Center for Hereditary Breast and Ovarian Cancer, Technical University of Munich (TUM), School of Medicine and Health, Klinikum rechts der Isar, TUM University Hospital, Munich, Germany\\
\textsuperscript{14} Institute of Artificial Intelligence and Informatics in Medicine (AIIM), TUM University Hospital, Technical University of Munich (TUM), Munich 81675, Germany\\
\textsuperscript{15} German Cancer Consortium (DKTK), partner site Munich, a partnership between DKFZ and TUM University Hospital, Munich, Germany\\
\textsuperscript{16} Technical University of Munich, School of Medicine and Health, TUM University Hospital, Klinikum rechts der Isar, Department of Diagnostic and Interventional Radiology, Germany\\
\textsuperscript{17} Technical University of Munich, School of Medicine and Health, TUM University Hospital, German Heart Center, Department of Cardiovascular Radiology and Nuclear Medicine, Germany
\vspace{0.45em}
\textbf{Corresponding author:} Keno K. Bressem, MD. \texttt{Keno.bressem@tum.de}
\end{minipage}
\end{center}

\begin{abstract}

Large language models achieve high scores on medical knowledge assessments, yet clinical reasoning requires actively deciding what to investigate under uncertainty. We developed an agentic evaluation framework in hematologic oncology in which models must proactively request clinical data across three sequential rounds before committing to a diagnosis and treatment plan. Across 32 frontier models, the best achieved only 68\% overall accuracy. Information utilization - the fraction of available data actually requested - was the strongest predictor of diagnostic accuracy (R = 0.69, P \textless{} 0.001), yet utilization collapsed from 57\% to 26\% in the final round, leaving molecular and cytogenetic data critical for treatment selection unexamined. Reasoning traces scored high on a clinical reasoning rubric (91\% above threshold) but decorrelated from accuracy, revealing a gap between locally coherent rationales and globally correct conclusions. Error analysis identified search satisficing, anchoring and premature closure as the dominant failure modes - the same cognitive biases that characterize novice clinicians under dual-process models of diagnostic reasoning. These findings demonstrate that the primary limitation of current models in clinical oncology is not insufficient medical knowledge but a systematic failure of information-seeking under uncertainty.

\end{abstract}

\section{Introduction}\label{introduction}

Clinical decision-making in hematologic oncology requires the sequential
integration of incomplete and evolving information. Resource constraints
and clinical urgency demand a stepwise, case-adapted approach in which
each diagnostic step is selected based on its expected impact on the
therapeutic decision at hand. Molecular profiling results arrive
incrementally, treatment decisions must account for prior therapies and
transplant history, and concurrent complications may fundamentally alter
the therapeutic calculus. At each stage, clinicians must decide how to
interpret partial results and when to act under uncertainty. This
capacity to identify what information is needed and to prioritize
appropriately constitutes a core competence of an experienced
hematologist. In cognitive science, this process is formalized in
dual-process models of diagnostic reasoning, where the transition from
rapid pattern recognition to deliberate hypothesis testing depends
critically on recognizing when initial impressions are
insufficient.\cite{Wasterlid2022-el}

Large language models (LLMs) have demonstrated strong performance on
medical knowledge assessments and oncology decision
support.\cite{Singhal2023-bm,Nori2023-aj,Gilson2023-jp} Yet reliable
active information-seeking by these systems remains insufficiently
tested in subspecialty
oncology.\cite{Ferber2024-vz,Truhn2023-br,Thirunavukarasu2023-gr}
Multimodal AI systems that integrate information from radiology,
histopathology, genomics, and clinical
records\cite{Acosta2022-yr,Lipkova2022-ol,Moor2023-rh,Tu2024-pl}
and autonomous AI agents have demonstrated promising results in
structured clinical tasks. However, all existing evaluations share a
critical limitation: they present models with complete clinical
information and assess the ability to interpret it, not the ability to
decide what information to
seek.\cite{Ferber2025-bs,Lee2024-mc}

Existing benchmarks present clinical information
passively.\cite{Bedi2026-he,Tordjman2025-hw,Singhal2025-fw}
In clinical practice, the direction of information flow is reversed:
clinicians must actively decide what to investigate, interpret results
as they arrive, and revise plans accordingly. This process of sequential
planning under uncertainty separates clinical reasoning from medical
knowledge retrieval. In human cognition, novice clinicians
characteristically fail at exactly this juncture, exhibiting search
satisficing, premature closure, and anchoring bias as described in
Croskerry's dual-process model. Whether large language models exhibit
analogous information-seeking failures has not been systematically
tested.\cite{Valmeekam2023-ti,Cao2025-fj}

Here we test this directly using OncoRounds, an agentic evaluation
framework that requires proactive information-seeking in hematologic
oncology. Models are presented with an initial clinical scenario and
must actively request clinical information, one item per turn, across
three rounds of increasing complexity. This design separates the ability
to identify relevant clinical data from the ability to interpret it, and
evaluates both within a single, integrated framework. Across 32 frontier
LLMs and 20 complex cases,the highest-performing model achieved only
68\% overall accuracy, diagnostic performance deteriorated as cases
progressed from syndromic impressions to molecular diagnoses, and models
converged on the same information-seeking failure modes that
characterize novice human clinicians, despite sharing no developmental
pathway with them.

\section{Results}\label{results}

We evaluated 32 frontier LLMs (Supplementary Table 1) across 20 complex
hematologic malignancy cases (Table 1, Figure 1c/d), each structured
across three sequential rounds that mirror the patient journey from
initial presentation through the first week on the ward to long-term
definitive treatment planning, with diagnostic results staged to reflect
realistic clinical timelines. Of the 32 models, 15 had open weights, and
23 provided at least partial access to reasoning traces (Figure 1c/d).
In each round, models were presented with a clinical scenario and
allowed to request additional clinical information one item per turn
before being assessed on diagnostic accuracy, differential reasoning,
and treatment recommendations. Given the scale of evaluation, we
employed an LLM-as-judge approach,\cite{Gu2025-iw} validating its
accuracy against an expert hematologist with 10 years of clinical
experience.

\subsection{Overall Performance}\label{overall-performance}

The highest-scoring model, Claude Opus 4.6, reached an overall accuracy
of 68.1\%, while the lowest-performing model, Llama 3.3 70B, scored
40.2\% (mean across all models 56.3 ± 7.4\%; Table 2). Performance
varied substantially across scoring components. Diagnostic accuracy was
highest (mean 66.6 ± 9.2\%), followed by treatment recommendations (58.8
± 6.5\%), while differential reasoning proved most challenging (43.6 ±
9.4\%). All 32 evaluated models showed differential reasoning as their
weakest component, regardless of overall performance level. The gap
between diagnostic accuracy and differential reasoning averaged 23.0
percentage points, suggesting that models can identify a likely
diagnosis from available information but struggle to systematically
consider and weigh competing alternatives. More recent models performed
better (R = 0.51, p = 0.002; Figure 1a), consistent with ongoing
capability gains across model generations. Model size showed a modest
positive correlation with overall performance among models with publicly
available parameter counts (R = 0.45, p = 0.046; Figure 1b), though this
analysis was limited to 15 of 32 models, where the number of parameters
was known.

The volume of information requests declined across rounds, from a mean
of 7.5 per round (SD 3.9) in R1 to 4.2 (SD 2.2) in R2 and 1.8 (SD 1.0)
in R3 (Kruskal-Wallis p \textless{} 0.001; Supplementary Figure 1a). Of
these requests, 58.8\% yielded released information, 17.2\% were
invalid, 8.8\% received guidance redirecting to a similar item, and
15.2\% targeted items not yet available in the current round. The
utilization rate (fraction of available clinical information actually
obtained) showed a strong positive correlation with overall accuracy (R
= 0.69, p \textless{} 0.001, Supplementary Figure 1b), indicating that
models gathering more of the available information performed better.
However, utilization declined sharply in later rounds (Supplementary
Figure 1c). In R1, models obtained 56.8\% of available diagnostics,
remaining similar in R2 (56.7\%) before dropping to 25.7\% in R3, with
several models requesting only 1--2 items in R3 before attempting to
solve, leaving molecular panels and treatment response data unexamined.

\subsection{Round Dynamics}\label{round-dynamics}

Model performance declined across rounds (Table 3; Figure 2a--c). Mean
composite scores dropped from 58.0\% in Round 1 (outpatient
presentation) to 55.1\% in Round 2 (inpatient workup) and 51.2\% in
Round 3 (definitive characterization), and 25 of 32 models scored lower
in R3 than R1. Correct diagnosis rates fell from 48.7\% in R1 to 41.6\%
in R3, while partially correct diagnoses rose from 34.3\% to 46.8\%.
Fully incorrect diagnoses decreased from 17.0\% to 11.6\%. This pattern
reflects the increasing specificity required in later rounds, as R3
demand the integrative synthesis of molecular profiles, treatment
history, and concurrent comorbidities to arrive at actionable
therapeutic recommendations. Models often achieved partial diagnostic
precision as additional clinical data became available but failed to
reach the specificity required for full credit.

Not all models finished all cases, being cancelled by the benchmark
after too many wrong requests. Analysis of 630 model-case pairs with
complete R1 and R3 diagnoses revealed pronounced diagnostic anchoring
(Figure 2d,e). Of 323 pairs with an incorrect or partially correct R1
diagnosis, 212 (65.6\%) remained incorrect through R3, yielding a
recovery rate of 34.4\%. Conversely, of 307 pairs with a correct R1
diagnosis, 156 (50.8\%) deteriorated to incorrect or partially correct
by R3.

The asymmetry of incorrect diagnoses persisting at nearly twice the rate
at which correct diagnoses were maintained, suggests that early errors
are far harder to recover from. However, recovery and deterioration
rates varied substantially across models (Supplementary Table 2;
Supplementary Figure 2c). Claude Opus 4.5 corrected all six of its
initially incorrect diagnoses by R3 (100\% recovery), and Claude Opus
4.6 recovered seven of eight (87.5\%). Qwen3 Next 80b recovered only one
of 14 (7.1\%). Deterioration rates ranged from 18.2\% for GPT-5.2
Reasoning (two of 11) to 83.3\% for Qwen3 Next 80b (five of six). Models
with the highest recovery rates also had fewer initial errors.

Anchoring patterns also differed by case (Supplementary Table 5,
Supplementary Figure 5). Case 20 (Richter transformation of CLL) showed
zero deterioration among 30 models with a correct R1 diagnosis,
consistent with a case where the core disease identity is apparent at
presentation and later rounds add prognostic detail without changing the
diagnostic category. By contrast, Case 3 (relapsed AML with complex
karyotype) had 26 of 29 initially correct diagnoses deteriorate by R3
(89.7\%), followed by Case 1 (relapsed AML with FLT3-ITD; 18/19,
94.7\%). Examination of the actual diagnosis text revealed that
deterioration predominantly reflected subtype refinement failure rather
than diagnostic category shifts. Models maintained the correct broad
disease identity (e.g., "AML," "multiple myeloma") but failed to
integrate molecular subtyping, WHO 2022 classification, or
treatment-refractory status when these became available in R3
(Supplementary Table 6). In Case 3, for example, models correctly
identified myelodysplastic syndrome in R1 and appropriately updated to
AML in R3, but achieved only "partially correct" scores because they
could not specify the complete WHO/ELN classification (AML,
myelodysplasia-related (AML-MR per WHO 2022), with complex karyotype,
ELN 2022 adverse risk). Notably, most deterioration resulted in
partially correct rather than fully incorrect R3 diagnoses
(Supplementary Figure 5b), indicating that models incorporated some
molecular findings but could not achieve the precision required for
actionable clinical classification.

\subsection{Clinical Reasoning
Quality}\label{clinical-reasoning-quality}

To assess reasoning quality beyond diagnostic accuracy, an adapted
R-IDEA\cite{Schaye2022-zc} rubric was applied to 1,008 evaluations
across 17 models that produced full thinking traces (Figure 3,
Supplementary Figure 4). Models providing only reasoning summaries
(GPT-5 series, Gemini series) were excluded. R-IDEA scores reasoning
across four dimensions: interpretive summary (I; 0--4), differential
diagnosis (D; 0--2), explanation (E; 0--2), and alternatives considered
(A; 0--2), yielding a total score of 0--10.\cite{Schaye2022-zc}

The mean R-IDEA total was 8.22 (SD 1.59), and 91.1\% of evaluations
scored at or above 6, a threshold previously proposed for high-quality
clinical reasoning.\cite{Schaye2022-zc} Model-level means ranged from
6.45 (GPT-OSS 20B) to 9.58 (Claude Opus 4.5). Claude Opus 4.5, GLM 5
(9.40), and Claude Sonnet 4.6 (9.35) occupied the top three positions,
while GPT-OSS 20B (6.45), Qwen3 235B A22B (6.70), and DeepSeek V3.2
(7.38) formed the lower tier (Table 4). R-IDEA scores slightly increased
in R3 compared to R1 for most models, paired with longer traces but
fewer requests (Supplementary Figure 3).

Dimension-level analysis revealed an asymmetry in reasoning quality.
When scores were normalized to the proportion of each
dimension\textquotesingle s maximum, explanation was strongest (E:
90.8\%), followed by differential diagnosis (D: 89.0\%), while
alternatives (A: 77.5\%) and interpretive summary (I: 76.9\%) showed the
largest deficits. The interpretive summary gap suggests that models
underutilize the interactive structure of the benchmark, requesting
fewer diagnostic tests than would be optimal, also reflected in the
declining number of information requests across rounds (often only 1--2
in Round 3 before the next solve).

R-IDEA scores showed a positive correlation with overall benchmark
accuracy (R = 0.56, p = 0.02), suggesting that models producing
higher-quality reasoning traces tended to achieve better diagnostic
outcomes. However, outliers such as GPT-OSS 120B demonstrated
below-median accuracy (56.4\%) despite high reasoning quality (R-IDEA
8.90).

\subsection{Error Taxonomy}\label{error-taxonomy}

Alongside R-IDEA scoring, reasoning traces were annotated for cognitive
biases and process errors using a taxonomy adapted from
Croskerry\textquotesingle s dual-process model of clinical
reasoning.\cite{Croskerry2009-pe} Errors were identified in 89 of
1,008 evaluations (8.8\%), yielding 93 individual error annotations
across the 17 models with full thinking traces (Table 4).

Process errors were more frequent than cognitive biases. The most common
error types were search satisficing, premature closure, and anchoring
bias. Anchoring bias, the most frequent cognitive bias, appeared across
multiple models, consistent with the diagnostic anchoring observed in
the round-level trajectory analysis. Hallucinated information was rare,
suggesting that models generally constrained their reasoning to the
presented clinical data (Figure 3b).

Error rates varied widely across models (Table 4). Claude Sonnet 4.5,
Claude Opus 4.5, and GLM 5 had the lowest error rates (1.7\% each),
while Qwen3 Next 80B had the highest (26.7\%). The majority of errors
were of medium severity (56/93, 60.2\%), followed by low (27/93, 29.0\%)
and high severity (10/93, 10.8\%). No critical errors with the potential
to cause direct patient harm were identified (Supplementary Figure 4).
Detailed examples on reasoning errors are available in the supplementary
materials.

\section{Discussion}\label{discussion}

Across 20 complex hematologic malignancy cases and 32 frontier LLMs, the
best-performing model achieved 68.1\% overall accuracy when required to
actively gather clinical information across sequential rounds.
Performance was highest for diagnostic accuracy (mean 66.6\%), lower for
treatment recommendations (58.8\%), and lowest for differential
reasoning (43.6\%), a ranking that held across all models regardless of
family, size, or generation.These results indicate that the primary
challenge for current LLMs in clinical oncology lies not in medical
knowledge retrieval but in the sequential processes of information
gathering, prioritization, and diagnostic updating that characterize
real clinical reasoning. Strikingly, the failure modes observed - search
satisficing, anchoring at an initial level of diagnostic resolution, and
premature closure - mirror those documented in novice human clinicians
under dual-process models of diagnostic reasoning. This convergence
suggests that the information-seeking deficit reflects a structural
property of how these systems process sequential, incomplete information
rather than a domain-specific knowledge gap.

LLMs now achieve above 90\% accuracy on USMLE-style
questions,\cite{Singhal2023-bm,Nori2023-aj,Gilson2023-jp,Tordjman2025-hw}
and multi-task evaluation frameworks such as MedHELM span dozens of
clinical domains.\cite{Bedi2026-he} However, these benchmarks
present complete clinical vignettes and ask models to select or generate
an answer. AgentClinic demonstrated that models excelling on MedQA
perform poorly when required to gather information through multi-turn
dialogue,\cite{Schmidgall2024-ht} and MedDialogRubrics,
evaluating over 5,000 synthetic cases with fine-grained rubrics,
concluded that improvements require advances in dialogue management
architectures rather than incremental model
tuning.\cite{Gong2026-yh} MedAgentSim introduced
self-improvement mechanisms for doctor agents and found partial but
incomplete gains.\cite{Almansoori2025-uf} These studies have
focused on general internal medicine. OncoRounds extends interactive
evaluation into subspecialty hematology/oncology, using 20 cases
deliberately constructed to represent the most diagnostically demanding
scenarios in hematologic practice, including relapsed, refractory, and
post-transplant presentations that require the integration of molecular
diagnostics, treatment history, and evolving clinical complications into
coherent management strategies. A controlled, turn-based design
separates information-seeking from interpretation and tracks both across
three clinically staged rounds.

Information utilization, the fraction of available clinical data that a
model actually requested, was the strongest single predictor of overall
accuracy (R = 0.69, p \textless{} 0.001). Yet models systematically
under-requested information, particularly in Round 3, where utilization
dropped from approximately 57\% in Rounds 1 and 2 to 25.7\%. Several
models requested only one or two items before committing to a diagnosis,
leaving molecular panels, cytogenetic results, and treatment response
data unexamined. R-IDEA evaluation of reasoning traces confirmed this
asymmetry. Interpretive summary and alternatives considered were the
weakest dimensions across 17 models, while explanation scored near
ceiling. In human clinical reasoning, search satisficing, the premature
termination of information search after identifying a plausible
explanation, is a well-characterized failure
mode.\cite{Croskerry2009-pe} The present data suggest that LLMs
exhibit an analogous pattern, likely compounded by two known
architectural and behavioral limitations. First, LLMs exhibit positional
attention biases that degrade retrieval of information positioned in the
middle of long contexts,\cite{Liu2023-sm} and performance on complex
reasoning tasks deteriorates as context length increases, with models
producing shorter and less reliable outputs.\cite{Gupta2024-dz}
OncoRounds is demanding in this regard. By the final round, the
accumulated interaction history of requests, released information items,
prior diagnostic assessments, and reasoning traces can exceed 10,000
tokens per case, placing critical molecular data at exactly the context
positions where retrieval is weakest. Second, Laban et al.~demonstrated
that LLMs perform significantly worse in multi-turn conversations than
in equivalent single-turn instructions, with accuracy declining as
interaction depth increases, compounded by premature assumptions from
which models fail to recover.\cite{Laban2025-ns} The
round-over-round performance decline observed here is consistent with
both phenomena, and the interaction between context saturation and
multi-turn degradation may explain why the largest utilization drop
occurs in the final round, where context is longest and the number of
prior turns is greatest. Training predominantly on complete-case
presentations, where active information gathering is never modeled, may
further amplify this tendency.

Composite performance declined across rounds, with 25 of 32 models
scoring lower in the final round than the first. This pattern does not
indicate that additional data harms performance. Rather, later rounds
demand progressively greater diagnostic precision as the reference
standard shifts from syndromic identification to molecular subtyping per
WHO classification, risk stratification, and treatment-refractory
designations. Models consistently identified the correct disease
category but failed to achieve the specificity required for actionable
clinical classification, a pattern the trajectory analysis characterized
as subtype refinement failure. Anchoring bias and premature closure,
among the most frequently cited cognitive errors in clinical
medicine,\cite{Croskerry2009-pe} offer a partial mechanistic
explanation. Recent work confirms that LLMs exhibit analogous
vulnerabilities. Moëll et al.~identified anchoring and difficulty
integrating conflicting data as the most frequent errors in DeepSeek
R1\textquotesingle s medical reasoning,\cite{Moell2025-wa} and
Schmidgall et al.~showed that cognitive biases reduce diagnostic
accuracy across medical LLMs with varying resilience by model
family.\cite{Schmidgall2024-ra} Ke et al.~demonstrated that multi-agent
frameworks with designated devil\textquotesingle s advocate agents can
partially mitigate such biases, improving accuracy from 0\% to 76\% in
diagnostically challenging scenarios.\cite{Ke2024-ea} The OncoRounds
data extend these observations by showing that anchoring operates at a
finer granularity than previously documented. Models did not anchor on
incorrect disease categories but on an initial level of diagnostic
resolution and failed to update toward the molecular precision that
later-round evidence demanded. However, in hematologic oncology, where
treatment selection depends directly on molecular subtype and risk
classification, this level of detail is needed.

Differential reasoning was the weakest scoring component for all 32
models, averaging 23 percentage points below diagnostic accuracy. This
pattern was corroborated by the R-IDEA evaluation, where the
alternatives dimension was among the lowest-scoring, and by the error
taxonomy, where search satisficing and premature closure were the most
frequent error types. However, differential diagnosis is also the most
subjective component in the benchmark, as the clinical relevance and
ranking of alternative diagnoses are less standardized than the primary
diagnosis or treatment plan, making automated scoring less reliable for
this category. With that caveat, the consistent deficit across all
models and evaluation modalities suggests a genuine limitation rather
than a scoring artifact. In hematologic oncology, the differential
diagnosis determines which molecular and cytogenetic studies to pursue.
A model that commits to AML without adequately considering ALL, or
mixed-phenotype acute leukemia will fail to request the investigations
needed to distinguish between them, compounding the
information-gathering deficit observed above.

The positive correlation between R-IDEA scores and benchmark accuracy (R
= 0.56, p = 0.02) indicates that reasoning quality contributes to
diagnostic outcomes, but the relationship is moderate. R-IDEA scores
exceeded the high-quality threshold (6 of 10) in 91\% of evaluations,
yet mean accuracy remained 56.3\%. This disconnect aligns with the
faithfulness-plausibility gap described in recent reviews of medical LLM
reasoning,\cite{Wang2025-sy} whereby models produce well-structured
reasoning traces that nonetheless arrive at incorrect conclusions.
MedThink-Bench addressed this by evaluating step-by-step rationales
against expert-authored references and achieved strong correlation with
expert judgment (Pearson r = 0.87), but within a static format where all
relevant information is provided upfront.\cite{Zhou2025-pr}
OncoRounds adds a temporal dimension. Even when reasoning quality within
a single solve attempt is high, the cumulative effect of inadequate
information gathering together with the context-length limitations
discussed above can undermine the final diagnosis. High R-IDEA scores
may therefore reflect local coherence within individual solve attempts
rather than cumulative reasoning fidelity across the full clinical
trajectory. This distinction is particularly relevant if reasoning
traces are used to calibrate clinician trust in model outputs. Wysocki
et al.~showed that AI-generated explanations can accentuate model
over-reliance and confirmation bias among healthcare professionals
rather than support critical assessment of model
limitations.\cite{Wysocki2023-ns} Plausible-sounding
rationales that accompany incorrect conclusions could therefore
paradoxically increase overtrust rather than mitigate it.

These findings have implications that extend beyond clinical medicine
into the broader understanding of how large language models reason under
uncertainty. The convergence between LLM and human novice failure modes
- despite fundamentally different architectures and developmental
histories - suggests that information-seeking deficits may be an
emergent property of systems that learn primarily from complete-case
presentations, whether through medical education or training corpora.
Static benchmarks remain valuable for assessing encoded knowledge but do
not capture the sequential decision-making that determines real-world
performance. The gap identified here, between what models know and what
they seek to learn, represents a structural limitation that scaling
alone is unlikely to resolve; the positive correlation between model
release date and performance (R = 0.51, p = 0.002) indicates
generational improvement, yet the 68\% ceiling suggests a qualitative
rather than quantitative barrier. Several directions follow from these
results. First, agentic evaluation paradigms that require active
information gathering should complement static benchmarks as a standard
component of AI assessment. Second, training approaches that explicitly
model sequential hypothesis testing and information-seeking under
uncertainty may address the utilization deficit observed here. Third,
the faithfulness gap between reasoning trace quality and diagnostic
accuracy warrants dedicated investigation, as plausible but incorrect
reasoning chains may paradoxically increase user overtrust rather than
support critical evaluation of model outputs.

\subsection{Limitations}\label{limitations}

This study has limitations. The benchmark comprises 20 cases in a single
subspecialty, all representing complex scenarios involving relapsed,
refractory, or post-transplant presentations. Performance on routine
hematologic conditions or in other medical specialties may differ
substantially. Differential diagnosis and treatment recommendations are
inherently more subjective than the primary diagnosis, and acceptable
answers vary across clinical settings and geographic regions, as
treatment algorithms in the United States differ from those in Europe.
Defining a definitive gold standard at this level of subspecialty
complexity is therefore difficult, though some reference standard
remains necessary for systematic evaluation. The approach taken here,
expert-authored references reviewed by independent subspecialists,
represents one solution but does not eliminate this ambiguity. The
LLM-as-judge scoring approach, while validated against expert
hematologist review (quadratic weighted κ = 0.804), may introduce
systematic bias in nuanced partial-credit decisions, and using a single
model family (GPT-5 Mini) as judge could introduce model-specific
scoring tendencies. All models were accessed through a single API
provider with identical system prompts, and performance could differ
with provider-specific optimizations, agentic scaffolding, or
retrieval-augmented generation. The benchmark architecture supports
custom agent workflows, but implementing model-specific optimizations
for each of 32 candidates was beyond the scope of this evaluation.
Finally, the benchmark evaluates text-based reasoning only and does not
capture the visual information from imaging, pathology slides, and
bedside assessment that informs real clinical decisions.

\subsection{Conclusion}\label{conclusion}

Across 32 frontier large language models evaluated in complex
hematologic oncology, the primary performance-limiting factor was not
medical knowledge but a systematic failure to seek the clinical
information on which treatment decisions depend. Information utilization
was the strongest predictor of accuracy, yet models exhibited search
satisficing, anchoring and premature closure - the same failure modes
documented in novice human clinicians under dual-process models of
diagnostic reasoning. The convergence of these failure patterns across
fundamentally different cognitive architectures suggests that
information-seeking under uncertainty represents a general challenge for
systems trained primarily on complete-case presentations. Addressing
this limitation will require training paradigms that explicitly model
active information gathering and iterative hypothesis testing, moving
beyond the passive interpretation tasks on which current systems excel.

\section{Methods}\label{methods}

\subsection{Study Design}\label{study-design}

OncoRounds is an interactive, turn-based benchmark for evaluating LLM
clinical reasoning in hematologic oncology. The benchmark simulates
clinical decision-making through three progressive rounds of information
disclosure. Round 1 (outpatient presentation) provides vital signs,
chief complaint, and point-of-care diagnostics. Round 2 (inpatient
workup) adds imaging, biopsies, and specialized consultations. Round 3
(definitive characterization) includes full pathology, cytogenetics, and
molecular markers. A detailed treatment history, including prior lines
of therapy and refractoriness status, was withheld until Round 3 to
prevent premature unblinding of disease subtypes in earlier rounds. At
each turn, models choose between requesting a single additional piece of
clinical information or committing to a diagnostic and treatment
solution. This single-request-per-turn constraint enforces clinical
prioritization, requiring models to reason about which information would
be most valuable given what is already known. The benchmark architecture
is depicted in Figure 4. Thirty-two models (Supplementary Table 1) were
evaluated across 20 cases (Table 1), each progressing through all three
clinical rounds.

\subsection{Case Development}\label{case-development}

Twenty clinical cases were developed spanning the spectrum of
hematologic malignancies. Each case comprises a patient profile
(demographics, chief complaint, vital signs), round-specific clinical
settings describing the available diagnostic capabilities, a set of
information items gated by round availability, and a reference standard
for each round. Reference standards consist of a working diagnosis,
differential diagnoses, and a treatment plan in which each action is
annotated with its intention (therapeutic, diagnostic, or supportive)
and priority level (1 through 3). Information items represent clinical
data obtainable in each round\textquotesingle s setting: complete blood
counts, coagulation studies, and bedside imaging in the outpatient
round; tissue biopsies, immunophenotyping, and focused imaging during
inpatient workup; and molecular panels, cytogenetics, and measurable
residual disease assessments in the definitive phase. Cases were
developed de novo by a board-certified hematologist (KB) with 10 years
of experience, rather than adapted from published case reports or public
repositories, to minimize the risk of data leakage into model training
corpora. They were then sent out to board certified hematologist (MH,
LKS, AB, TW, KT, IM, SK) for feedback and suggestions for treatment and
differential diagnosis. Each expert evaluated three cases. Discrepancies
in clinical judgement were resolved in discussion.

\subsection{Models Evaluated}\label{models-evaluated}

Thirty-two models from 12 developers were evaluated: Claude (Haiku 4.5,
Sonnet 4.5, Sonnet 4.6, Opus 4.5, Opus 4.6; Anthropic PBC, San
Francisco, CA, United States), GPT (4.1, 5 Mini, 5 Nano, 5.2 Instant,
5.2 Reasoning, OSS-20B, OSS-120B; OpenAI, San Francisco, CA, United
States), Gemini (3 Flash, 3 Pro, 3.1 Pro; Google LLC Mountain View, CA,
United States), Kimi (K2 thinking, K2.5; Beijing Moonshot AI Technology
Co., Ltd., Beijing, China), DeepSeek (V3.2, V3.2 Speciale; Hangzhou
DeepSeek Artificial Intelligence Basic Technology Research Co., Ltd.,
Hangzhou, Zhejiang, China), MiniMax (M2.1, M2.5; MiniMax Group Inc,
Shanghai, China), Qwen (Qwen3 Max, Qwen3 Next 80B, Qwen3 235B 22B;
Alibaba Group Holding Limited, Hangzhou, Zhejiang, China), Grok (4, 4.1
Fast; X.AI Corp., California, United States), GLM (4.7, 5; Zhipu AI,
Beijing, China), Mistral (Large, Medium 3.1; Mistral AI SAS, Paris,
France), Llama 3.3 70B (Meta Platforms, Inc., Menlo Park, California,
U.S), and Arcee Trinity Large (Arcee AI, Venice, FL, United States).
Total parameter counts ranged from 20 billion to approximately one
trillion, encompassing both proprietary and open-weight architectures.
Of the 32 models, 17 produced full chain-of-thought thinking traces, six
provided only summary-style reasoning outputs (classified as obscured),
and nine produced no reasoning traces. All models were accessed via
unified API (OpenRouter) using identical system prompts and response
schemas.

\subsection{Automated Evaluation
Pipeline}\label{automated-evaluation-pipeline}

Each model\textquotesingle s natural-language information requests were
mapped to standardized case information items by a parser (GPT-5 Mini).
The parser received the candidate\textquotesingle s free-text request
together with the list of available information items and their
round-availability constraints, and returned a structured mapping
indicating whether the request was valid and which item, if any, it
corresponded to.

Solution attempts were evaluated by a judge ensemble of five parallel
GPT-5 Mini instances. Each judge independently compared the
candidate\textquotesingle s working diagnosis, differential diagnoses,
and treatment plan against the round-specific reference standard,
assigning one of three labels to each element: correct (clinically
accurate and complete), partially correct (directionally correct but
incomplete or imprecise), or incorrect. For the working diagnosis, the
correct label required an exact match or accepted synonym; partially
correct was assigned for the correct disease family with a missing
critical modifier. Each candidate differential diagnosis was
independently labeled against the reference set, regardless of how many
the model proposed. Treatment plan items were scored across three
sub-dimensions: action (whether the intervention matched the reference),
intention (therapeutic, diagnostic, or supportive), and priority (scored
correct if matching, partially correct if off by one level, incorrect if
off by two or more). Final labels were determined by majority vote
across the five judges. When no label achieved a strict majority,
partially correct was selected as the tiebreaker, reflecting a
conservative approach to borderline assessments. The judge ensemble was
calibrated against independent scoring by a board-certified hematologist
(KB) across all 20 cases (1,007 paired evaluation items spanning all
three clinical rounds) to verify alignment on labeling criteria and
partial-credit boundaries. Because the three scoring labels (correct,
partially correct, incorrect) form an ordered scale on which a
two-category disagreement carries greater clinical consequence than a
one-category disagreement, quadratic weighted Cohen\textquotesingle s κ
was used to assess inter-rater agreement, yielding substantial agreement
(κ = 0.804). The AI judge was systematically stricter than the human
expert, with 10.7\% of ratings being AI-stricter versus 7.1\%
expert-stricter. The most frequent source of disagreement was the AI
downgrading expert partially correct ratings to incorrect (48 items).

\subsection{Scoring}\label{scoring}

Overall performance was quantified as balanced accuracy, defined as the
unweighted mean of three category-level accuracies: diagnosis,
differential diagnosis, and treatment. Labels were converted to numeric
scores of 1.0 (correct), 0.5 (partially correct), and 0.0 (incorrect),
and category accuracy was computed as the ratio of earned points to
maximum possible points, expressed as a percentage.

For differential diagnoses, a greedy best-N matching strategy was
applied: all candidate labels were sorted by quality (correct first,
then partially correct, then incorrect) and only the top N were counted
toward the score, where N equaled the number of reference differentials.
A model proposing 10 differentials against five references would thus
have its five best-matching labels scored. To discourage list-padding, a
penalty of one point was subtracted for each candidate differential
beyond a threshold of max(10, 2N), with the total floored at zero.
Treatment was scored identically using the same best-N selection and
anti-gaming penalty applied to the three sub-dimensions (action,
intention, priority) jointly, yielding a maximum of three points per
reference treatment item. Scores accumulated across all three clinical
rounds, so that the final score reflected performance from the
outpatient presentation through definitive characterization.

\subsection{Clinical Reasoning
Evaluation}\label{clinical-reasoning-evaluation}

Clinical reasoning quality was assessed using an adapted version of the
Revised-IDEA (R-IDEA) rubric.\cite{Schaye2022-zc} The rubric evaluates
four dimensions: Interpretive Summary (I; 0 to 4 points), reflecting
synthesis of clinical data into a coherent narrative; Differential
Diagnosis (D; 0 to 2), assessing breadth and prioritization of
diagnostic hypotheses; Explanation (E; 0 to 2), capturing mechanistic
reasoning linking findings to diagnoses; and Alternatives (A; 0 to 2),
evaluating consideration of competing hypotheses and disconfirming
evidence. Dimension scores sum to a composite total ranging from 0 to
10, with six or above considered high-quality reasoning. This analysis
was restricted to the 17 models that produced full thinking traces,
yielding 1,008 model-case-round entries. Evaluator agents (Claude Opus
4.6) independently scored each thinking trace given the clinical
scenario and scoring. A random sample of 50 entries was independently
reviewed by the authors (KKB) to verify agent scoring. Bootstrap 95\%
confidence intervals for mean R-IDEA scores were computed using 10,000
replicates with the percentile method.

\subsection{Cognitive Error Taxonomy}\label{cognitive-error-taxonomy}

Reasoning errors were annotated using a taxonomy adapted from
Croskerry\textquotesingle s dual-process model of clinical
reasoning.\cite{Croskerry2009-pe} The taxonomy distinguished
cognitive biases from process errors. Cognitive biases comprised
anchoring bias (fixating on an initial impression despite contradictory
information), confirmation bias (selectively attending to supportive
evidence), premature closure (accepting a diagnosis before adequate
verification), availability bias (overweighting easily recalled
diagnoses), and search satisficing (terminating the diagnostic search
after finding one plausible explanation). Process errors comprised
hallucinated information (incorporating findings not present in the case
data), logical contradiction (making mutually incompatible claims), and
over-differentiating (generating excessively broad differential lists
without stratification). Three error types captured by the
benchmark\textquotesingle s existing scoring mechanism and
information-gathering analysis were excluded: incomplete workup, missed
key findings, and incorrect diagnosis. Each error was graded on a
four-level severity scale (low, medium, high, critical) based on
potential clinical impact, and was judged against the information
available to the model at the time of the error rather than
retrospectively against the complete case. Error annotations were
produced by the same evaluator agents (Claude Opus 4.6) used for R-IDEA
scoring, applied simultaneously during trace evaluation.

\subsection{Information Gathering
Analysis}\label{information-gathering-analysis}

Information-gathering strategy was characterized by two metrics computed
per model-case-round entry. Request precision was defined as the
proportion of valid requests (those successfully matched to an available
information item and released to the model) out of total requests made.
Information utilization was defined as the proportion of available
information items in a given round that were actually requested by the
model.

\subsection{Diagnostic Trajectory
Analysis}\label{diagnostic-trajectory-analysis}

Diagnostic trajectories tracked changes in working diagnosis accuracy
from the first to the final clinical round. Model-case pairs with valid
diagnosis labels across all three rounds were classified into four
mutually exclusive categories: sustained correct (correct at both R1 and
R3), recovered (incorrect at R1 but correct at R3), deteriorated
(correct at R1 but incorrect at R3), and sustained incorrect (incorrect
at both R1 and R3). Partially correct diagnoses were grouped with
incorrect for trajectory classification.

\subsection{Statistical Analysis}\label{statistical-analysis}

Pearson correlations with one-sided tests assessed associations between
model-level information-gathering metrics (request precision and
information utilization). R-IDEA scores and overall benchmark accuracy.
Kruskal-Wallis tests evaluated differences in number of requests across
clinical rounds. Pairwise Wilcoxon rank-sum tests with
Benjamini-Hochberg correction for multiple comparisons were used for
post-hoc group comparisons of R-IDEA scores and error rates. The maximum
sample space comprised 1,920 model-case-round entries (32 models by 20
cases by three rounds), of which 1,896 were completed (634 in R1, 631
each in R2 and R3). R-IDEA evaluation encompassed 1,008 entries across
the 17 models with full thinking traces. Diagnostic trajectory analysis
included 630 complete model-case triads. A p value of \textless{} 0.05
was considered statistically significant.

\subsection{Data Availability}\label{data-availability}

The OncoRounds benchmark dataset, including all 20 clinical cases,
reference standards, and model outputs, will be made publicly available
upon publication at https://github.com/kbressem/oncorounds. Due to the
synthetic nature of the clinical cases, no patient consent or ethical
approval was required.

\subsection{Code Availability}\label{code-availability}

The benchmark evaluation pipeline, including the parser, judge ensemble,
and scoring algorithms, is available at
https://github.com/kbressem/oncorounds.

\subsection{Ethics Statement}\label{ethics-statement}

All clinical cases were developed de novo and do not contain
identifiable patient information. No human subjects were involved in
this study. Institutional review board approval was not required.

\subsection{Author Contributions}\label{author-contributions}

KB designed the study and developed all clinical cases. KKB and LCA
jointly conceived the study and supervised the project. KKB implemented
the benchmark evaluation pipeline and conducted statistical analyses.
LCA and KKB contributed to data interpretation and manuscript
preparation. LKS, TW, AB, IM, KT, SK, and MH provided expert hematologic
review of clinical cases and reference standards. SZ contributed to data
analysis. FB and MM provided critical revision of the manuscript. All
authors reviewed and approved the final manuscript.

\subsection{Competing Interests}\label{competing-interests}

FB received honoraria and/or travel/accommodation expenses from Bristol
Myers Squibb (BMS), AbbVie, Amgen, and Johnson \& Johnson. KKB reports
speaker fees from Canon Medical Systems and GE Healthcare; and Data
Safety Monitoring Board participation for Philips (IHI Project IMAGIO).
All other authors declare no competing interests.

\subsection{Acknowledgements}\label{acknowledgements}

We thank Oliver Weigert and Martin Dreyling (Department of Medicine III,
LMU University Hospital, Munich) for helpful discussions on lymphoma
cases. This work was supported by the European Research Council
(Consolidator Grant BCM-UPS-682473 {[}FB{]}), a TUM Innovation Network
Grant (NextGenDrugs {[}FB{]}), the Else Kröner Fresenius Foundation
(Exzellenzstipendium {[}LCA, KKB{]}), and the Wilhelm Sander Foundation
({[}LCA, KKB{]}).

\subsection{Role of the Funding
Source}\label{role-of-the-funding-source}

The funders of the study had no role in study design, data collection,
data analysis, data interpretation, or writing of the report. The
corresponding author had full access to all the data in the study and
had final responsibility for the decision to submit for publication.

\bibliographystyle{unsrtnat}
\bibliography{references}

@ARTICLE{Wasterlid2022-el,
  title     = "Application of precision medicine in clinical routine in
               haematology-Challenges and opportunities",
  author    = "W{\"a}sterlid, Tove and Cavelier, Lucia and Haferlach, Claudia
               and Konopleva, Marina and Fr{\"o}hling, Stefan and {\"O}stling,
               P{\"a}ivi and Bullinger, Lars and Fioretos, Thoas and Smedby,
               Karin E",
  journal   = "J. Intern. Med.",
  publisher = "Wiley",
  volume    =  292,
  number    =  2,
  pages     = "243--261",
  month     =  aug,
  year      =  2022,
  copyright = "http://creativecommons.org/licenses/by-nc/4.0/",
  language  = "en",
  doi           = "10.1111/joim.13508"
}

@ARTICLE{Singhal2023-bm,
  title     = "Large language models encode clinical knowledge",
  author    = "Singhal, Karan and Azizi, Shekoofeh and Tu, Tao and Mahdavi, S
               Sara and Wei, Jason and Chung, Hyung Won and Scales, Nathan and
               Tanwani, Ajay and Cole-Lewis, Heather and Pfohl, Stephen and
               Payne, Perry and Seneviratne, Martin and Gamble, Paul and Kelly,
               Chris and Babiker, Abubakr and Sch{\"a}rli, Nathanael and
               Chowdhery, Aakanksha and Mansfield, Philip and Demner-Fushman,
               Dina and Ag{\"u}era Y Arcas, Blaise and Webster, Dale and
               Corrado, Greg S and Matias, Yossi and Chou, Katherine and
               Gottweis, Juraj and Tomasev, Nenad and Liu, Yun and Rajkomar,
               Alvin and Barral, Joelle and Semturs, Christopher and
               Karthikesalingam, Alan and Natarajan, Vivek",
  journal   = "Nature",
  publisher = "Springer Science and Business Media LLC",
  volume    =  620,
  number    =  7972,
  pages     = "172--180",
  month     =  aug,
  year      =  2023,
  copyright = "https://creativecommons.org/licenses/by/4.0",
  language  = "en",
  doi           = "10.1038/s41586-023-06291-2"
}

@MISC{Nori2023-aj,
  title         = "Capabilities of {GPT-4} on medical challenge problems",
  author        = "Nori, Harsha and King, Nicholas and McKinney, Scott Mayer
                   and Carignan, Dean and Horvitz, Eric",
  month         =  mar,
  year          =  2023,
  copyright     = "http://arxiv.org/licenses/nonexclusive-distrib/1.0/",
  archivePrefix = "arXiv",
  primaryClass  = "cs.CL",
  eprint        = "2303.13375",
  howpublished = "arXiv preprint arXiv:2303.13375",
  url          = "https://arxiv.org/abs/2303.13375",
  doi           = "10.48550/arXiv.2303.13375"
}

@ARTICLE{Gilson2023-jp,
  title     = "How does {ChatGPT} perform on the United States Medical
               Licensing Examination ({USMLE)}? The implications of large
               language models for medical education and knowledge assessment",
  author    = "Gilson, Aidan and Safranek, Conrad W and Huang, Thomas and
               Socrates, Vimig and Chi, Ling and Taylor, Richard Andrew and
               Chartash, David",
  journal   = "JMIR Med. Educ.",
  publisher = "JMIR Publications Inc.",
  volume    =  9,
  pages     = "e45312",
  month     =  feb,
  year      =  2023,
  language  = "en",
  doi           = "10.2196/45312"
}

@ARTICLE{Ferber2024-vz,
  title     = "{GPT-4} for information retrieval and comparison of medical
               oncology guidelines",
  author    = "Ferber, Dyke and Wiest, Isabella C and W{\"o}lflein, Georg and
               Ebert, Matthias P and Beutel, Gernot and Eckardt, Jan-Niklas and
               Truhn, Daniel and Springfeld, Christoph and J{\"a}ger, Dirk and
               Kather, Jakob Nikolas",
  journal   = "NEJM AI",
  publisher = "Massachusetts Medical Society",
  volume    =  1,
  number    =  6,
  month     =  may,
  year      =  2024,
  language  = "en",
  doi           = "10.1056/AIcs2300235"
}

@ARTICLE{Truhn2023-br,
  title    = "Large language models should be used as scientific reasoning
              engines, not knowledge databases",
  author   = "Truhn, Daniel and Reis-Filho, Jorge S and Kather, Jakob Nikolas",
  journal  = "Nat. Med.",
  volume   =  29,
  number   =  12,
  pages    = "2983--2984",
  month    =  dec,
  year     =  2023,
  language = "en",
  doi           = "10.1038/s41591-023-02594-z"
}

@ARTICLE{Thirunavukarasu2023-gr,
  title    = "Large language models in medicine",
  author   = "Thirunavukarasu, Arun James and Ting, Darren Shu Jeng and
              Elangovan, Kabilan and Gutierrez, Laura and Tan, Ting Fang and
              Ting, Daniel Shu Wei",
  journal  = "Nat. Med.",
  volume   =  29,
  number   =  8,
  pages    = "1930--1940",
  month    =  aug,
  year     =  2023,
  language = "en",
  doi           = "10.1038/s41591-023-02448-8"
}

@ARTICLE{Acosta2022-yr,
  title     = "Multimodal biomedical {AI}",
  author    = "Acosta, Juli{\'a}n N and Falcone, Guido J and Rajpurkar, Pranav
               and Topol, Eric J",
  journal   = "Nat. Med.",
  publisher = "Springer Science and Business Media LLC",
  volume    =  28,
  number    =  9,
  pages     = "1773--1784",
  month     =  sep,
  year      =  2022,
  language  = "en",
  doi           = "10.1038/s41591-022-01981-2"
}

@ARTICLE{Lipkova2022-ol,
  title     = "Artificial intelligence for multimodal data integration in
               oncology",
  author    = "Lipkova, Jana and Chen, Richard J and Chen, Bowen and Lu, Ming Y
               and Barbieri, Matteo and Shao, Daniel and Vaidya, Anurag J and
               Chen, Chengkuan and Zhuang, Luoting and Williamson, Drew F K and
               Shaban, Muhammad and Chen, Tiffany Y and Mahmood, Faisal",
  journal   = "Cancer Cell",
  publisher = "Elsevier BV",
  volume    =  40,
  number    =  10,
  pages     = "1095--1110",
  month     =  oct,
  year      =  2022,
  copyright = "http://creativecommons.org/licenses/by-nc-nd/4.0/",
  language  = "en",
  doi           = "10.1016/j.ccell.2022.09.012"
}

@ARTICLE{Moor2023-rh,
  title     = "Foundation models for generalist medical artificial intelligence",
  author    = "Moor, Michael and Banerjee, Oishi and Abad, Zahra Shakeri
               Hossein and Krumholz, Harlan M and Leskovec, Jure and Topol,
               Eric J and Rajpurkar, Pranav",
  journal   = "Nature",
  publisher = "Springer Science and Business Media LLC",
  volume    =  616,
  number    =  7956,
  pages     = "259--265",
  month     =  apr,
  year      =  2023,
  copyright = "https://www.springernature.com/gp/researchers/text-and-data-mining",
  language  = "en",
  doi           = "10.1038/s41586-023-05881-4"
}

@ARTICLE{Tu2024-pl,
  title     = "Towards generalist biomedical {AI}",
  author    = "Tu, Tao and Azizi, Shekoofeh and Driess, Danny and Schaekermann,
               Mike and Amin, Mohamed and Chang, Pi-Chuan and Carroll, Andrew
               and Lau, Charles and Tanno, Ryutaro and Ktena, Ira and Palepu,
               Anil and Mustafa, Basil and Chowdhery, Aakanksha and Liu, Yun
               and Kornblith, Simon and Fleet, David and Mansfield, Philip and
               Prakash, Sushant and Wong, Renee and Virmani, Sunny and Semturs,
               Christopher and Mahdavi, S Sara and Green, Bradley and
               Dominowska, Ewa and Arcas, Blaise Aguera y and Barral, Joelle
               and Webster, Dale and Corrado, Greg S and Matias, Yossi and
               Singhal, Karan and Florence, Pete and Karthikesalingam, Alan and
               Natarajan, Vivek",
  journal   = "NEJM AI",
  publisher = "Massachusetts Medical Society",
  volume    =  1,
  number    =  3,
  month     =  feb,
  year      =  2024,
  language  = "en",
  doi           = "10.1056/AIoa2300138"
}

@ARTICLE{Ferber2025-bs,
  title     = "Development and validation of an autonomous artificial
               intelligence agent for clinical decision-making in oncology",
  author    = "Ferber, Dyke and El Nahhas, Omar S M and W{\"o}lflein, Georg and
               Wiest, Isabella C and Clusmann, Jan and Le{\ss}mann,
               Marie-Elisabeth and Foersch, Sebastian and Lammert, Jacqueline
               and Tschochohei, Maximilian and J{\"a}ger, Dirk and
               Salto-Tellez, Manuel and Schultz, Nikolaus and Truhn, Daniel and
               Kather, Jakob Nikolas",
  journal   = "Nat. Cancer",
  publisher = "Springer Science and Business Media LLC",
  volume    =  6,
  number    =  8,
  pages     = "1337--1349",
  month     =  aug,
  year      =  2025,
  copyright = "https://creativecommons.org/licenses/by/4.0",
  language  = "en",
  doi           = "10.1038/s43018-025-00991-6"
}

@ARTICLE{Lee2024-mc,
  title     = "How {AI} agents will change cancer research and oncology",
  author    = "Lee, Yongju and Ferber, Dyke and Rood, Jennifer E and Regev,
               Aviv and Kather, Jakob Nikolas",
  journal   = "Nat. Cancer",
  publisher = "Springer Science and Business Media LLC",
  volume    =  5,
  number    =  12,
  pages     = "1765--1767",
  month     =  dec,
  year      =  2024,
  copyright = "https://www.springernature.com/gp/researchers/text-and-data-mining",
  language  = "en",
  doi           = "10.1038/s43018-024-00861-7"
}

@ARTICLE{Bedi2026-he,
  title    = "Holistic evaluation of large language models for medical tasks
              with {MedHELM}",
  author   = "Bedi, Suhana and Cui, Hejie and Fuentes, Miguel and Unell, Alyssa
              and Wornow, Michael and Banda, Juan M and Kotecha, Nikesh and
              Keyes, Timothy and Mai, Yifan and Oez, Mert and Qiu, Hao and
              Jain, Shrey and Schettini, Leonardo and Kashyap, Mehr and Fries,
              Jason Alan and Swaminathan, Akshay and Chung, Philip and
              Haredasht, Fateme Nateghi and Lopez, Ivan and Aali, Asad and Tse,
              Gabriel and Nayak, Ashwin and Vedak, Shivam and Jain, Sneha S and
              Patel, Birju and Fayanju, Oluseyi and Shah, Shreya and Goh, Ethan
              and Yao, Dong-Han and Soetikno, Brian and Reis, Eduardo and
              Gatidis, Sergios and Divi, Vasu and Capasso, Robson and Saralkar,
              Rachna and Chiang, Chia-Chun and Jindal, Jenelle and Pham, Tho
              and Ghoddusi, Faraz and Lin, Steven and Chiou, Albert S and Hong,
              Hyo Jung and Roy, Mohana and Gensheimer, Michael F and Patel,
              Hinesh and Schulman, Kevin and Dash, Dev and Char, Danton and
              Downing, Lance and Grolleau, Francois and Black, Kameron and
              Mieso, Bethel and Zahedivash, Aydin and Yim, Wen-Wai and Sharma,
              Harshita and Lee, Tony and Kirsch, Hannah and Lee, Jennifer and
              Ambers, Nerissa and Lugtu, Carlene and Sharma, Aditya and Mawji,
              Bilal and Alekseyev, Alex and Zhou, Vicky and Kakkar, Vikas and
              Helzer, Jarrod and Revri, Anurang and Bannett, Yair and
              Daneshjou, Roxana and Chen, Jonathan and Alsentzer, Emily and
              Morse, Keith and Ravi, Nirmal and Aghaeepour, Nima and Kennedy,
              Vanessa and Chaudhari, Akshay and Wang, Thomas and Koyejo, Sanmi
              and Lungren, Matthew P and Horvitz, Eric and Liang, Percy and
              Pfeffer, Michael A and Shah, Nigam H",
  journal  = "Nat. Med.",
  volume   =  32,
  number   =  3,
  pages    = "943--951",
  month    =  jan,
  year     =  2026,
  language = "en",
  doi           = "10.1038/s41591-025-04151-2"
}

@ARTICLE{Tordjman2025-hw,
  title     = "Comparative benchmarking of the {DeepSeek} large language model
               on medical tasks and clinical reasoning",
  author    = "Tordjman, Mickael and Liu, Zelong and Yuce, Murat and Fauveau,
               Valentin and Mei, Yunhao and Hadjadj, Jerome and Bolger, Ian and
               Almansour, Haidara and Horst, Carolyn and Parihar, Ashwin Singh
               and Geahchan, Amine and Meribout, Anis and Yatim, Nader and Ng,
               Nicole and Robson, Phillip and Zhou, Alexander and Lewis, Sara
               and Huang, Mingqian and Deyer, Timothy and Taouli, Bachir and
               Lee, Hao-Chih and Fayad, Zahi A and Mei, Xueyan",
  journal   = "Nat. Med.",
  publisher = "Springer Science and Business Media LLC",
  volume    =  31,
  number    =  8,
  pages     = "2550--2555",
  month     =  aug,
  year      =  2025,
  copyright = "https://www.springernature.com/gp/researchers/text-and-data-mining",
  language  = "en",
  doi           = "10.1038/s41591-025-03726-3"
}

@ARTICLE{Singhal2025-fw,
  title     = "Toward expert-level medical question answering with large
               language models",
  author    = "Singhal, Karan and Tu, Tao and Gottweis, Juraj and Sayres, Rory
               and Wulczyn, Ellery and Amin, Mohamed and Hou, Le and Clark,
               Kevin and Pfohl, Stephen R and Cole-Lewis, Heather and Neal,
               Darlene and Rashid, Qazi Mamunur and Schaekermann, Mike and
               Wang, Amy and Dash, Dev and Chen, Jonathan H and Shah, Nigam H
               and Lachgar, Sami and Mansfield, Philip Andrew and Prakash,
               Sushant and Green, Bradley and Dominowska, Ewa and Ag{\"u}era Y
               Arcas, Blaise and Toma{\v s}ev, Nenad and Liu, Yun and Wong,
               Renee and Semturs, Christopher and Mahdavi, S Sara and Barral,
               Joelle K and Webster, Dale R and Corrado, Greg S and Matias,
               Yossi and Azizi, Shekoofeh and Karthikesalingam, Alan and
               Natarajan, Vivek",
  journal   = "Nat. Med.",
  publisher = "Springer Science and Business Media LLC",
  volume    =  31,
  number    =  3,
  pages     = "943--950",
  month     =  mar,
  year      =  2025,
  copyright = "https://creativecommons.org/licenses/by-nc-nd/4.0",
  language  = "en",
  doi           = "10.1038/s41591-024-03423-7"
}

@MISC{Valmeekam2023-ti,
  title         = "On the planning abilities of large language models : A
                   critical investigation",
  author        = "Valmeekam, Karthik and Marquez, Matthew and Sreedharan,
                   Sarath and Kambhampati, Subbarao",
  month         =  may,
  year          =  2023,
  copyright     = "http://arxiv.org/licenses/nonexclusive-distrib/1.0/",
  archivePrefix = "arXiv",
  primaryClass  = "cs.AI",
  eprint        = "2305.15771",
  howpublished = "arXiv preprint arXiv:2305.15771",
  url          = "https://arxiv.org/abs/2305.15771",
  doi           = "10.48550/arXiv.2305.15771"
}

@MISC{Cao2025-fj,
  title         = "Large Language Models for planning: A comprehensive and
                   systematic survey",
  author        = "Cao, Pengfei and Men, Tianyi and Liu, Wencan and Zhang,
                   Jingwen and Li, Xuzhao and Lin, Xixun and Sui, Dianbo and
                   Cao, Yanan and Liu, Kang and Zhao, Jun",
  month         =  may,
  year          =  2025,
  copyright     = "http://arxiv.org/licenses/nonexclusive-distrib/1.0/",
  archivePrefix = "arXiv",
  primaryClass  = "cs.AI",
  eprint        = "2505.19683",
  howpublished = "arXiv preprint arXiv:2505.19683",
  url          = "https://arxiv.org/abs/2505.19683",
  doi           = "10.48550/arXiv.2505.19683"
}

@MISC{Gu2025-iw,
  title         = "A Survey on {LLM-as-a-Judge}",
  author        = "Gu, Jiawei and Jiang, Xuhui and Shi, Zhichao and Tan,
                   Hexiang and Zhai, Xuehao and Xu, Chengjin and Li, Wei and
                   Shen, Yinghan and Ma, Shengjie and Liu, Honghao and Wang,
                   Saizhuo and Zhang, Kun and Wang, Yuanzhuo and Gao, Wen and
                   Ni, Lionel and Guo, Jian",
  month         =  oct,
  year          =  2025,
  copyright     = "http://creativecommons.org/publicdomain/zero/1.0/",
  archivePrefix = "arXiv",
  primaryClass  = "cs.CL",
  eprint        = "2411.15594",
  howpublished = "arXiv preprint arXiv:2411.15594",
  url          = "https://arxiv.org/abs/2411.15594",
  doi           = "10.48550/arXiv.2411.15594"
}

@ARTICLE{Schaye2022-zc,
  title     = "Development of a clinical reasoning documentation assessment
               tool for resident and fellow admission notes: A shared mental
               model for feedback",
  author    = "Schaye, Verity and Miller, Louis and Kudlowitz, David and Chun,
               Jonathan and Burk-Rafel, Jesse and Cocks, Patrick and Guzman,
               Benedict and Aphinyanaphongs, Yindalon and Marin, Marina",
  journal   = "J. Gen. Intern. Med.",
  publisher = "Springer Science and Business Media LLC",
  volume    =  37,
  number    =  3,
  pages     = "507--512",
  month     =  feb,
  year      =  2022,
  language  = "en",
  doi           = "10.1007/s11606-021-06805-6"
}

@ARTICLE{Croskerry2009-pe,
  title     = "A universal model of diagnostic reasoning",
  author    = "Croskerry, Pat",
  journal   = "Acad. Med.",
  publisher = "Oxford University Press (OUP)",
  volume    =  84,
  number    =  8,
  pages     = "1022--1028",
  month     =  aug,
  year      =  2009,
  language  = "en",
  doi           = "10.1097/ACM.0b013e3181ace703"
}

@MISC{Schmidgall2024-ht,
  title         = "{AgentClinic}: a multimodal agent benchmark to evaluate {AI}
                   in simulated clinical environments",
  author        = "Schmidgall, Samuel and Ziaei, Rojin and Harris, Carl and
                   Reis, Eduardo and Jopling, Jeffrey and Moor, Michael",
  month         =  may,
  year          =  2024,
  copyright     = "http://creativecommons.org/licenses/by/4.0/",
  archivePrefix = "arXiv",
  primaryClass  = "cs.HC",
  eprint        = "2405.07960",
  howpublished = "arXiv preprint arXiv:2405.07960",
  url          = "https://arxiv.org/abs/2405.07960",
  doi           = "10.48550/arXiv.2405.07960"
}

@MISC{Gong2026-yh,
  title         = "{MedDialogRubrics}: A comprehensive benchmark and evaluation
                   framework for multi-turn medical consultations in large
                   language models",
  author        = "Gong, Lecheng and Fang, Weimin and Yang, Ting and Tao,
                   Dongjie and Guo, Chunxiao and Wei, Peng and Xie, Bo and
                   Guan, Jinqun and Chen, Zixiao and Shi, Fang and Gu, Jinjie
                   and Liu, Junwei",
  month         =  jan,
  year          =  2026,
  copyright     = "http://arxiv.org/licenses/nonexclusive-distrib/1.0/",
  archivePrefix = "arXiv",
  primaryClass  = "cs.CL",
  eprint        = "2601.03023",
  howpublished = "arXiv preprint arXiv:2601.03023",
  url          = "https://arxiv.org/abs/2601.03023",
  doi           = "10.48550/arXiv.2601.03023"
}

@MISC{Almansoori2025-uf,
  title         = "Self-evolving multi-agent simulations for realistic clinical
                   interactions",
  author        = "Almansoori, Mohammad and Kumar, Komal and Cholakkal, Hisham",
  month         =  oct,
  year          =  2025,
  copyright     = "http://creativecommons.org/licenses/by-nc-sa/4.0/",
  archivePrefix = "arXiv",
  primaryClass  = "cs.CL",
  eprint        = "2503.22678",
  howpublished = "arXiv preprint arXiv:2503.22678",
  url          = "https://arxiv.org/abs/2503.22678",
  doi           = "10.48550/arXiv.2503.22678"
}

@MISC{Liu2023-sm,
  title         = "Lost in the middle: How language models use long contexts",
  author        = "Liu, Nelson F and Lin, Kevin and Hewitt, John and Paranjape,
                   Ashwin and Bevilacqua, Michele and Petroni, Fabio and Liang,
                   Percy",
  month         =  jul,
  year          =  2023,
  copyright     = "http://arxiv.org/licenses/nonexclusive-distrib/1.0/",
  archivePrefix = "arXiv",
  primaryClass  = "cs.CL",
  eprint        = "2307.03172",
  howpublished = "arXiv preprint arXiv:2307.03172",
  url          = "https://arxiv.org/abs/2307.03172",
  doi           = "10.48550/arXiv.2307.03172"
}

@MISC{Gupta2024-dz,
  title         = "Systematic evaluation of long-context {LLMs} on financial
                   concepts",
  author        = "Gupta, Lavanya and Sharma, Saket and Zhao, Yiyun",
  month         =  dec,
  year          =  2024,
  copyright     = "http://creativecommons.org/licenses/by/4.0/",
  archivePrefix = "arXiv",
  primaryClass  = "cs.CL",
  eprint        = "2412.15386",
  howpublished = "arXiv preprint arXiv:2412.15386",
  url          = "https://arxiv.org/abs/2412.15386",
  doi           = "10.48550/arXiv.2412.15386"
}

@MISC{Laban2025-ns,
  title         = "{LLMs} Get Lost In {Multi-Turn} Conversation",
  author        = "Laban, Philippe and Hayashi, Hiroaki and Zhou, Yingbo and
                   Neville, Jennifer",
  month         =  may,
  year          =  2025,
  copyright     = "http://creativecommons.org/licenses/by/4.0/",
  archivePrefix = "arXiv",
  primaryClass  = "cs.CL",
  eprint        = "2505.06120",
  howpublished = "arXiv preprint arXiv:2505.06120",
  url          = "https://arxiv.org/abs/2505.06120",
  doi           = "10.48550/arXiv.2505.06120"
}

@ARTICLE{Moell2025-wa,
  title    = "Medical reasoning in {LLMs}: an in-depth analysis of {DeepSeek}
              {R1}",
  author   = "Mo{\"e}ll, Birger and Sand Aronsson, Fredrik and Akbar, Sanian",
  journal  = "Front. Artif. Intell.",
  volume   =  8,
  pages    = "1616145",
  month    =  jun,
  year     =  2025,
  language = "en",
  doi           = "10.3389/frai.2025.1616145"
}

@MISC{Schmidgall2024-ra,
  title         = "Addressing cognitive bias in medical language models",
  author        = "Schmidgall, Samuel and Harris, Carl and Essien, Ime and
                   Olshvang, Daniel and Rahman, Tawsifur and Kim, Ji Woong and
                   Ziaei, Rojin and Eshraghian, Jason and Abadir, Peter and
                   Chellappa, Rama",
  month         =  feb,
  year          =  2024,
  copyright     = "http://creativecommons.org/licenses/by/4.0/",
  archivePrefix = "arXiv",
  primaryClass  = "cs.CL",
  eprint        = "2402.08113",
  howpublished = "arXiv preprint arXiv:2402.08113",
  url          = "https://arxiv.org/abs/2402.08113",
  doi           = "10.48550/arXiv.2402.08113"
}

@ARTICLE{Ke2024-ea,
  title     = "Mitigating cognitive biases in clinical decision-making through
               multi-agent conversations using large language models:
               Simulation study",
  author    = "Ke, Yuhe and Yang, Rui and Lie, Sui An and Lim, Taylor Xin Yi
               and Ning, Yilin and Li, Irene and Abdullah, Hairil Rizal and
               Ting, Daniel Shu Wei and Liu, Nan",
  journal   = "J. Med. Internet Res.",
  publisher = "JMIR Publications Inc.",
  volume    =  26,
  pages     = "e59439",
  month     =  nov,
  year      =  2024,
  language  = "en",
  doi           = "10.2196/59439"
}

@MISC{Wang2025-sy,
  title         = "Medical reasoning in the era of {LLMs}: A systematic review
                   of enhancement techniques and applications",
  author        = "Wang, Wenxuan and Ma, Zizhan and Ding, Meidan and Zheng,
                   Shiyi and Liu, Shengyuan and Liu, Jie and Ji, Jiaming and
                   Chen, Wenting and Li, Xiang and Shen, Linlin and Yuan,
                   Yixuan",
  month         =  aug,
  year          =  2025,
  copyright     = "http://arxiv.org/licenses/nonexclusive-distrib/1.0/",
  archivePrefix = "arXiv",
  primaryClass  = "cs.CL",
  eprint        = "2508.00669",
  howpublished = "arXiv preprint arXiv:2508.00669",
  url          = "https://arxiv.org/abs/2508.00669",
  doi           = "10.48550/arXiv.2508.00669"
}

@ARTICLE{Zhou2025-pr,
  title     = "Automating expert-level medical reasoning evaluation of large
               language models",
  author    = "Zhou, Shuang and Xie, Wenya and Li, Jiaxi and Zhan, Zaifu and
               Song, Meijia and Yang, Han and Espinoza, Cheyenna and Welton,
               Lindsay and Mai, Xinnie and Jin, Yanwei and Xu, Zidu and Chung,
               Yuen-Hei and Xing, Yiyun and Tsai, Meng-Han and Schaffer, Emma
               and Shi, Yucheng and Liu, Ninghao and Liu, Zirui and Zhang, Rui",
  journal   = "NPJ Digit. Med.",
  publisher = "Springer Science and Business Media LLC",
  volume    =  9,
  number    =  1,
  pages     = "34",
  month     =  dec,
  year      =  2025,
  copyright = "https://creativecommons.org/licenses/by/4.0",
  language  = "en",
  doi           = "10.1038/s41746-025-02208-7"
}

@ARTICLE{Wysocki2023-ns,
  title     = "Assessing the communication gap between {AI} models and
               healthcare professionals: Explainability, utility and trust in
               {AI-driven} clinical decision-making",
  author    = "Wysocki, Oskar and Davies, Jessica Katharine and Vigo, Markel
               and Armstrong, Anne Caroline and Landers, D{\'o}nal and Lee,
               Rebecca and Freitas, Andr{\'e}",
  journal   = "Artif. Intell.",
  publisher = "Elsevier BV",
  volume    =  316,
  pages     = "103839",
  month     =  mar,
  year      =  2023,
  copyright = "http://creativecommons.org/licenses/by/4.0/",
  language  = "en",
  doi           = "10.1016/j.artint.2022.103839"
}

\section{Tables}\label{tables}

\subsection{Table 1: Case Overview}\label{table-1-case-overview}

\begin{longtable}[]{@{}
  >{\raggedright\arraybackslash}p{(\columnwidth - 8\tabcolsep) * \real{0.0378}}
  >{\raggedright\arraybackslash}p{(\columnwidth - 8\tabcolsep) * \real{0.1189}}
  >{\raggedright\arraybackslash}p{(\columnwidth - 8\tabcolsep) * \real{0.0378}}
  >{\raggedright\arraybackslash}p{(\columnwidth - 8\tabcolsep) * \real{0.6811}}
  >{\raggedright\arraybackslash}p{(\columnwidth - 8\tabcolsep) * \real{0.1081}}@{}}
\toprule\noalign{}
\begin{minipage}[b]{\linewidth}\raggedright
Case
\end{minipage} & \begin{minipage}[b]{\linewidth}\raggedright
Diagnosis
\end{minipage} & \begin{minipage}[b]{\linewidth}\raggedright
Age/

Sex
\end{minipage} & \begin{minipage}[b]{\linewidth}\raggedright
Case Title*
\end{minipage} & \begin{minipage}[b]{\linewidth}\raggedright
Mean Accuracy (\%)
\end{minipage} \\
\midrule\noalign{}
\endhead
\bottomrule\noalign{}
\endlastfoot
1 & AML & 35/M & Relapsed AML with FLT3-ITD in a 35-year-old male &
48.1 \\
2 & AML & 24/F & Post-transplant extramedullary relapse of NPM1-mutated
AML in a 24-year-old female & 60.7 \\
3 & AML & 71/M & Relapsed/refractory AML-MR with complex karyotype in a
71-year-old male & 58.2 \\
4 & MDS & 56/F & Relapsed high-risk MDS with TP53 mutation and severe
chronic GvHD in a 56-year-old female after alloSCT & 49.0 \\
5 & AML & 56/F & AML-pCT in a 56-year-old female with prior multiple
myeloma & 59.1 \\
6 & BPDCN & 67/M & Relapsed/refractory BPDCN after alloSCT in a
67-year-old male & 56.9 \\
7 & DLBCL & 70/M & Relapsed high-grade B-cell lymphoma with MYC and BCL2
rearrangements in a 70-year-old male & 52.1 \\
8 & AML & 78/M & Relapsed AML-MR with septic shock in a 78-year-old male
after alloSCT & 67.3 \\
9 & Multiple Myeloma & 81/M & Relapsed/refractory IgG kappa multiple
myeloma with therapy-related MDS in an 81-year-old male & 46.4 \\
10 & ALL & 21/F & Early T-cell precursor acute lymphoblastic leukemia
with molecular failure in a 21-year-old female & 59.7 \\
11 & DLBCL & 32/F & High-grade B-cell lymphoma with MYC and BCL2
rearrangements in a 32-year-old female with prior classical Hodgkin
lymphoma & 47.5 \\
12 & ALCL & 67/F & Relapsed ALK-negative anaplastic large cell lymphoma
after haploidentical alloSCT in a 67-year-old female & 49.5 \\
13 & Multiple Myeloma & 67/M & Relapsed triple-class refractory multiple
myeloma in a 67-year-old male & 52.7 \\
14 & AML & 27/F & Relapsed AML with mutated TP53 after second alloSCT
and invasive fungal pneumonia in a 27-year-old female & 48.4 \\
15 & Multiple Myeloma & 75/M & Relapsed/refractory multiple myeloma with
secondary TP53-mutated MDS in a 75-year-old male & 51.9 \\
16 & Follicular Lymphoma & 55/M & Early post-CAR-T relapse of follicular
lymphoma grade 1-2 in a 55-year-old male & 53.2 \\
17 & ALL & 44/M & Molecular relapse of B-ALL with CD19/CD22-negative
AUL-like phenotype in a 44-year-old male after alloSCT and CD19 CAR-T &
59.5 \\
18 & Aplastic Anemia & 34/F & Refractory very severe aplastic anemia
with large PNH clone and monosomy 7 in a 34-year-old female with
invasive fusariosis & 66.1 \\
19 & AML & 76/F & AML-pCT with TP53 mutation after follicular lymphoma
in a 76-year-old woman with recent traumatic ICH & 72.0 \\
20 & CLL/Richter & 77/M & Richter transformation with BTK C481S mutation
after multiple CLL therapies & 68.6 \\
\end{longtable}

* Case title not disclosed to the LLM during evaluation.

\subsection{Table 2: Overall
Performance}\label{table-2-overall-performance}

\begin{longtable}[]{@{}
  >{\raggedright\arraybackslash}p{(\columnwidth - 10\tabcolsep) * \real{0.0714}}
  >{\raggedright\arraybackslash}p{(\columnwidth - 10\tabcolsep) * \real{0.2551}}
  >{\raggedright\arraybackslash}p{(\columnwidth - 10\tabcolsep) * \real{0.1122}}
  >{\raggedright\arraybackslash}p{(\columnwidth - 10\tabcolsep) * \real{0.1633}}
  >{\raggedright\arraybackslash}p{(\columnwidth - 10\tabcolsep) * \real{0.1939}}
  >{\raggedright\arraybackslash}p{(\columnwidth - 10\tabcolsep) * \real{0.1633}}@{}}
\toprule\noalign{}
\begin{minipage}[b]{\linewidth}\raggedright
Rank
\end{minipage} & \begin{minipage}[b]{\linewidth}\raggedright
Model
\end{minipage} & \begin{minipage}[b]{\linewidth}\raggedright
Overall

(\%)
\end{minipage} & \begin{minipage}[b]{\linewidth}\raggedright
Diagnosis (\%)
\end{minipage} & \begin{minipage}[b]{\linewidth}\raggedright
Differential (\%)
\end{minipage} & \begin{minipage}[b]{\linewidth}\raggedright
Treatment (\%)
\end{minipage} \\
\midrule\noalign{}
\endhead
\bottomrule\noalign{}
\endlastfoot
1 & Claude Opus 4.6 & \textbf{68.1} & \textbf{84.2} & 55.2 & 65.1 \\
2 & MiniMax M2.1 & 66.3 & 72.5 & \textbf{61.4} & 65.0 \\
3 & Claude Opus 4.5 & 64.9 & 81.7 & 49.8 & 63.2 \\
4 & Claude Sonnet 4.6 & 64.8 & 77.5 & 51.8 & 65.2 \\
5 & MiniMax M2.5 & 64.3 & 72.5 & 57.7 & 62.8 \\
6 & Claude Sonnet 4.5 & 63.8 & 72.8 & 51.9 & 66.8 \\
7 & GPT-5.2 Reasoning & 62.3 & 76.3 & 47.3 & 63.3 \\
8 & Claude Haiku 4.5 & 62.2 & 67.5 & 54.1 & 65.1 \\
9 & GPT-5.2 Instant & 62.1 & 78.3 & 46.5 & 61.4 \\
10 & GPT-5 Mini & 61.2 & 65.4 & 47.5 & \textbf{70.7} \\
11 & Kimi K2.5 & 60.9 & 70.8 & 50.1 & 61.7 \\
12 & Kimi K2 Thinking & 59.0 & 60.0 & 52.2 & 64.7 \\
13 & Gemini 3 Pro & 58.8 & 69.2 & 46.4 & 60.8 \\
14 & GLM 5 & 58.6 & 70.0 & 45.8 & 59.9 \\
15 & Grok 4 & 58.4 & 74.2 & 39.2 & 61.8 \\
16 & GPT-4.1 & 58.1 & 72.5 & 45.0 & 56.9 \\
17 & Gemini 3 Flash & 57.4 & 63.3 & 53.3 & 55.5 \\
18 & DeepSeek V3.2 Speciale & 57.1 & 70.0 & 38.9 & 62.4 \\
19 & GPT-OSS 120B & 56.4 & 71.7 & 40.6 & 56.9 \\
20 & DeepSeek V3.2 & 55.9 & 70.0 & 43.3 & 54.5 \\
21 & Grok 4.1 Fast & 55.0 & 67.5 & 39.1 & 58.5 \\
22 & GPT-5 Nano & 54.9 & 57.9 & 47.9 & 58.8 \\
23 & Gemini 3.1 Pro & 54.8 & 61.4 & 42.9 & 60.1 \\
24 & Mistral 3 Large & 51.1 & 54.2 & 36.9 & 62.2 \\
25 & GLM 4.7 & 50.3 & 65.8 & 31.8 & 53.2 \\
26 & GPT-OSS 20B & 50.0 & 64.2 & 33.6 & 52.1 \\
27 & Qwen3 Max & 49.8 & 56.7 & 38.4 & 54.2 \\
28 & Arcee Trinity & 46.9 & 58.3 & 32.4 & 49.9 \\
29 & Qwen3 235B A22B & 45.6 & 50.8 & 35.9 & 49.9 \\
30 & Mistral 3.1 Medium & 42.7 & 49.2 & 27.4 & 51.6 \\
31 & Qwen3 Next 80b & 40.9 & 48.3 & 30.1 & 44.3 \\
32 & Llama 3.3 70B & 40.2 & 56.7 & 20.8 & 43.1 \\
\end{longtable}

\subsection{Table 3: Round Dynamics}\label{table-3-round-dynamics}

\begin{longtable}[]{@{}
  >{\raggedright\arraybackslash}p{(\columnwidth - 6\tabcolsep) * \real{0.3571}}
  >{\raggedright\arraybackslash}p{(\columnwidth - 6\tabcolsep) * \real{0.1825}}
  >{\raggedright\arraybackslash}p{(\columnwidth - 6\tabcolsep) * \real{0.1825}}
  >{\raggedright\arraybackslash}p{(\columnwidth - 6\tabcolsep) * \real{0.2619}}@{}}
\toprule\noalign{}
\begin{minipage}[b]{\linewidth}\raggedright
Metric
\end{minipage} & \begin{minipage}[b]{\linewidth}\raggedright
R1 (Outpatient)
\end{minipage} & \begin{minipage}[b]{\linewidth}\raggedright
R2 (Inpatient)
\end{minipage} & \begin{minipage}[b]{\linewidth}\raggedright
R3 (Definitive)
\end{minipage} \\
\midrule\noalign{}
\endhead
\bottomrule\noalign{}
\endlastfoot
Performance & & & \\
Model-case pairs, N & 630 & 630 & 630 \\
Score, \% (mean ± SD) & 58.0 ± 18.2 & 55.1 ± 19.4 & 51.2 ± 20.2 \\
Score, \% (median {[}IQR{]}) & 58.9 {[}45.4--71.1{]} & 55.6
{[}41.7--70.5{]} & 50.9 {[}35.8--65.6{]} \\
Score range, \% & 4.2--100.0 & 5.6--98.6 & 0.0--100.0 \\
Information requests per round (mean ± SD) & 7.5 ± 3.9 & 4.2 ± 2.2 & 1.8
± 1.0 \\
Diagnosis accuracy & & & \\
Diagnosis correct & 307 (48.7\%) & 281 (44.6\%) & 262 (41.6\%) \\
Diagnosis partially correct & 216 (34.3\%) & 307 (48.7\%) & 295
(46.8\%) \\
Diagnosis incorrect & 107 (17.0\%) & 42 (6.7\%) & 73 (11.6\%) \\
Model range & & & \\
Best model, \% (mean) & GPT-5 Mini (72.9) & GPT-5 Mini (66.9) & Claude
Sonnet 4.5 (63.0) \\
Worst model, \% (mean) & Llama 3.3 70B (28.1) & Llama 3.3 70B (27.1) &
Qwen3 Next 80B Thinking (26.3) \\
Score transitions & & & \\
Score improved from prior round & - & 297 (46.8\%) & 276 (43.5\%) \\
Score declined from prior round & - & 319 (50.3\%) & 343 (54.1\%) \\
\end{longtable}

\subsection{Table 4: Clinical Reasoning and Error Taxonomy
Evaluation}\label{table-4-clinical-reasoning-and-error-taxonomy-evaluation}

\textbf{Table 4a. R-IDEA reasoning evaluation.}

\begin{longtable}[]{@{}lrrrrrr@{}}
\toprule\noalign{}
Model & N & I & D & E & A & Total \\
\midrule\noalign{}
\endhead
\bottomrule\noalign{}
\endlastfoot
Claude Opus 4.5 & 60 & 3.63 & 1.98 & 2 & 1.97 & 9.58 \\
GLM 5 & 60 & 3.63 & 2 & 1.98 & 1.78 & 9.4 \\
Claude Sonnet 4.6 & 60 & 3.72 & 1.9 & 1.98 & 1.75 & 9.35 \\
Claude Opus 4.6 & 60 & 3.65 & 1.95 & 1.95 & 1.68 & 9.23 \\
Kimi K2 Thinking & 60 & 3.57 & 1.87 & 1.93 & 1.57 & 8.93 \\
GPT-OSS 120B & 60 & 3.17 & 1.98 & 1.98 & 1.77 & 8.9 \\
GLM 4.7 & 60 & 3.18 & 1.82 & 1.87 & 1.57 & 8.43 \\
Kimi K2.5 & 60 & 3.18 & 1.72 & 1.87 & 1.55 & 8.32 \\
DeepSeek V3.2 Speciale & 55* & 2.44 & 1.85 & 1.89 & 1.89 & 8.07 \\
MiniMax M2.5 & 56* & 2.91 & 1.73 & 1.82 & 1.59 & 8.05 \\
Claude Sonnet 4.5 & 60 & 2.92 & 1.65 & 1.87 & 1.5 & 7.93 \\
MiniMax M2.1 & 60 & 3.02 & 1.77 & 1.8 & 1.23 & 7.82 \\
Qwen3 Next 80B Thinking & 60 & 2.82 & 1.65 & 1.87 & 1.2 & 7.53 \\
Claude Haiku 4.5 & 60 & 2.93 & 1.32 & 1.83 & 1.32 & 7.4 \\
DeepSeek V3.2 & 60 & 2.05 & 1.73 & 1.72 & 1.88 & 7.38 \\
Qwen3 235B A22B & 57* & 2.81 & 1.72 & 1.09 & 1.09 & 6.7 \\
GPT-OSS 20B & 60 & 2.55 & 1.57 & 1.38 & 0.95 & 6.45 \\
\end{longtable}

\textbf{Table 4b. R-IDEA score dispersion and confidence intervals.}

\begin{longtable}[]{@{}lrrr@{}}
\toprule\noalign{}
Model & SD & CI Low & CI High \\
\midrule\noalign{}
\endhead
\bottomrule\noalign{}
\endlastfoot
Claude Opus 4.5 & 0.56 & 9.43 & 9.72 \\
GLM 5 & 0.87 & 9.18 & 9.62 \\
Claude Sonnet 4.6 & 0.99 & 9.08 & 9.58 \\
Claude Opus 4.6 & 1.36 & 8.87 & 9.55 \\
Kimi K2 Thinking & 1.41 & 8.55 & 9.27 \\
GPT-OSS 120B & 0.8 & 8.68 & 9.08 \\
GLM 4.7 & 1.61 & 8.02 & 8.83 \\
Kimi K2.5 & 1.3 & 7.98 & 8.63 \\
DeepSeek V3.2 Speciale & 1.2 & 7.73 & 8.36 \\
MiniMax M2.5 & 1.41 & 7.66 & 8.39 \\
Claude Sonnet 4.5 & 2.32 & 7.3 & 8.48 \\
MiniMax M2.1 & 1.43 & 7.45 & 8.17 \\
Qwen3 Next 80B Thinking & 1.26 & 7.22 & 7.83 \\
Claude Haiku 4.5 & 1.14 & 7.12 & 7.68 \\
DeepSeek V3.2 & 1.24 & 7.05 & 7.68 \\
Qwen3 235B A22B & 1.16 & 6.4 & 7 \\
GPT-OSS 20B & 1.43 & 6.1 & 6.8 \\
\end{longtable}

\textbf{Table 4c. Error taxonomy.}

\begin{longtable}[]{@{}lrrrrr@{}}
\toprule\noalign{}
Model & Errors & Low & Medium & High & Error (\%) \\
\midrule\noalign{}
\endhead
\bottomrule\noalign{}
\endlastfoot
Claude Opus 4.5 & 1 & 1 & 0 & 0 & 1.7 \\
GLM 5 & 1 & 1 & 0 & 0 & 1.7 \\
Claude Sonnet 4.6 & 6 & 0 & 5 & 1 & 10 \\
Claude Opus 4.6 & 2 & 1 & 1 & 0 & 3.3 \\
Kimi K2 Thinking & 12 & 3 & 6 & 3 & 18.3 \\
GPT-OSS 120B & 4 & 3 & 1 & 0 & 6.7 \\
GLM 4.7 & 2 & 1 & 1 & 0 & 3.3 \\
Kimi K2.5 & 6 & 2 & 4 & 0 & 10 \\
DeepSeek V3.2 Speciale & 3 & 0 & 3 & 0 & 5.5 \\
MiniMax M2.5 & 6 & 2 & 3 & 1 & 10.7 \\
Claude Sonnet 4.5 & 1 & 0 & 1 & 0 & 1.7 \\
MiniMax M2.1 & 3 & 1 & 2 & 0 & 5 \\
Qwen3 Next 80B Thinking & 16 & 3 & 9 & 4 & 26.7 \\
Claude Haiku 4.5 & 3 & 1 & 2 & 0 & 5 \\
DeepSeek V3.2 & 2 & 1 & 1 & 0 & 3.3 \\
Qwen3 235B A22B & 7 & 2 & 5 & 0 & 12.3 \\
GPT-OSS 20B & 18 & 5 & 12 & 1 & 25 \\
\end{longtable}

*Missing data from models not requesting new information during final
round (DeepSeek V3.2. Speciale, Qwen3 235B A22B) or failing a case
(MiniMax 2.5).

\section{Figures}\label{figures}

\clearpage

\subsection{Figure 1. Overall model performance on the OncoRounds
benchmark.}\label{figure-1.-overall-model-performance-on-the-oncorounds-benchmark.}

\begin{center}
\includegraphics[width=\linewidth,height=0.62\textheight,keepaspectratio]{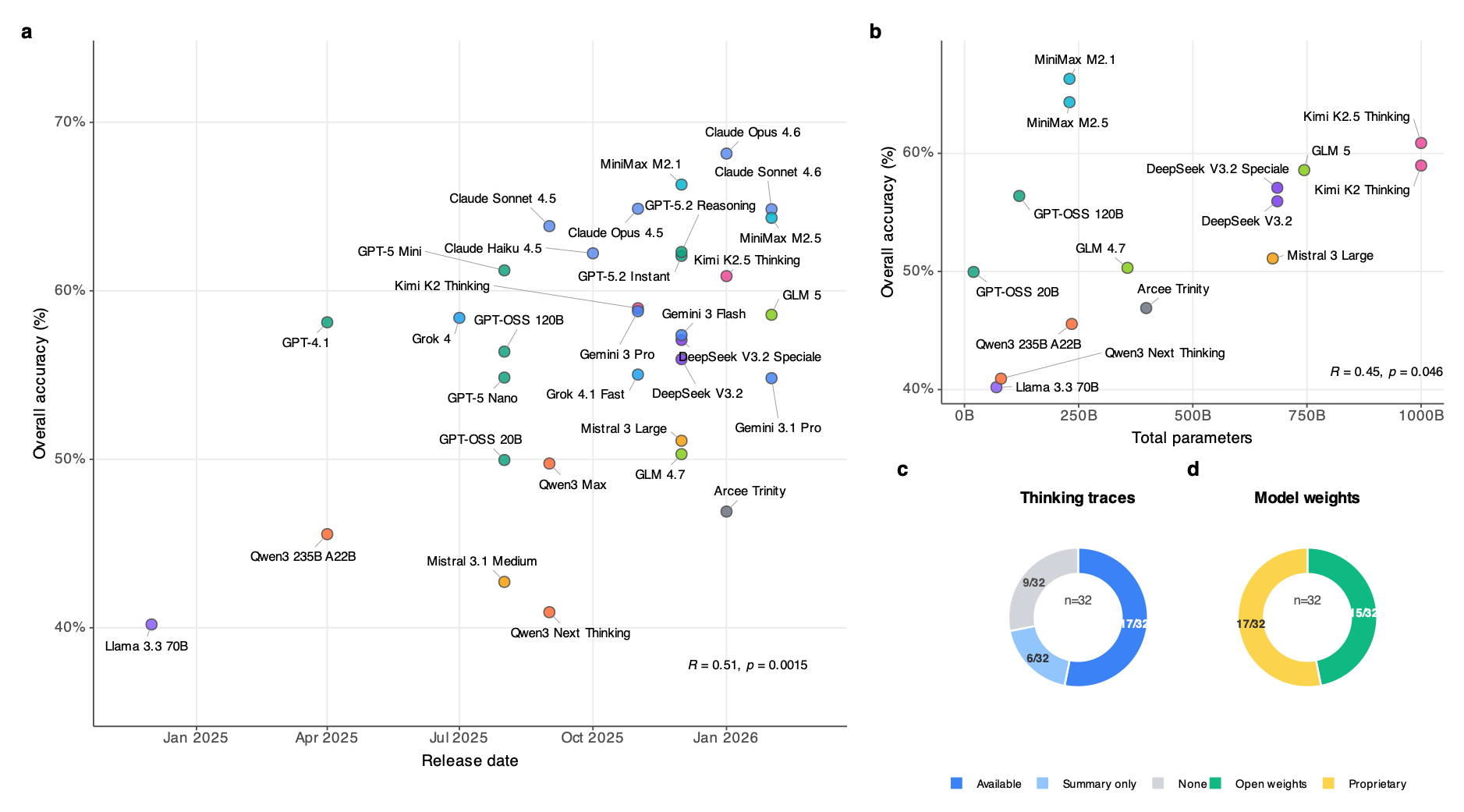}
\end{center}

\textbf{Figure 1. Overall model performance on the OncoRounds benchmark.
a}) Overall accuracy versus model release date for the 32 language
models evaluated across 20 oncology cases. Each point represents one
model, colored by model family. \textbf{b}) Overall accuracy versus
total parameter count for the 15 models with publicly known architecture
sizes. MiniMax M2.5 (230B total, 10B active parameters) outperformed all
models with 3--4 times more parameters. \textbf{c}), Distribution of
thinking trace availability across evaluated models. Of 32 models, 17
provided full reasoning traces, 6 provided reasoning summaries only, and
9 did not expose any reasoning process. \textbf{d)}, Distribution of
model weight availability. 15 models evaluated had publicly available
weights.

\clearpage

\subsection{Figure 2: Diagnostic performance across clinical
rounds.}\label{figure-2-diagnostic-performance-across-clinical-rounds.}

\begin{center}
\includegraphics[width=\linewidth,height=0.66\textheight,keepaspectratio]{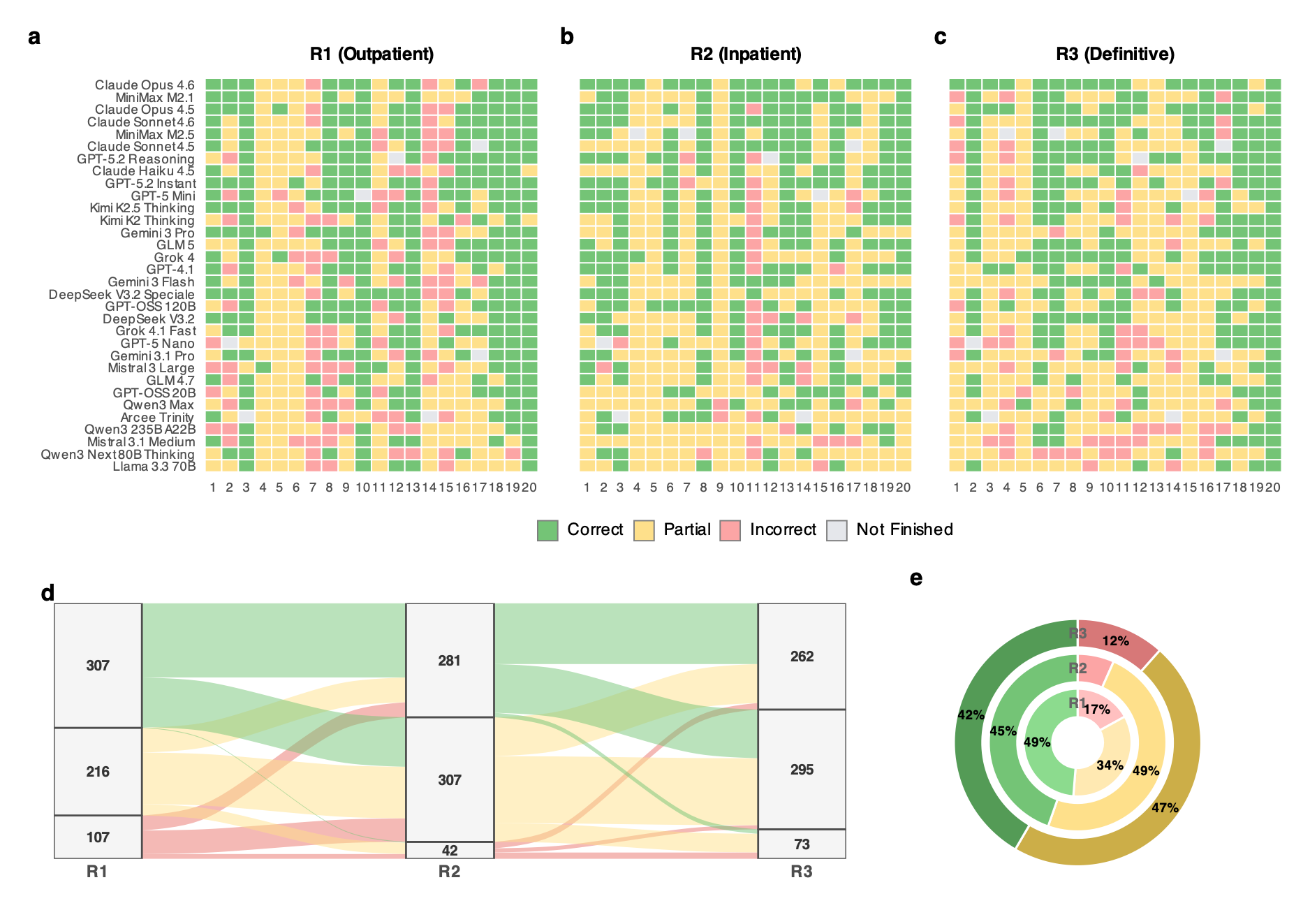}
\end{center}

\textbf{Figure 2. Diagnostic performance across clinical rounds.} a--c,
Per-round diagnosis heatmaps showing the outcome of each
model\textquotesingle s diagnostic attempt. Each tile represents one
model--case pair. d, Alluvial diagram showing diagnosis transitions
across rounds. e, Concentric donut chart summarizing the aggregate
diagnosis distribution by round (R1, inner ring; R2, middle; R3, outer).

\clearpage

\subsection{Figure 3: Clinical reasoning quality and error
taxonomy}\label{figure-3-clinical-reasoning-quality-and-error-taxonomy}

\begin{center}
\includegraphics[width=\linewidth,height=0.72\textheight,keepaspectratio]{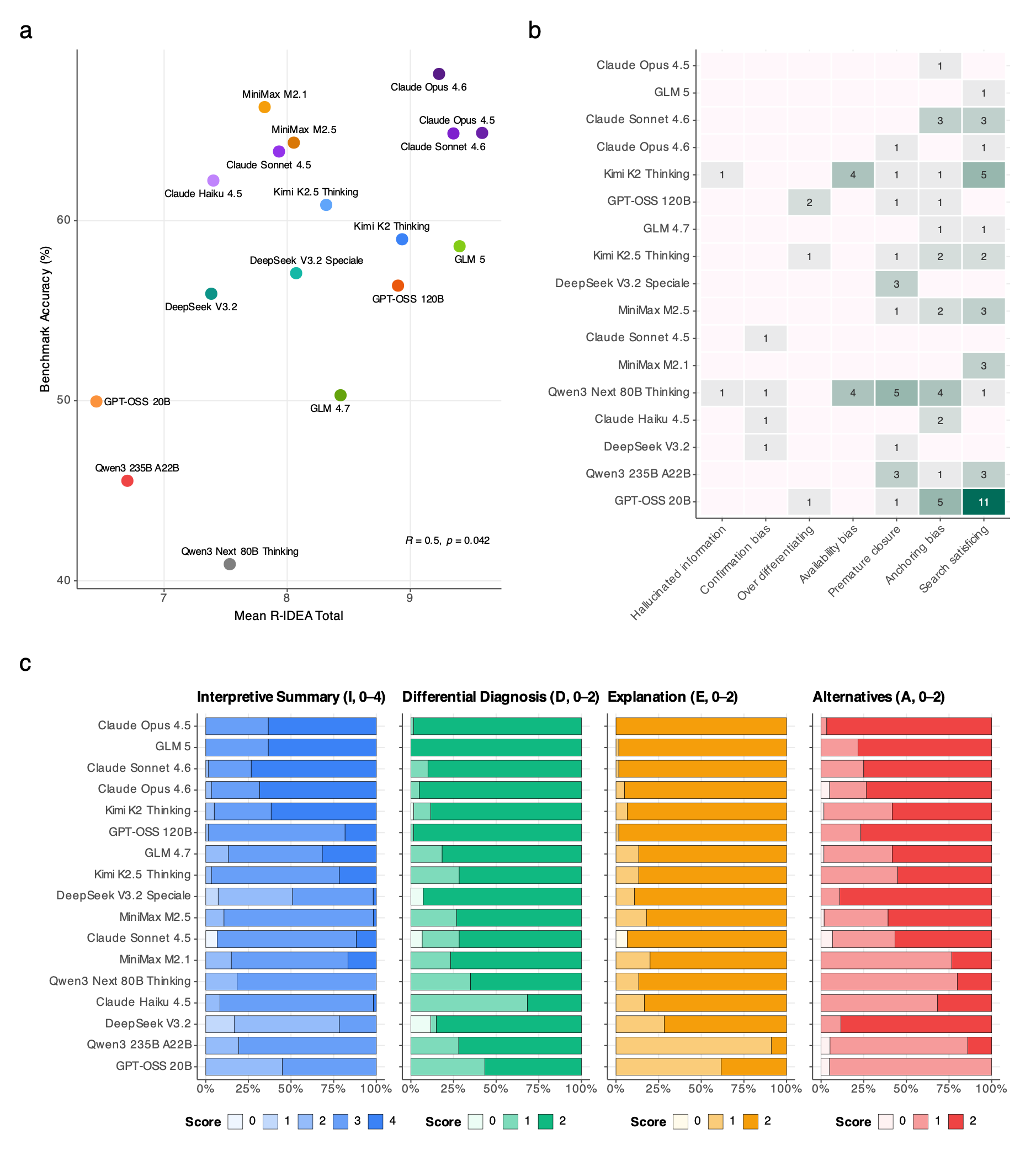}
\end{center}

\textbf{Figure 3. Clinical reasoning quality and error taxonomy.} a,
R-IDEA reasoning scores across 17 models with full thinking traces.
Scores are shown per dimension (I, interpretive summary; D, differential
diagnosis; E, explanation of lead diagnosis; A, alternatives considered)
and as total scores (0--10). Models are ordered by mean R-IDEA total. b,
Distribution of cognitive bias and process error annotations across
models. Severity is indicated by color (low, medium, high). c,
distribution of R-IDEA scores across all dimensions.

\clearpage

\subsection{Figure 4: OncoRounds benchmark
design.}\label{figure-4-oncorounds-benchmark-design.}

\begin{center}
\includegraphics[width=\linewidth,height=0.68\textheight,keepaspectratio]{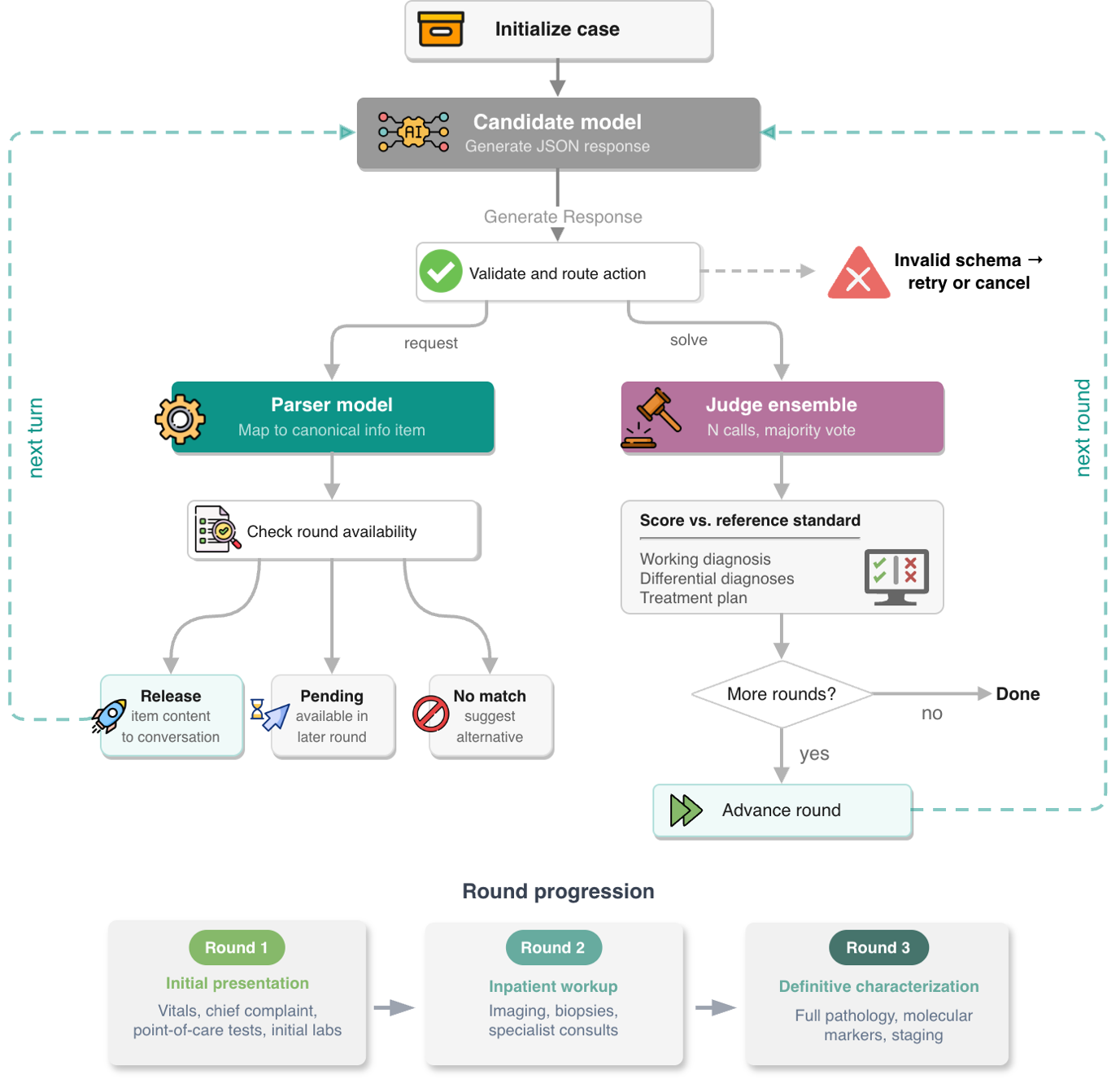}
\end{center}

\textbf{Figure 4. OncoRounds benchmark design.} Overview of the agentic
evaluation framework. Each of 20 hematologic malignancy cases is
structured across three sequential rounds (R1, outpatient presentation;
R2, inpatient workup; R3, definitive characterization). Models receive
an initial clinical scenario and actively request additional clinical
information one item per turn. After each round, models are compared
against a reference standard on diagnostic accuracy, differential
reasoning, and treatment recommendations. Cases progress from initial
syndromic impressions to molecular-level diagnoses, mirroring the
clinical workflow from admission through definitive treatment planning.

\clearpage

\section{Supplementary Material}\label{supplementary-material}

\section{Supplementary Tables}\label{supplementary-tables}

\subsection{Supplementary Table 1: Model
Overview}\label{supplementary-table-1-model-overview}

\begin{longtable}[]{@{}
  >{\raggedright\arraybackslash}p{(\columnwidth - 12\tabcolsep) * \real{0.0256}}
  >{\raggedright\arraybackslash}p{(\columnwidth - 12\tabcolsep) * \real{0.1282}}
  >{\raggedright\arraybackslash}p{(\columnwidth - 12\tabcolsep) * \real{0.2154}}
  >{\raggedright\arraybackslash}p{(\columnwidth - 12\tabcolsep) * \real{0.0718}}
  >{\raggedright\arraybackslash}p{(\columnwidth - 12\tabcolsep) * \real{0.1333}}
  >{\raggedright\arraybackslash}p{(\columnwidth - 12\tabcolsep) * \real{0.0564}}
  >{\raggedright\arraybackslash}p{(\columnwidth - 12\tabcolsep) * \real{0.3590}}@{}}
\toprule\noalign{}
\begin{minipage}[b]{\linewidth}\raggedright
\end{minipage} & \begin{minipage}[b]{\linewidth}\raggedright
Model Name
\end{minipage} & \begin{minipage}[b]{\linewidth}\raggedright
Endpoint Used
\end{minipage} & \begin{minipage}[b]{\linewidth}\raggedright
Producer
\end{minipage} & \begin{minipage}[b]{\linewidth}\raggedright
No Param / Active Param
\end{minipage} & \begin{minipage}[b]{\linewidth}\raggedright
Released
\end{minipage} & \begin{minipage}[b]{\linewidth}\raggedright
Weights Availability
\end{minipage} \\
\midrule\noalign{}
\endhead
\bottomrule\noalign{}
\endlastfoot
1 & Arcee Trinity & arcee-ai/arcee-trinity-large-preview & Arcee AI &
398B / A13B & 2026-01 &
https://huggingface.co/arcee-ai/Trinity-Large-Preview \\
2 & Claude Haiku 4.5 & anthropic/claude-haiku-4-5-20251001 & Anthropic &
Unknown & 2025-10 & Unavailable \\
3 & Claude Opus 4.5 & anthropic/claude-opus-4-5-20251101 & Anthropic &
Unknown & 2025-11 & Unavailable \\
4 & Claude Opus 4.6 & anthropic/claude-opus-4-6 & Anthropic & Unknown &
2026-01 & Unavailable \\
5 & Claude Sonnet 4.6 & anthropic/claude-sonnet-4-6 & Anthropic &
Unknown & 2026-02 & Unavailable \\
6 & Claude Sonnet 4.5 & anthropic/claude-sonnet-4-5-20250929 & Anthropic
& Unknown & 2025-09 & Unavailable \\
7 & DeepSeek V3.2 & deepseek/deepseek-v3.2 & DeepSeek & 685B / A37B &
2025-12 & https://huggingface.co/deepseek-ai/DeepSeek-V3.2 \\
8 & DeepSeek V3.2 Speciale & deepseek/deepseek-v3.2-speciale & DeepSeek
& 685B / A37B & 2025-12 &
https://huggingface.co/deepseek-ai/DeepSeek-V3.2-Speciale \\
9 & Gemini 3 Flash & google/gemini-3-flash-preview & Google & Unknown &
2025-12 & Unavailable \\
10 & Gemini 3.1 Pro & google/gemini-3.1-pro-preview & Google & Unknown &
2026-02 & Unavailable \\
11 & Gemini 3 Pro & google/gemini-3-pro-preview & Google & Unknown &
2025-11 & Unavailable \\
12 & GLM 4.7 & zhipu/glm-4.7 & Zhipu AI & 357B / 32B & 2025-12 &
https://huggingface.co/zai-org/GLM-4.7 \\
13 & GLM 5 & zhipu/glm-5 & Zhipu AI & 744B / 40B & 2026-02 &
https://huggingface.co/zai-org/GLM-5 \\
14 & GPT-4.1 & openai/gpt-4.1-2025-04-14 & OpenAI & Unknown & 2025-04 &
Unavailable \\
15 & GPT-5 Mini & openai/gpt-5-2025-08-07 & OpenAI & Unknown & 2025-08 &
Unavailable \\
16 & GPT-5 Nano & openai/gpt-5-nano-2025-08-07 & OpenAI & Unknown &
2025-08 & Unavailable \\
17 & GPT-5.2 Instant & openai/gpt-5.2-2025-12-11 & OpenAI & Unknown &
2025-12 & Unavailable \\
18 & GPT-5.2 Reasoning & openai/gpt-5.2-2025-12-11 & OpenAI & Unknown &
2025-12 & Unavailable \\
19 & GPT-OSS 120B & openai/gpt-oss-120b & OpenAI & 120B & 2025-08 &
https://huggingface.co/openai/gpt-oss-120b \\
20 & GPT-OSS 20B & openai/gpt-oss-20b & OpenAI & 20B & 2025-08 &
https://huggingface.co/openai/gpt-oss-20b \\
21 & Grok 4 & x-ai/grok-4-0709 & xAI & Unknown & 2025-07 &
Unavailable \\
22 & Grok 4.1 Fast & x-ai/grok-4-1-fast-non-reasoning & xAI & Unknown &
2025-11 & Unavailable \\
23 & Kimi K2 Thinking & moonshot/kimi-k2 & Moonshot AI & 1T / A32B &
2025-11 & https://huggingface.co/moonshotai/Kimi-K2-thinking \\
24 & Kimi K2.5 & moonshot/kimi-2.5 & Moonshot AI & 1T / A32B & 2026-01 &
https://huggingface.co/moonshotai/Kimi-K2.5 \\
25 & Llama 3.3 70B & meta-llama/llama-3.3-70b-instruct-turbo & Meta &
70B & 2024-12 &
https://huggingface.co/meta-llama/Llama-3.3-70B-Instruct \\
26 & MiniMax M2.1 & minimax/minimax-m2p1 & MiniMax & 230B / A10B &
2025-12 & https://huggingface.co/MiniMaxAI/MiniMax-M2.1 \\
27 & MiniMax M2.5 & minimax/minimax-m2.5 & MiniMax & 230B / A10B &
2026-02 & https://huggingface.co/MiniMaxAI/MiniMax-M2.5 \\
28 & Mistral 3 Large & mistralai/mistral-large-2512 & Mistral AI & 675B
/ 41B & 2025-12 &
https://huggingface.co/mistralai/Mistral-Large-3-675B-Instruct-2512 \\
29 & Mistral 3.1 Medium & mistralai/mistral-medium-3.1 & Mistral AI &
Unknown & 2025-08 & Unavailable \\
30 & Qwen3 235B A22B & qwen/qwen3-vl-235b-a22b & Alibaba & 235B / A22B &
2025-04 & https://huggingface.co/Qwen/Qwen3-235B-A22B-Instruct \\
31 & Qwen3 Max & qwen/qwen3-max & Alibaba & \textasciitilde1T (MoE) &
2025-09 & Unavailable \\
32 & Qwen3 Next 80b & qwen/qwen3-next-80b-a3b & Alibaba & 80B / A3B &
2025-09 & https://huggingface.co/Qwen/Qwen3-Next-80B-A3B-Thinking \\
\end{longtable}

For MoE models, parameters are listed as Total / Active (e.g.~685B /
A37B). "Unknown" indicates closed-weight models with undisclosed
parameter counts.

\subsection{Supplementary Table 2: Diagnostic trajectory analysis across
clinical
rounds}\label{supplementary-table-2-diagnostic-trajectory-analysis-across-clinical-rounds}

\begin{longtable}[]{@{}
  >{\raggedright\arraybackslash}p{(\columnwidth - 14\tabcolsep) * \real{0.1712}}
  >{\raggedright\arraybackslash}p{(\columnwidth - 14\tabcolsep) * \real{0.0342}}
  >{\raggedright\arraybackslash}p{(\columnwidth - 14\tabcolsep) * \real{0.1370}}
  >{\raggedright\arraybackslash}p{(\columnwidth - 14\tabcolsep) * \real{0.0822}}
  >{\raggedright\arraybackslash}p{(\columnwidth - 14\tabcolsep) * \real{0.1027}}
  >{\raggedright\arraybackslash}p{(\columnwidth - 14\tabcolsep) * \real{0.1507}}
  >{\raggedright\arraybackslash}p{(\columnwidth - 14\tabcolsep) * \real{0.1370}}
  >{\raggedright\arraybackslash}p{(\columnwidth - 14\tabcolsep) * \real{0.1712}}@{}}
\toprule\noalign{}
\begin{minipage}[b]{\linewidth}\raggedright
Model
\end{minipage} & \begin{minipage}[b]{\linewidth}\raggedright
N
\end{minipage} & \begin{minipage}[b]{\linewidth}\raggedright
Sustained correct
\end{minipage} & \begin{minipage}[b]{\linewidth}\raggedright
Recovered
\end{minipage} & \begin{minipage}[b]{\linewidth}\raggedright
Deteriorated
\end{minipage} & \begin{minipage}[b]{\linewidth}\raggedright
Sustained incorrect
\end{minipage} & \begin{minipage}[b]{\linewidth}\raggedright
Recovery rate (\%)
\end{minipage} & \begin{minipage}[b]{\linewidth}\raggedright
Deterioration rate (\%)
\end{minipage} \\
\midrule\noalign{}
\endhead
\bottomrule\noalign{}
\endlastfoot
Claude Opus 4.6 & 20 & 9 & 7 & 3 & 1 & 87.5 & 25.0 \\
MiniMax M2.1 & 20 & 4 & 5 & 6 & 5 & 50.0 & 60.0 \\
Claude Opus 4.5 & 20 & 9 & 6 & 5 & 0 & 100.0 & 35.7 \\
Claude Sonnet 4.6 & 20 & 7 & 5 & 5 & 3 & 62.5 & 41.7 \\
MiniMax M2.5 & 18 & 6 & 3 & 5 & 4 & 42.9 & 45.5 \\
Claude Sonnet 4.5 & 19 & 8 & 2 & 3 & 6 & 25.0 & 27.3 \\
GPT-5.2 Reasoning & 19 & 9 & 4 & 2 & 4 & 50.0 & 18.2 \\
Claude Haiku 4.5 & 20 & 3 & 5 & 6 & 6 & 45.5 & 66.7 \\
GPT-5.2 Instant & 20 & 9 & 2 & 6 & 3 & 40.0 & 40.0 \\
GPT-5 Mini & 18 & 3 & 3 & 7 & 5 & 37.5 & 70.0 \\
Kimi K2.5 & 20 & 4 & 4 & 8 & 4 & 50.0 & 66.7 \\
Kimi K2 Thinking & 20 & 2 & 5 & 5 & 8 & 38.5 & 71.4 \\
Gemini 3 Pro & 20 & 5 & 2 & 8 & 5 & 28.6 & 61.5 \\
GLM 5 & 20 & 4 & 4 & 6 & 6 & 40.0 & 60.0 \\
Grok 4 & 20 & 8 & 6 & 3 & 3 & 66.7 & 27.3 \\
GPT-4.1 & 20 & 6 & 5 & 4 & 5 & 50.0 & 40.0 \\
Gemini 3 Flash & 20 & 4 & 3 & 4 & 9 & 25.0 & 50.0 \\
DeepSeek V3.2 Speciale & 20 & 5 & 4 & 7 & 4 & 50.0 & 58.3 \\
GPT-OSS 120B & 20 & 5 & 4 & 6 & 5 & 44.4 & 54.5 \\
DeepSeek V3.2 & 20 & 7 & 3 & 5 & 5 & 37.5 & 41.7 \\
Grok 4.1 Fast & 20 & 6 & 2 & 4 & 8 & 20.0 & 40.0 \\
GPT-5 Nano & 19 & 4 & 3 & 3 & 9 & 25.0 & 42.9 \\
Gemini 3.1 Pro & 19 & 4 & 0 & 6 & 9 & 0.0 & 60.0 \\
Mistral 3 Large & 20 & 2 & 3 & 5 & 10 & 23.1 & 71.4 \\
GLM 4.7 & 20 & 2 & 5 & 7 & 6 & 45.5 & 77.8 \\
GPT-OSS 20B & 20 & 4 & 1 & 5 & 10 & 9.1 & 55.6 \\
Qwen3 Max & 20 & 3 & 2 & 3 & 12 & 14.3 & 50.0 \\
Arcee Trinity & 18 & 2 & 3 & 4 & 9 & 25.0 & 66.7 \\
Qwen3 235B A22B & 20 & 2 & 2 & 2 & 14 & 12.5 & 50.0 \\
Mistral 3.1 Medium & 20 & 2 & 4 & 5 & 9 & 30.8 & 71.4 \\
Qwen3 Next 80b & 20 & 1 & 1 & 5 & 13 & 7.1 & 83.3 \\
Llama 3.3 70B & 20 & 2 & 3 & 3 & 12 & 20.0 & 60.0 \\
\end{longtable}

Per-model classification of diagnostic outcomes between rounds.
Trajectories are defined as: sustained correct (correct in both R1 and
R3), recovered (incorrect in R1, correct in R3), deteriorated (correct
in R1, incorrect in R3), or sustained incorrect (incorrect in both R1
and R3). Recovery rate denotes the proportion of initially incorrect
diagnoses corrected by R3; deterioration rate denotes the proportion of
initially correct diagnoses lost by R3. Partially correct diagnoses were
counted as incorrect for trajectory classification. Models are ordered
by overall composite accuracy. N, number of model--case pairs with
complete R1 and R3 data; values below 20 indicate cases where the model
failed to complete the benchmark case.

\subsection{Supplementary Table 3: Mean information requests per
round.}\label{supplementary-table-3-mean-information-requests-per-round.}

\begin{longtable}[]{@{}
  >{\raggedright\arraybackslash}p{(\columnwidth - 8\tabcolsep) * \real{0.3784}}
  >{\raggedright\arraybackslash}p{(\columnwidth - 8\tabcolsep) * \real{0.1486}}
  >{\raggedright\arraybackslash}p{(\columnwidth - 8\tabcolsep) * \real{0.1486}}
  >{\raggedright\arraybackslash}p{(\columnwidth - 8\tabcolsep) * \real{0.1486}}
  >{\raggedright\arraybackslash}p{(\columnwidth - 8\tabcolsep) * \real{0.1486}}@{}}
\toprule\noalign{}
\begin{minipage}[b]{\linewidth}\raggedright
Model
\end{minipage} & \begin{minipage}[b]{\linewidth}\raggedright
R1
\end{minipage} & \begin{minipage}[b]{\linewidth}\raggedright
R2
\end{minipage} & \begin{minipage}[b]{\linewidth}\raggedright
R3
\end{minipage} & \begin{minipage}[b]{\linewidth}\raggedright
Overall
\end{minipage} \\
\midrule\noalign{}
\endhead
\bottomrule\noalign{}
\endlastfoot
Claude Opus 4.6 & 9.4 & 6.3 & 2.4 & 6.0 \\
MiniMax M2.1 & 10.3 & 5.3 & 1.8 & 5.8 \\
Claude Opus 4.5 & 12.0 & 7.9 & 3.0 & 7.6 \\
Claude Sonnet 4.6 & 6.7 & 4.3 & 2.0 & 4.3 \\
MiniMax M2.5 & 10.5 & 5.6 & 2.0 & 6.2 \\
Claude Sonnet 4.5 & 10.9 & 5.5 & 3.1 & 6.5 \\
GPT-5.2 Reasoning & 7.9 & 4.3 & 2.3 & 4.8 \\
Claude Haiku 4.5 & 8.1 & 5.2 & 1.7 & 5.0 \\
GPT-5.2 Instant & 10.0 & 3.3 & 1.9 & 5.1 \\
GPT-5 Mini & 8.0 & 4.8 & 1.0 & 4.6 \\
Kimi K2.5 & 9.9 & 5.5 & 2.5 & 6.0 \\
Kimi K2 Thinking & 4.2 & 3.1 & 1.7 & 3.0 \\
Gemini 3 Pro & 8.4 & 4.3 & 1.7 & 4.8 \\
GLM 5 & 8.0 & 4.2 & 2.2 & 4.8 \\
Grok 4 & 8.5 & 5.2 & 3.4 & 5.7 \\
GPT-4.1 & 8.4 & 3.9 & 1.7 & 4.7 \\
Gemini 3 Flash & 8.8 & 4.5 & 1.7 & 5.0 \\
DeepSeek V3.2 Speciale & 5.5 & 4.0 & 2.4 & 4.0 \\
GPT-OSS 120B & 10.3 & 3.8 & 2.2 & 5.5 \\
DeepSeek V3.2 & 6.8 & 3.3 & 1.8 & 3.9 \\
Grok 4.1 Fast & 4.2 & 3.7 & 1.9 & 3.2 \\
GPT-5 Nano & 5.5 & 2.4 & 1.0 & 2.9 \\
Gemini 3.1 Pro & 6.7 & 3.5 & 1.4 & 3.8 \\
Mistral 3 Large & 2.9 & 2.6 & 1.4 & 2.3 \\
GLM 4.7 & 9.0 & 5.3 & 2.4 & 5.6 \\
GPT-OSS 20B & 6.6 & 2.8 & 1.4 & 3.6 \\
Qwen3 Max & 4.3 & 2.7 & 1.5 & 2.8 \\
Arcee Trinity & 7.9 & 4.4 & 1.5 & 4.6 \\
Qwen3 235B A22B & 5.2 & 2.3 & 1.1 & 2.9 \\
Mistral 3.1 Medium & 2.6 & 2.1 & 1.1 & 1.9 \\
Qwen3 Next 80b & 6.5 & 3.7 & 1.1 & 3.8 \\
Llama 3.3 70B & 7.5 & 3.7 & 1.2 & 4.1 \\
\end{longtable}

\subsection{Supplementary Table 4: Thinking Trace Statistics per
Model}\label{supplementary-table-4-thinking-trace-statistics-per-model}

\begin{longtable}[]{@{}
  >{\raggedright\arraybackslash}p{(\columnwidth - 10\tabcolsep) * \real{0.2427}}
  >{\raggedright\arraybackslash}p{(\columnwidth - 10\tabcolsep) * \real{0.0971}}
  >{\raggedright\arraybackslash}p{(\columnwidth - 10\tabcolsep) * \real{0.1359}}
  >{\raggedright\arraybackslash}p{(\columnwidth - 10\tabcolsep) * \real{0.1748}}
  >{\raggedright\arraybackslash}p{(\columnwidth - 10\tabcolsep) * \real{0.1553}}
  >{\raggedright\arraybackslash}p{(\columnwidth - 10\tabcolsep) * \real{0.1748}}@{}}
\toprule\noalign{}
\begin{minipage}[b]{\linewidth}\raggedright
Model
\end{minipage} & \begin{minipage}[b]{\linewidth}\raggedright
N Steps
\end{minipage} & \begin{minipage}[b]{\linewidth}\raggedright
Total Words
\end{minipage} & \begin{minipage}[b]{\linewidth}\raggedright
Mean Words/Step
\end{minipage} & \begin{minipage}[b]{\linewidth}\raggedright
SD Words/Step
\end{minipage} & \begin{minipage}[b]{\linewidth}\raggedright
Mean Chars/Step
\end{minipage} \\
\midrule\noalign{}
\endhead
\bottomrule\noalign{}
\endlastfoot
Qwen3 Next 80b & 227 & 622,655 & 2,743.0 & 1,406.7 & 18,069.4 \\
DeepSeek V3.2 Speciale & 210 & 384,179 & 1,829.4 & 1,521.5 & 12,614.0 \\
GLM 4.7 & 350 & 465,054 & 1,328.7 & 448.4 & 9,365.9 \\
Kimi K2 Thinking & 182 & 150,441 & 826.6 & 695.6 & 5,592.4 \\
Qwen3 235B A22B & 163 & 128,939 & 791.0 & 519.7 & 5,087.9 \\
DeepSeek V3.2 & 236 & 186,159 & 788.8 & 505.5 & 5,459.8 \\
Kimi K2.5 & 409 & 199,987 & 489.0 & 500.2 & 3,330.9 \\
Claude Haiku 4.5 & 301 & 120,496 & 400.3 & 171.2 & 2,898.6 \\
GPT-OSS 20B & 216 & 79,959 & 370.2 & 276.3 & 2,553.4 \\
GLM 5 & 307 & 101,452 & 330.5 & 223.5 & 2,284.1 \\
Claude Sonnet 4.5 & 409 & 123,495 & 301.9 & 114.3 & 2,157.1 \\
Claude Opus 4.5 & 457 & 120,838 & 264.4 & 116.1 & 1,901.8 \\
MiniMax M2.1 & 350 & 90,775 & 259.4 & 126.1 & 1,864.7 \\
MiniMax M2.5 & 357 & 79,132 & 221.7 & 94.2 & 1,544.7 \\
Claude Sonnet 4.6 & 266 & 58,690 & 220.6 & 255.7 & 1,579.2 \\
Claude Opus 4.6 & 323 & 57,624 & 178.4 & 140.7 & 1,296.2 \\
GPT-OSS 120B & 327 & 58,123 & 177.7 & 134.2 & 1,291.0 \\
\end{longtable}

\clearpage
\begin{landscape}

\subsection{Supplementary Table 5: Per-Case Deterioration
Summary}\label{supplementary-table-5-per-case-deterioration-summary}

\begin{longtable}[]{@{}
  >{\raggedright\arraybackslash}p{(\columnwidth - 14\tabcolsep) * \real{0.0400}}
  >{\raggedright\arraybackslash}p{(\columnwidth - 14\tabcolsep) * \real{0.0900}}
  >{\raggedright\arraybackslash}p{(\columnwidth - 14\tabcolsep) * \real{0.0600}}
  >{\raggedright\arraybackslash}p{(\columnwidth - 14\tabcolsep) * \real{0.0600}}
  >{\raggedright\arraybackslash}p{(\columnwidth - 14\tabcolsep) * \real{0.0750}}
  >{\raggedright\arraybackslash}p{(\columnwidth - 14\tabcolsep) * \real{0.0700}}
  >{\raggedright\arraybackslash}p{(\columnwidth - 14\tabcolsep) * \real{0.2850}}
  >{\raggedright\arraybackslash}p{(\columnwidth - 14\tabcolsep) * \real{0.3200}}@{}}
\toprule\noalign{}
\begin{minipage}[b]{\linewidth}\raggedright
Case
\end{minipage} & \begin{minipage}[b]{\linewidth}\raggedright
Cancer type
\end{minipage} & \begin{minipage}[b]{\linewidth}\raggedright
R1 correct
\end{minipage} & \begin{minipage}[b]{\linewidth}\raggedright
Sust. correct
\end{minipage} & \begin{minipage}[b]{\linewidth}\raggedright
Deteriorated
\end{minipage} & \begin{minipage}[b]{\linewidth}\raggedright
Det. rate (\%)
\end{minipage} & \begin{minipage}[b]{\linewidth}\raggedright
R1 reference
\end{minipage} & \begin{minipage}[b]{\linewidth}\raggedright
R3 reference
\end{minipage} \\
\midrule\noalign{}
\endhead
\bottomrule\noalign{}
\endlastfoot
3 & AML & 29 & 3 & 26 & 89.7 & Myelodysplastic syndrome (MDS) &
Relapsed/refractory AML with myelodysplasia-related changes and complex
karyotype, ELN 2022 risk: Adverse \\
13 & Multiple Myeloma & 29 & 3 & 26 & 89.7 & Multiple Myeloma in VGPR
with ongoing treatment & Plasma cell myeloma (WHO 2022), R-ISS stage
III; High Risk IgA Multiple Myeloma with prior Plasma Cell Leukemia
transformation; Triple-class refractory; Cytopenia likely
drug-associated with concurrent CHIP \\
1 & AML & 19 & 1 & 18 & 94.7 & Acute leukemia; Recurrent pulmonary
embolism & Relapsed AML with other defined genetic alterations (WHO
2022); Pulmonary embolism; Right ventricular heart failure \\
8 & AML & 22 & 10 & 12 & 54.5 & Acute leukemia & Relapsed AML-MR (WHO
2022) after alloSCT \\
10 & ALL & 30 & 18 & 12 & 40.0 & T-cell acute lymphoblastic leukemia
(T-ALL), post-induction & Early T-cell precursor acute lymphoblastic
leukemia, molecular failure \\
12 & ALCL & 17 & 6 & 11 & 64.7 & Progression or relapse of lymphoma &
Peripheral T-cell lymphoma, NOS (WHO 2022), relapsed after
haploidentical alloSCT \\
9 & Multiple Myeloma & 17 & 8 & 9 & 52.9 & Relapsed multiple myeloma
with extramedullary manifestation & Multiple myeloma IgG kappa with
FISH-defined high-risk cytogenetics (R-ISS II) and MDS-pCT with TP53
mutation \\
19 & AML & 29 & 20 & 9 & 31.0 & Acute leukemia with hyperleukocytosis,
most likely AML pCT & AML pCT with TP53 mutation, ELN 2024 adverse
risk \\
16 & Follicular Lymphoma & 16 & 8 & 8 & 50.0 & Lymphoma with high tumor
burden, tumor lysis syndrome, and infection of unclear origin & Relapsed
classical follicular lymphoma grade 1-2, Ann Arbor stage IV, with early
post-CAR-T relapse, TP53 deletion, and high-risk FLIPI \\
17 & ALL & 15 & 8 & 7 & 46.7 & Acute leukemia in remission post
allogeneic HCT with suspected sinusitis & Molecular relapse of
B-lymphoblastic leukemia/lymphoma NOS (WHO 2022) with antigen-loss
(CD19-/CD22-) and phenotypic switch toward AUL-like profile \\
18 & Aplastic Anemia & 25 & 19 & 6 & 24.0 & Transfusion-dependent
aplastic anemia with PNH clone and residual disseminated fusariosis &
vSAA not in remission after immunosuppressive therapy with ATG; MDS with
monosomy 7 (WHO 2022) \\
15 & Multiple Myeloma & 5 & 0 & 5 & 100.0 & Cytopenia and new vertebral
compression fractures in a heavily pretreated multiple myeloma patient
in apparent biochemical remission, raising concern for therapy-related
bone marrow failure or evolving secondary myeloid neoplasm. &
Relapsed/refractory multiple myeloma (rrMM) and myelodysplastic syndrome
(MDS) with TP53 mutation, post-chemotherapy (pCT), WHO 2022; IPSS-M 2.92
(very high risk), IPSS-R 8.0 (very high risk); not eligible for
allogeneic transplantation due to age and comorbidities (HCT-CI 6-7
points). \\
4 & MDS & 2 & 0 & 2 & 100.0 & Acute febrile hypoxemic respiratory
illness in an immunocompromised post-alloSCT patient with chronic
pulmonary and cutaneous GvHD, highly suspicious for viral and/or
invasive fungal pneumonia with concurrent cytopenias requiring further
evaluation for possible disease relapse. & Relapsed myelodysplastic
syndrome with TP53 mutation (WHO 2022) after alloSCT, with clonal
evolution on a background of complex cytogenetics and persistent mixed
chimerism, complicated by severe steroid-refractory chronic GvHD of skin
and joints, and ongoing fungal pneumonia. \\
7 & DLBCL & 6 & 4 & 2 & 33.3 & Relapsed (hematologic) malignancy with
mass on the left ankle & High-grade B-cell lymphoma with MYC and BCL2
rearrangements (WHO 2022); Isolated extranodal cutaneous involvement;
HCT-CI score 5 \\
2 & AML & 11 & 10 & 1 & 9.1 & Relapse or secondary malignancy in
post-alloSCT patient & Post-transplant extramedullary relapse of
NPM1-mutated AML \\
5 & AML & 2 & 1 & 1 & 50.0 & AML post cytotoxic therapy (AML-pCT) &
AML-pCT (WHO 2022) after treatment of multiple myeloma with tandem ASCT
and lenalidomide maintenance; ELN 2022 adverse risk (17p deletion,
complex karyotype, TP53 mutation); refractory to
azacitidine/venetoclax \\
11 & DLBCL & 2 & 1 & 1 & 50.0 & Relapsed Hodgkin Lymphoma & High-grade
B-cell lymphoma with MYC and BCL2 rearrangements (WHO 2022). Ann Arbor
Stage IV. \\
6 & BPDCN & 1 & 1 & 0 & 0.0 & Unclear lymphadenopathy,
hepatosplenomegaly and nodular skin lesions & Relapsed BPDCN after
alloSCT with skin, lymph node, hepatosplenic, pericardial, pleural, and
bone marrow involvement, harboring TET2, ASXL1, and TP53 mutations \\
14 & AML & 0 & 0 & 0 & - & Suspected MDS/AML relapse after alloSCT with
concurrent pulmonary infection & AML with mutated TP53 (WHO 2022),
ELN2022 adverse risk \\
20 & CLL/ Richter & 30 & 30 & 0 & 0.0 & Suspected CLL with Richter
transformation & Chronic lymphocytic leukemia with Richter
transformation to diffuse large B-cell lymphoma, Ann Arbor stage IV \\
\end{longtable}

\clearpage
\subsection{Supplementary Table 6: Representative Diagnosis Text for
High-Deterioration
Cases.}\label{supplementary-table-6-representative-diagnosis-text-for-high-deterioration-cases.}

\begin{longtable}[]{@{}
  >{\raggedright\arraybackslash}p{(\columnwidth - 10\tabcolsep) * \real{0.0450}}
  >{\raggedright\arraybackslash}p{(\columnwidth - 10\tabcolsep) * \real{0.1000}}
  >{\raggedright\arraybackslash}p{(\columnwidth - 10\tabcolsep) * \real{0.2900}}
  >{\raggedright\arraybackslash}p{(\columnwidth - 10\tabcolsep) * \real{0.3000}}
  >{\raggedright\arraybackslash}p{(\columnwidth - 10\tabcolsep) * \real{0.0750}}
  >{\raggedright\arraybackslash}p{(\columnwidth - 10\tabcolsep) * \real{0.1900}}@{}}
\toprule\noalign{}
\begin{minipage}[b]{\linewidth}\raggedright
\end{minipage} & \begin{minipage}[b]{\linewidth}\raggedright
Source
\end{minipage} & \begin{minipage}[b]{\linewidth}\raggedright
R1 diagnosis
\end{minipage} & \begin{minipage}[b]{\linewidth}\raggedright
R3 diagnosis
\end{minipage} & \begin{minipage}[b]{\linewidth}\raggedright
R3 Score
\end{minipage} & \begin{minipage}[b]{\linewidth}\raggedright
Explanation
\end{minipage} \\
\midrule\noalign{}
\endhead
\bottomrule\noalign{}
\endlastfoot
Case 3 & & & & & \\
& Reference & Myelodysplastic syndrome (MDS) & Relapsed/refractory AML
with myelodysplasia-related changes and complex karyotype, ELN 2022
risk: Adverse & & \\
& Kimi K2 Thinking & High-risk myelodysplastic syndrome (MDS) with
transfusion-dependent pancytopenia & Acute Myeloid Leukemia with
Myelodysplasia-Related Changes (AML-MR), therapy-related type, with
high-risk molecular features (ASXL1, RUNX1, TET2 mutations) and
suspected germline DDX41 predisposition syndrome & partial & Correct
AML-MR reclassification but invoked DDX41 germline predisposition not in
reference; missed complex karyotype \\
& Claude Haiku 4.5 & Myelodysplastic syndrome (MDS) with cytopenias &
Therapy-related acute myeloid leukemia (tAML) with complex karyotype,
high-risk cytogenetics, and myelodysplasia-related changes & partial &
Identified therapy-related AML with MDS changes but used tAML label
instead of WHO 2022 AML-MR; missed ELN adverse risk \\
& GLM 5 & Myelodysplastic Syndrome (MDS) with multilineage cytopenias &
Acute Myeloid Leukemia (AML) with Myelodysplasia-Related Features and
DDX41 Mutation & partial & Correct AML with MDS features but incomplete:
no complex karyotype, no ELN risk, no relapsed/refractory context \\
Case 1 & & & & & \\
& Reference & Acute leukemia; Recurrent pulmonary embolism & Relapsed
AML with other defined genetic alterations (WHO 2022); Pulmonary
embolism; Right ventricular heart failure & & \\
& Arcee Trinity & Acute myeloid leukemia with monocytic differentiation
(M4/M5 AML) with hyperleukocytosis and tumor lysis syndrome & Acute
myeloid leukemia with monocytic differentiation (M4/M5 AML) with
FLT3-ITD mutation and hyperleukocytosis & partial & Retained FAB M4/M5
instead of WHO 2022; missed relapsed status and pulmonary embolism
context \\
& Gemini 3 Flash & Acute myeloid leukemia with monocytic differentiation
& Relapsed Acute Myeloid Leukemia with FLT3-ITD mutation & partial &
Correct relapse and FLT3-ITD but incomplete: no WHO 2022 subtype, no PE,
no RV failure \\
& Grok 4 & Acute myeloid leukemia (likely monocytic subtype) with
leukostasis and tumor lysis syndrome & Acute myeloid leukemia with
monocytic differentiation (AML-M5) and FLT3-ITD mutation, complicated by
pulmonary embolism and possible disseminated intravascular coagulation.
& partial & Correct disease and mutation; used FAB (AML-M5) instead of
WHO 2022; added DIC not in reference \\
Case 10 & & & & & \\
& Reference & T-cell acute lymphoblastic leukemia (T-ALL),
post-induction & Early T-cell precursor acute lymphoblastic leukemia,
molecular failure & & \\
& DeepSeek V3.2 Speciale & Post-induction T-ALL in remission with
residual hepatic and hematologic abnormalities, and complications of
sinus vein thrombosis and intracerebral hemorrhage & T-ALL in
morphological remission but with detectable MRD positive (0.1\%) after
consolidation, with NRAS mutation and specific TCR/IGH rearrangements.
History of sinus vein thrombosis and ICH. Considered high-risk for
relapse. & partial & Identified MRD positivity and NRAS but missed
ETP-ALL subtype; stated remission instead of molecular failure \\
& Mistral 3 Large & Post-induction remission of T-ALL with persistent
cytopenias, coagulopathy (likely secondary to recent thrombosis and
hemorrhage), and mild transaminitis. Minimal residual disease (MRD)
status remains critical but is not yet provided. & Minimal residual
disease (MRD)-positive T-ALL post-induction therapy with NRAS mutation,
indicating high risk of relapse despite clinical remission. Persistent
cytopenias and coagulopathy secondary to prior therapy and thrombotic
complications. & partial & Recognized MRD-positive status but missed
ETP-ALL subtype; framed as high relapse risk rather than molecular
failure \\
& Qwen3 235B A22B & T-ALL in remission post-induction with complicated
sinus venous thrombosis (CVST) and intracerebral hemorrhage (ICH) &
T-ALL (ETP-ALL subtype) in morphologic remission post-induction with
high-risk disease requiring allogeneic hematopoietic stem cell
transplant (HSCT). & partial & Correctly identified ETP-ALL but stated
morphologic remission, contradicting molecular failure per reference \\
Case 8 & & & & & \\
& Reference & Acute leukemia & Relapsed AML-MR (WHO 2022) after alloSCT
& & \\
& Arcee Trinity & Acute Myeloid Leukemia with sepsis and multi-organ
dysfunction & Acute Myeloid Leukemia with complex karyotype and multiple
mutations (ASXL1, DNMT3A, SRSF2, STAG2, RAD21) & partial & Identified
AML with complex karyotype and mutations but missed relapsed status,
MDS-related changes, and alloSCT context \\
& GPT-4.1 & Likely acute leukemia (with blast crisis) presenting with
neutropenic sepsis and multi-organ dysfunction & Acute myeloid leukemia
(AML) with myelodysplasia-related changes, adverse molecular profile
(ASXL1, DNMT3A, SRSF2, STAG2, RAD21 mutations), relapsed/refractory
disease with ongoing neutropenic sepsis & partial & Recognized AML-MR
and adverse molecular profile but added neutropenic sepsis framing not
in reference diagnosis \\
& Qwen3 Next 80B Thinking & Acute Myeloid Leukemia with septic shock
secondary to probable pneumonia in neutropenic patient & Myelodysplastic
Syndrome with excess blasts type 2 (MDS-EB2) with septic shock and
disseminated intravascular coagulation (DIC) & incorrect & Misclassified
as MDS-EB2 instead of AML-MR; incorrect disease entity \\
\end{longtable}

For each case, the reference standard diagnosis is shown alongside
model-generated diagnoses at Round 1 (R1) and Round 3 (R3), the assigned
R3 score, and a post-hoc explanation of the scoring rationale.
Deterioration reflects cases where models achieved a correct R1
diagnosis but declined to partially correct or incorrect at R3,
typically because they failed to integrate new information disclosed in
later rounds. Common failure patterns include retention of outdated
classification systems (e.g., FAB instead of WHO 2022), omission of
prognostically essential features (e.g., ELN adverse risk, triple-class
refractory designation), and mischaracterization of disease status
(e.g., "morphologic remission" when molecular data indicated treatment
failure). Post-hoc explanations were added manually by the authors.

\end{landscape}
\clearpage

\section{Supplementary Figures}\label{supplementary-figures}

\subsection{Supplementary Figure 1: Information request
patterns.}\label{supplementary-figure-1-information-request-patterns.}

\begin{center}
\includegraphics[width=0.88\linewidth,height=0.70\textheight,keepaspectratio]{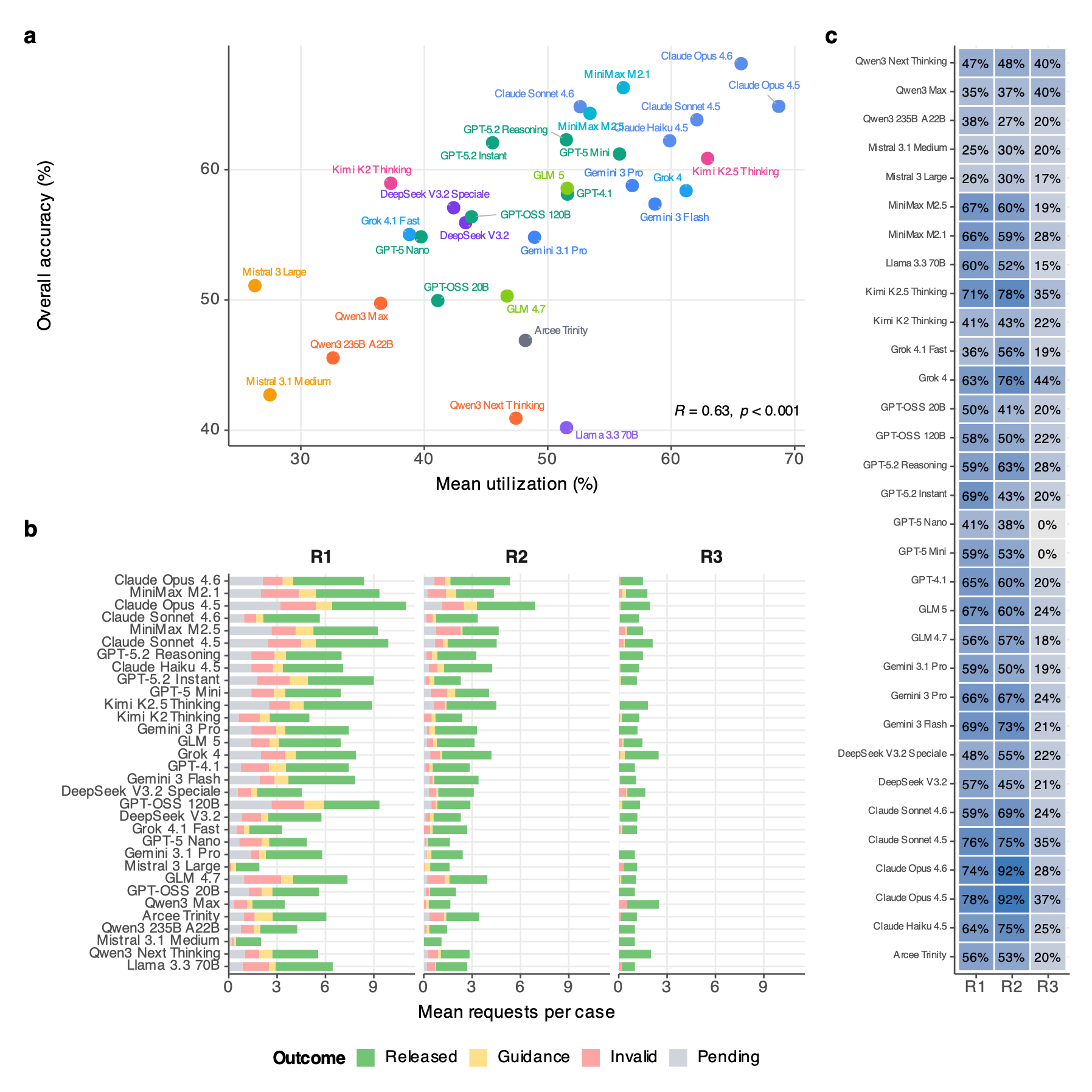}
\end{center}

\textbf{Supplementary Figure 1. Information request patterns.} a,
Distribution of information requests per round across all model--case
pairs showing a positive correlation between quantity of requested
information and overall accuracy. b, Request volume declined from R1
through R3. c, Per-round utilization rates showing the proportion of
available clinical data requested by models in each round.

\clearpage

\subsection{Supplementary Figure 2: Per-model round dynamics and
trajectory
analysis}\label{supplementary-figure-2-per-model-round-dynamics-and-trajectory-analysis}

\begin{center}
\includegraphics[width=\linewidth,height=0.78\textheight,keepaspectratio]{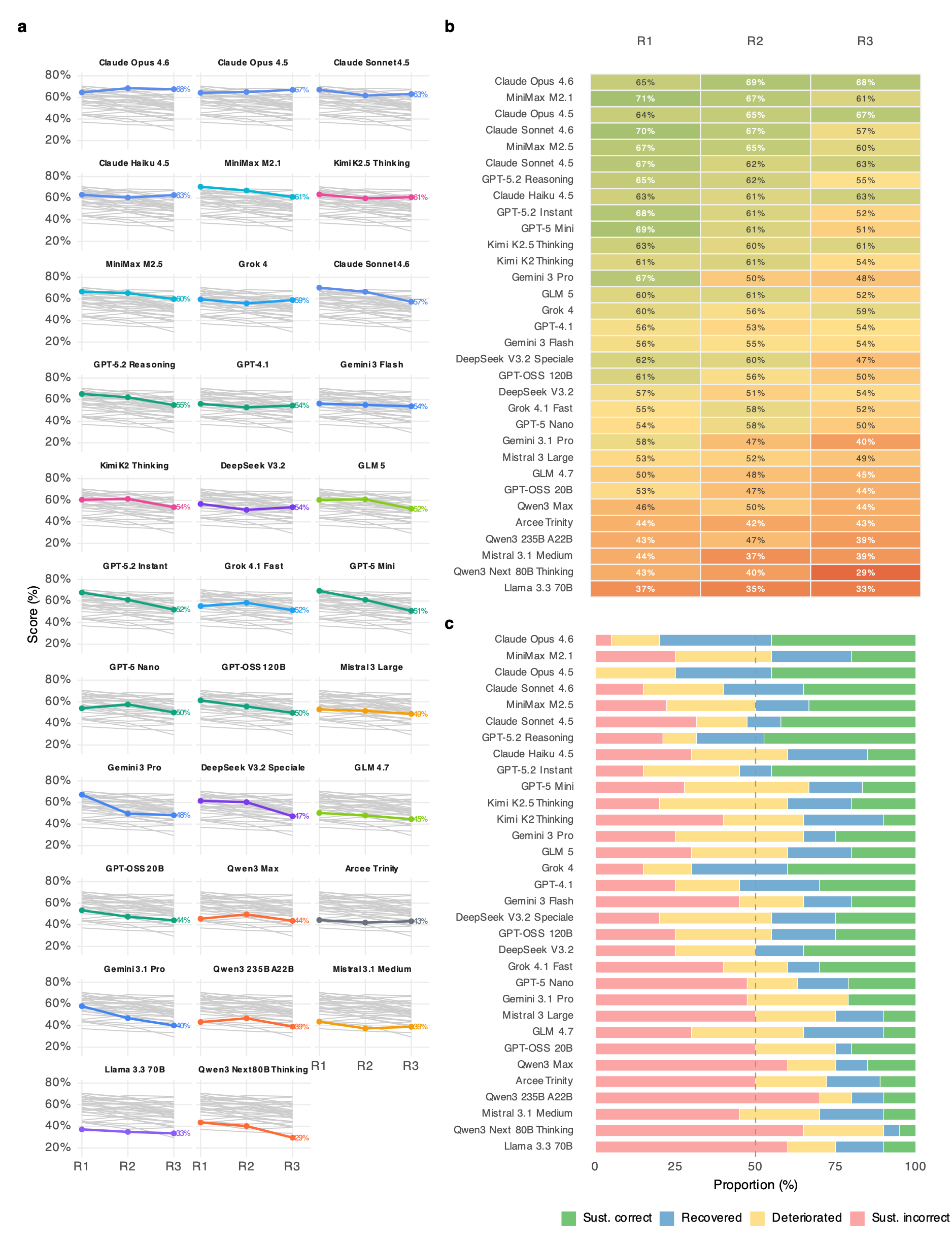}
\end{center}

\textbf{Supplementary Figure 2. Per-model round dynamics and trajectory
analysis.} a, Score trajectories for each of the 32 models across Rounds
1--3. End-point labels indicate the R3 mean score. b, Mean composite
score (\%) per model and round. c, Per-model diagnostic trajectory
distribution. Each bar shows the proportion of cases classified as
sustained correct (correct in both R1 and R3), recovered (incorrect in
R1, correct in R3), deteriorated (correct in R1, incorrect in R3), or
sustained incorrect (incorrect in both R1 and R3).

\clearpage

\subsection{Supplementary Figure 3: R-IDEA scores and thinking trace
length}\label{supplementary-figure-3-r-idea-scores-and-thinking-trace-length}

\begin{center}
\includegraphics[width=\linewidth,height=0.66\textheight,keepaspectratio]{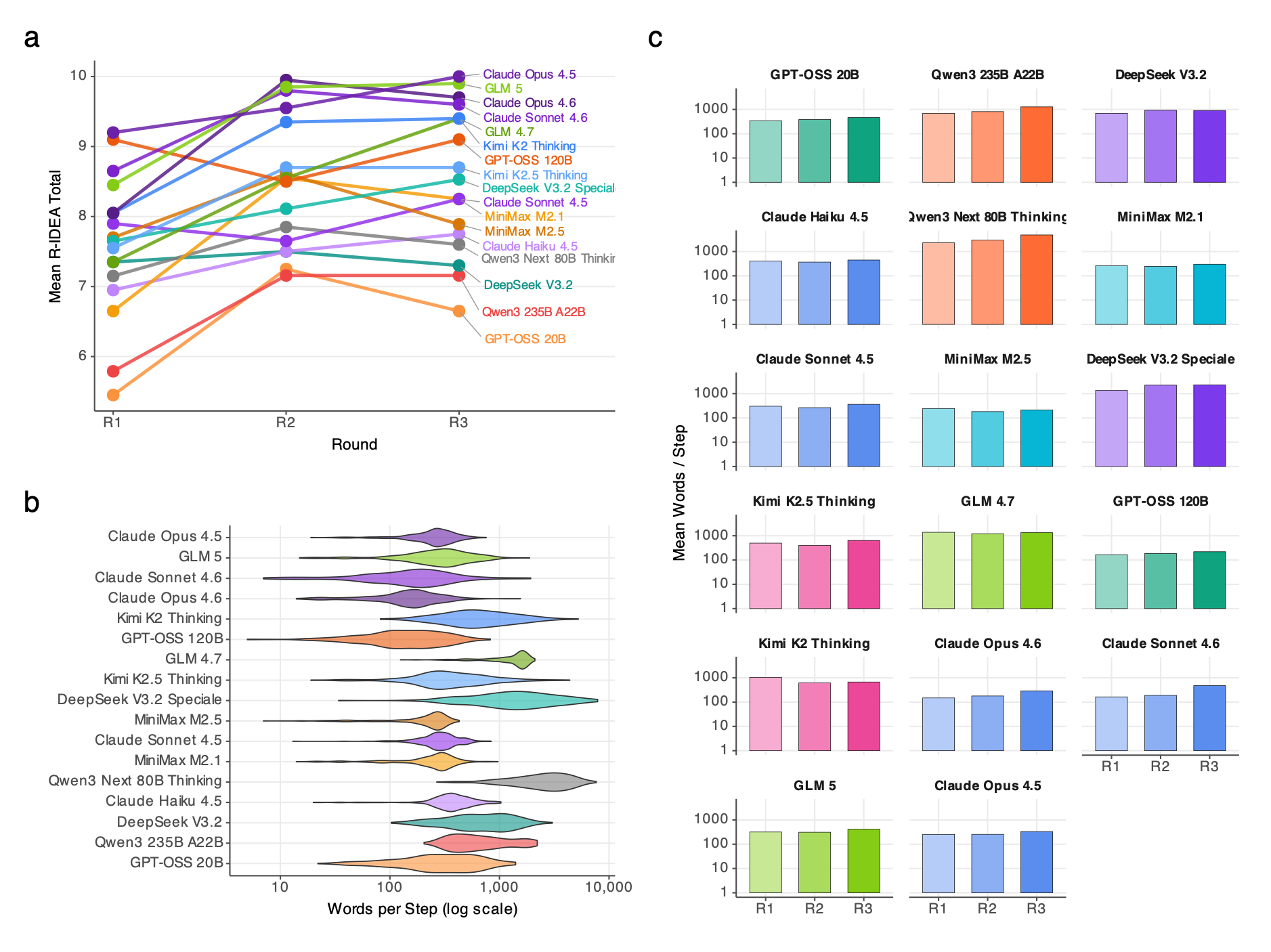}
\end{center}

\textbf{Supplementary Figure 3. R-IDEA scores and thinking trace
length.} a, Per-model R-IDEA scores by round, b, thinking trace length
(word count), c) and mean words per step. R-IDEA scores slightly
increased in later rounds for most models, paired with longer reasoning
traces.

\clearpage

\subsection{Supplementary Figure 4: Error severity
distribution.}\label{supplementary-figure-4-error-severity-distribution.}

\begin{center}
\includegraphics[width=\linewidth,height=0.48\textheight,keepaspectratio]{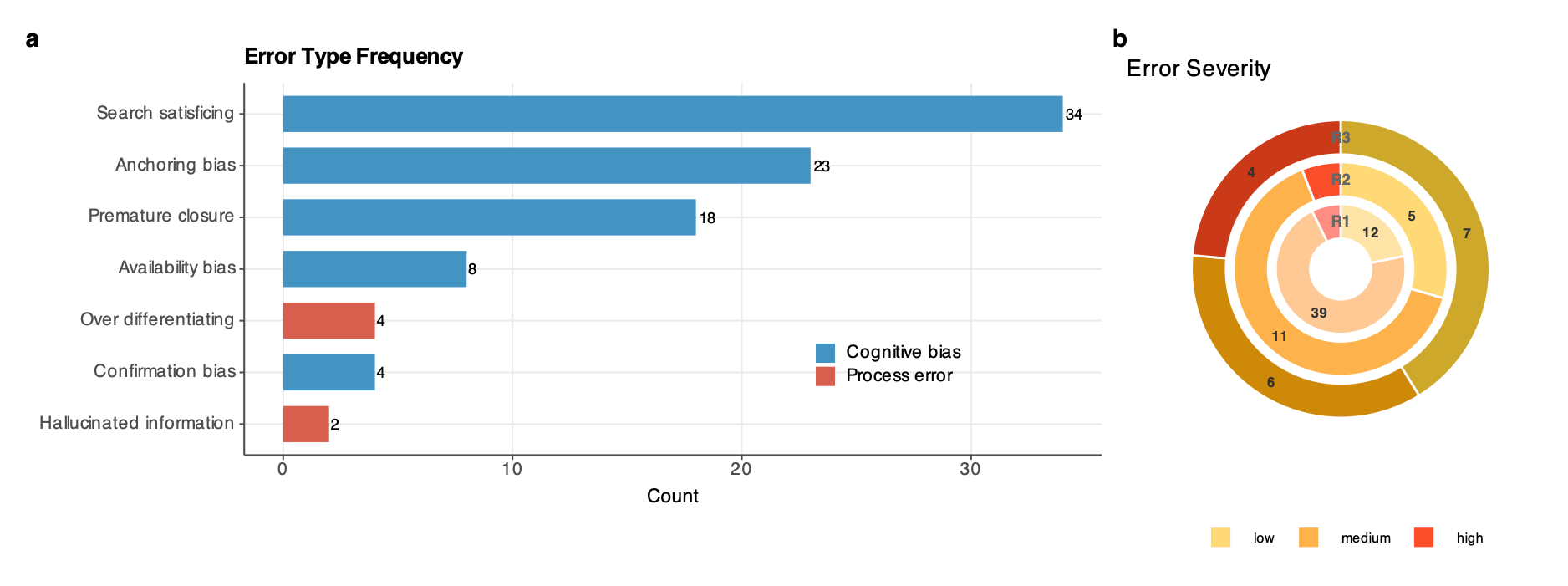}
\end{center}

\textbf{Supplementary Figure 4. Error severity distribution.} a,
Distribution of error severity (low, medium, high) across all annotated
reasoning errors. b, No critical errors with the potential to cause
direct patient harm were identified.

\clearpage

\subsection{Supplementary Figure 5: Deterioration Patterns in Key
Cases}\label{supplementary-figure-5-deterioration-patterns-in-key-cases}

\begin{center}
\includegraphics[width=\linewidth,height=0.68\textheight,keepaspectratio]{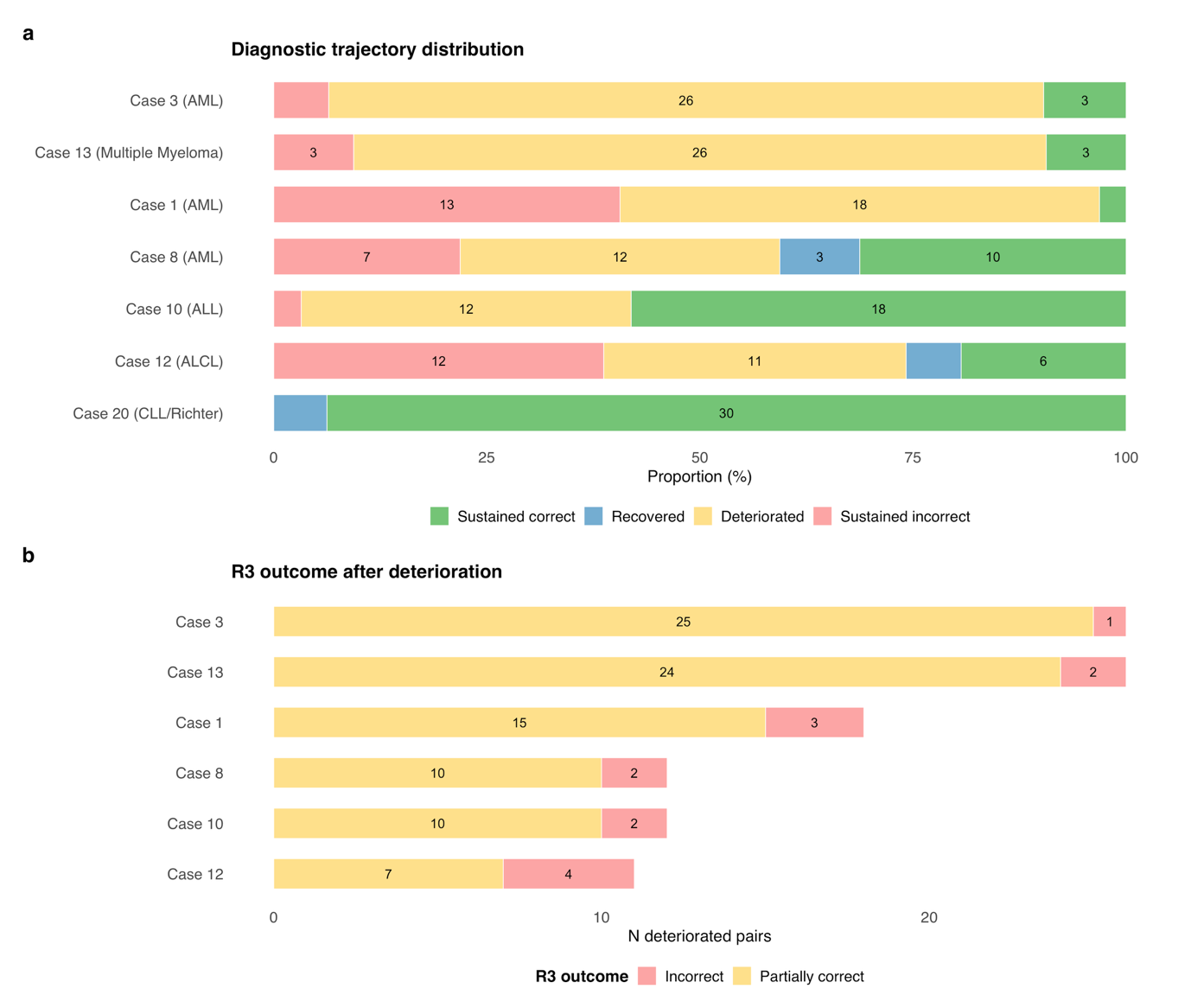}
\end{center}

\textbf{Supplementary Figure 5: Deterioration Patterns in Key Cases.} a,
Diagnostic trajectory distribution for selected cases with the highest
deterioration rates. Each bar shows the proportion of model-case pairs
classified as sustained correct, recovered, deteriorated, or sustained
incorrect between R1 and R3. Case 20 (Richter transformation) showed
near-universal diagnostic stability, while Cases 3 and 13 had the
highest deterioration rates (89.7\% each). b, R3 outcome after
deterioration, showing whether deteriorated diagnoses were scored as
partially correct or incorrect at R3. Most deterioration resulted in
partially correct rather than fully incorrect R3 diagnoses, indicating
that models incorporated some later-round findings but failed to achieve
the molecular precision required for full credit.

\clearpage

\section{Supplementary Appendix: Reasoning Trace
Excerpts}\label{supplementary-appendix-reasoning-trace-excerpts}

\subsection{Introduction}\label{introduction-1}

This supplementary appendix presents 49 curated reasoning trace excerpts
from 17 large language models, programmatically selected from 1,066
R-IDEA evaluation entries. Each excerpt includes the clinical context,
the evaluator's assessment, and a verbatim passage from the model's
internal reasoning trace.

Excerpts are organized into three sections:

\begin{itemize}
\item
  Error Type Taxonomy (26 excerpts), illustrating all 7 cognitive error
  types identified during evaluation
\item
  R-IDEA Dimension Analysis (16 excerpts), contrasting low and high
  performance on each of the four R-IDEA dimensions
\item
  Exemplary Reasoning (7 excerpts), showcasing high-fidelity clinical
  reasoning (total R-IDEA \(\geq\) 9, no errors)
\end{itemize}

\textbf{Selection methodology.} Excerpt selection and annotation were
performed with the assistance of Claude Opus 4.6 (Anthropic). A Python
script scored candidates by evaluator quote availability, error
severity, explanation quality, and model diversity, then extracted
verbatim passages from the corresponding thinking trace files. For error
type excerpts, candidates were scored by evaluator quote availability,
error severity, explanation quality, and model diversity. For R-IDEA
dimensions, the two lowest- and highest-scoring entries per dimension
were selected with family-level diversity constraints. Exemplary
excerpts required perfect or near-perfect scores (total \(\geq\) 9) with
no identified errors, diversified across cases and model families. All
excerpts were reviewed by the authors for accuracy and
representativeness.

\subsection{Error Type Taxonomy}\label{error-type-taxonomy}

This section presents excerpts illustrating the cognitive error types
identified during R-IDEA evaluation, organized into two categories
following Croskerry's dual-process taxonomy. Three error types captured
by the benchmark's scoring mechanism (incomplete workup, missed key
findings, incorrect diagnosis) are excluded here, consistent with the
main error taxonomy analysis.

\subsubsection{Cognitive Biases}\label{cognitive-biases}

Systematic deviations in the reasoning process.

\subsubsection{Anchoring Bias}\label{anchoring-bias}

5 excerpts from 3 models.

\textbf{Case 7} \textbar{} DLBCL \textbar{} Round 1 \textbar{} Claude
Sonnet 4.6

Severity: high

\textbf{Evaluator:} Weak R1 reasoning. The model correctly identifies
immunosuppression from the medication profile but incorrectly assumes
HIV as the cause rather than post-chemotherapy or post-transplant
prophylaxis. This leads to anchoring on Kaposi sarcoma throughout R1,
missing the more likely diagnosis of relapsed lymphoma. The dark brown
nodular lesion on the ankle is noted but misinterpreted. The
pancytopenia and hyponatremia are integrated but attributed to HIV
rather than lymphoma.

\textbf{Error detail:} Model assumes HIV from co-trimoxazole and
valaciclovir, then anchors on Kaposi sarcoma throughout R1. The patient
is actually HIV-negative; these medications were post-transplant
prophylaxis for a relapsed lymphoma. The model does not consider that
co-trimoxazole and valaciclovir are also standard post-chemotherapy or
post-transplant prophylaxis.

Thinking trace (step 0):

Let me analyze this case: - 70-year-old male - New nodular lesion on the
left ankle with swelling and discoloration - Vital signs are relatively
stable

This presentation is concerning for several possibilities: 1. Kaposi's
sarcoma - classic presentation with violaceous nodular lesions on lower
extremities, especially in elderly Mediterranean/Eastern European men
(classic KS) or HIV-positive patients 2. Melanoma 3. Other soft tissue
sarcoma 4. Vascular tumor (angiosarcoma) 5. Stasis dermatitis/venous
insufficiency with secondary changes 6. Lymphoma (cutaneous) 7. Merkel
cell carcinoma. Given his age and the ankle location with discoloration,
Kaposi's sarcoma remains high on my differential, particularly the
classic form in elderly men of Mediterranean or Eastern European
descent. I need to take a focused history and perform a careful physical
exam to better characterize this lesion.

\textbf{Case 4} \textbar{} MDS \textbar{} Round 3 \textbar{} Qwen3 Next
80B Thinking

Severity: high

\textbf{Evaluator:} Persistent anchoring on infectious complications
across all three rounds. Despite being in the definitive
characterization phase with molecular data available, the model never
identified the relapsed MDS with TP53 mutation. The infection management
reasoning was sound, but the fundamental diagnostic failure represents a
critical gap.

\textbf{Error detail:} Through all three rounds, the model remained
anchored on the infectious diagnosis and never identified the relapsed
MDS with TP53 mutation, which was the primary hematologic diagnosis.
This represents persistent failure to consider the underlying
malignancy.

Thinking trace (step 12):

{[}\ldots{]} Then the most critical therapies: antibiotics and
antifungal.

But for the antifungal, voriconazole loading dose is needed.

Also, checking whether ceftazidime-avibactam is correct; some guidelines
prefer meropenem for OXA-48, but it's resistant to carbapenems. So
avibactam is needed.

Ceftazidime-avibactam is correct choice for OXA-48 producing Klebsiella.

Now, the treatment plan actions should be:

\begin{itemize}
\item
  Supportive oxygen
\item
  Antibiotics correct
\item
  Antifungal correct
\item
  Antiviral correct
\end{itemize}

Taper steroids or reduce immunosuppression for GVHD

Diagnostic for HLH

In the previous attempt, HLH was a differential but the treatment plan
didn't address it. So perhaps adding the bone marrow biopsy to check for
HLH.

But HLH is usually confirmed by hemophagocytes in marrow, and other
criteria, but bone marrow biopsy can be done.

However, the patient has cytopenias, so bone marrow biopsy might be
risky, but needed.

So this seems logical.

So final solve could be:

Working diagnosis: Invasive Pulmonary Aspergillosis with ESBL-producing
Klebsiella pneumoniae and Influenza A co-infection in chronic
Graft-versus-Host Disease

Differentials: Hemophagocytic Lymphohistiocytosis, Organizing Pneumonia
related to chronic GVHD, Mucormycosis

Treatment plan:

Oxygen support: supportive, priority 1

IV ceftazidime-avibactam: therapeutic, priority 2

Voriconazole: therapeutic, priority 2

Oseltamivir: therapeutic, priority 3

Taper prednisolone gradually: therapeutic, priority 3

Bone marrow biopsy to assess hemophagocytosis: diagnostic, priority 2

\textbf{Case 6} \textbar{} BPDCN \textbar{} Round 2 \textbar{} Qwen3
Next 80B Thinking

Severity: high

\textbf{Evaluator:} Failed to update diagnosis despite additional
pathologic data being available. Continued to anchor on PTLD without
considering BPDCN, which is the hardest case across all models in this
benchmark. The differential did not include any plasmacytoid dendritic
cell entities.

\textbf{Error detail:} Despite moving to R2 with biopsy/flow cytometry
capabilities, the model maintained PTLD without incorporating new
pathologic data that would distinguish BPDCN from PTLD. BPDCN has
distinctive markers (CD4+, CD56+, CD123+, BDCA-2+) that differ from
typical PTLD.

Thinking trace (step 8):

{[}\ldots{]} So the user should request bone marrow biopsy.

Alternatively, the skin biopsy has already been requested in the
previous solve, but maybe the user needs to confirm with a specific
request for it.

Wait, but in the previous solve, the user included skin biopsy as part
of the solution. But according to the problem instructions, you must
request information one piece at a time until you are confident.
However, when solving, you provide the treatment plan. But in the
current round 2, the patient is now inpatient, so perhaps the user can
proceed to solve with the necessary tests.

But the previous solve was evaluated as partially correct, so maybe the
working diagnosis is correct but differentials or treatment plan
incorrect.

Looking at the evaluation: working\_diagnosis was partially\_correct.
The differentials had one correct and others incorrect. For the
treatment plan, first item correct, second priority partially correct,
others incorrect.

So in this round, to proceed correctly, perhaps:

\begin{itemize}
\item
  Working diagnosis: PTLD (correct)
\item
  Differentials: High-grade B-cell lymphoma (which is a possible PTLD
  subtype), maybe other lymphomas, but chronic GVHD might not be a major
  differential for nodules; maybe another lymphoma, or EBV-driven. Maybe
  also consider lymphomatoid papulosis or other skin lymphomas.
  Sarcoidosis and multiple myeloma might not fit.
\end{itemize}

So perhaps differentials should include other lymphomas (B-cell or
T-cell), perhaps cutaneous lymphoma, but PTLD is the main one.

But in post-transplant, PTLD is a key diagnosis.

Treatment plan: Skin biopsy is correct for diagnostic. PET-CT for
staging is good. Hypercalcemia treatment: yes, but maybe the action is
correct. Intravenous fluids and zoledronic acid. However, zoledronic
acid is used for hypercalcemia of malignancy, but if the hypercalcemia
is due to PTLD, then yes. {[}\ldots{]}

\textbf{Case 2} \textbar{} AML \textbar{} Round 1 \textbar{} Claude
Sonnet 4.6

Severity: medium

\textbf{Evaluator:} Good R1 reasoning with strong medication inference
(correctly identifies post-allo-HSCT status from ruxolitinib +
prophylaxis regimen). Integrates cardiac biomarkers with skin findings
and retroperitoneal masses into a multi-system picture. However, anchors
on PTLD rather than considering extramedullary AML relapse as the
leading diagnosis. The normal CBC with no cytopenias is correctly noted.
Information gathering strategy is reasonable with echo and abdominal US.

\textbf{Error detail:} Model anchors on PTLD as leading diagnosis rather
than considering extramedullary AML relapse, which is more likely given
the post-HSCT context and normal blood counts. The medication profile
(ruxolitinib for GvHD, posaconazole) suggests prior AML treatment
followed by allo-SCT, making relapse the more important consideration.

Thinking trace (step 8):

{[}\ldots{]} \textbf{CBC:} Normal values across the board - hemoglobin
14.2, WBC 6.8, platelets 240 - no cytopenias evident.

\textbf{Chemistry:} The labs show cardiac involvement with elevated
NT-proBNP at 1250 pg/mL and mildly elevated troponin T at 0.024 ng/mL
suggesting cardiac stress. She's tachycardic at 116, liver function is
normal, LDH is normal at 180, but creatinine is mildly elevated at 1.2.

\textbf{Echo:} Showing pericardial tissue thickening with septal
hypertrophy and an unclear mass lesion.

\textbf{Abdominal US:} Multiple rounded lesions measuring 2-3 cm
adjacent to the left ureter causing grade II hydronephrosis, with normal
liver and spleen.

The clinical picture here is striking - this post-HSCT patient on
ruxolitinib presents with new skin lesions, multi-organ involvement
including cardiac and pericardial changes, retroperitoneal masses
obstructing the ureter, and neurological symptoms. This constellation
strongly suggests post-transplant lymphoproliferative disorder, likely
EBV-associated PTLD with monomorphic features. The multi-site disease
across skin, heart, and retroperitoneum in an immunosuppressed
transplant patient is the classic presentation.

My differential includes relapse of the underlying malignancy, secondary
lymphoma, or disseminated infection, but PTLD sits at the top given the
immunosuppressive context and distribution pattern. I need to move
forward with urgent tissue diagnosis, EBV viral load testing, reduction
of immunosuppression, and staging imaging, along with specialist
consultations for the cardiac and renal involvement.

\textbf{Case 4} \textbar{} MDS \textbar{} Round 1 \textbar{} GPT-OSS 20B

Severity: medium

\textbf{Evaluator:} R1 reasoning correctly identifies the infectious
component (IPA suspicion based on halo sign) but misses the broader
clinical picture of a post-alloSCT patient where disease relapse should
be high on the differential. The model spends 8 steps primarily focused
on respiratory infection workup without requesting information about the
underlying hematologic condition. The reference expects concurrent
cytopenias requiring evaluation for possible disease relapse, which the
model does not adequately address.

\textbf{Error detail:} The model anchors heavily on invasive pulmonary
aspergillosis based on the halo sign, to the exclusion of considering
MDS relapse as the underlying driver. The reference standard emphasizes
that the acute illness occurs in a post-alloSCT patient with chronic
GvHD and possible disease relapse, but the model focuses almost entirely
on the infectious etiology without adequately considering the
hematologic dimension.

Thinking trace (step 0):

We have a case: 56-year-old female, post-transplant (immunocompromised),
now fever, dry cough, progressive fatigue, worsening joint pain and
stiffness. Vital signs: low BP, tachycardic, febrile, tachypneic, low O2
sat 87\%. She is in acute respiratory symptoms. We need to request a
single piece of information. Likely we want a chest imaging: chest X-ray
or CT. Also want labs: CBC with differential. She has joint pain, could
be infection or autoimmune. She is immunocompromised; could have
opportunistic infections: PCP pneumonia, CMV, fungal infections
(Aspergillus). Also could be viral infection like COVID-19 or influenza.
Also drug-related toxicity? We need more info.

Given that the scenario allows us to request routine blood evaluations.
The most immediate is CBC with differential. This gives WBC, anemia,
thrombocytopenia, maybe neutropenia. Also might do blood cultures. But
the immediate next step is a chest X-ray to assess cause of hypoxia. But
also we need to differentiate PCP vs bacterial pneumonia. In
immunocompromised, PCP is common, presenting as dry cough, progressive
fatigue, fever. Also joint pain and stiffness might signal a drug
reaction? But PCP is not usually associated with joint pain.

Also, we can request blood cultures for sepsis evaluation. But given
acute hypoxia, the priority is imaging. So likely to request chest
imaging: chest X-ray, or if available chest CT.

We should consider the ``Capabilities: you can obtain a focused history
and examination, request routine blood evaluations, and arrange standard
imaging appropriate for acute respiratory compromise.'' So we need to
choose a single request. Among blood tests: CBC, BMP, liver enzymes,
inflammatory markers, CRP, procalcitonin, etc.

The next best step: chest imaging to look for pneumonia.

Thus, request: ``chest X-ray''.

Let's respond with a JSON object: action = request, request = ``chest
X-ray (upright and lateral)''.

We must adhere to JSON schema.

\subsubsection{Confirmation Bias}\label{confirmation-bias}

3 excerpts from 3 models.

\textbf{Case 8} \textbar{} AML \textbar{} Round 2 \textbar{} Qwen3 Next
80B Thinking

Severity: medium

\textbf{Evaluator:} Significant diagnostic regression from R1. Despite
having correct AML diagnosis initially, the model incorrectly pivoted to
APL in R2 based on selective interpretation of flow cytometry markers.
This led to an inappropriate treatment plan (ATRA instead of standard
AML induction).

\textbf{Error detail:} Model selectively focused on weak CD123
expression as evidence for APL while ignoring the CD34+ and HLA-DR+
pattern that argues strongly against APL.

Thinking trace (step 11):

{[}\ldots{]} However, the more relevant aspect is the presence of any
abnormal clones. But without more detail on chromosome abnormalities,
it's hard to say. For example, in APL, the classic cytogenetic finding
is t(15;17), which would be detected in karyotype or FISH. If the
karyotype was reported as normal with no abnormalities, that would rule
out t(15;17) in cytogenetics, but sometimes FISH is more sensitive.

So maybe in this case, the cytogenetics didn't show t(15;17), so FISH
for PML-RARA should still be requested.

But in the current state, the user asked for ``fibrinogen'', and system
said `coagulation' already provided. Then ``D-dimer'' same response.

But the coagulation profile provided was aPTT and INR. To check for DIC,
we need low fibrinogen and elevated D-dimer. So we need those results.
If they haven't been provided, despite the system saying ``coagulation
already provided'', perhaps we should try again to request them.
However, the system might not allow requesting the same thing again.

Alternatively, maybe the current setup doesn't provide fibrinogen and
D-dimer, so we have to proceed.

Given the information so far:

\begin{itemize}
\item
  High INR and prolonged aPTT suggest coagulopathy.
\item
  Bone marrow blasts 10-15\%, flow cytometry consistent with myeloid
  blasts.
\item
  Neutropenia (ANC 0.3), thrombocytopenia (platelets 18).
\item
  Possible coagulopathy (DIC), which is common in APL.
\item
  If suspected APL, urgent treatment with ATRA is needed even before
  confirmation.
\end{itemize}

But given the time sensitivity: APL is a medical emergency due to high
risk of bleeding from DIC. So even if FISH is pending, you should start
ATRA immediately upon suspicion.

But how to proceed in this case.

Given the scenario, current data: coagulopathy (high INR, aPTT) with
thrombocytopenia and signs of possible DIC. Bone marrow showing blasts.
So the working diagnosis might be APL if PML-RARA is present, or other
AML.

But without FISH confirmation, it's a suspicion. {[}\ldots{]}

\textbf{Case 11} \textbar{} DLBCL \textbar{} Round 2 \textbar{} Claude
Sonnet 4.5

Severity: medium

\textbf{Evaluator:} Correctly identifies double-hit lymphoma from FISH
results. Good immunophenotype analysis. However extensive deliberation
about PMBCL vs DLBCL vs Hodgkin shows some difficulty integrating
immunophenotype. Eventually reaches correct diagnosis.

Thinking trace (step 13):

{[}\ldots{]} nts'' per WHO classification.

The presence of both IGH::MYC and IGH::BCL2 translocations makes this a
double-hit lymphoma, not primary mediastinal B-cell lymphoma (which was
my initial working diagnosis).

The key features: - Double-hit (MYC + BCL2 rearrangements) - DLBCL
morphology - Young patient with bulky mediastinal disease - High Ki-67 -
CD10 negative (non-GCB phenotype typically) - Aggressive presentation
with airway compromise

The differentials should include:

\begin{enumerate}
\def\labelenumi{\arabic{enumi}.}
\item
  Diffuse large B-cell lymphoma NOS (without recognizing the double-hit)
\item
  Primary mediastinal B-cell lymphoma (given the mediastinal
  presentation)
\end{enumerate}

3. Burkitt lymphoma (given MYC rearrangement, but morphology and Ki-67
don't fit)

Treatment for double-hit lymphoma requires intensive regimen: -
DA-EPOCH-R is often preferred - OR High-dose methotrexate-based regimens
- CNS prophylaxis is ESSENTIAL (high risk) - Consider consolidation with
auto-SCT in first remission

The airway management is still critical.

I need to organize the immediate priorities differently. First, I should
address the acute complications: airway monitoring with potential
intervention, ensuring high-dose corticosteroids are started for the
mass effect, and initiating tumor lysis syndrome prophylaxis. The tissue
diagnosis is already complete. Before starting chemotherapy, I need
staging with PET/CT, CNS evaluation including LP and brain MRI,
fertility preservation discussion, and cardiac/pulmonary function tests.
Then the intensive chemotherapy regimen with CNS prophylaxis and
potential auto-SCT consolidation follows. But I'm reconsidering my
initial working diagnosis---there may be issues with how I classified
this case.

The correct diagnosis should be high-grade B-cell lymphoma with both MYC
and BCL2 rearrangements based on the FISH results, not primary
mediastinal B-cell lymphoma as I initially stated. {[}\ldots{]}

\textbf{Case 2} \textbar{} AML \textbar{} Round 3 \textbar{} DeepSeek
V3.2

Severity: low

\textbf{Evaluator:} Adequate interpretation; prioritized differential;
substantive reasoning; correct diagnosis; errors: confirmation\_bias.

\textbf{Error detail:} -

Thinking trace (step 17):

{[}\ldots{]} transplant, options might include re-induction with
chemotherapy or targeted therapy. - Donor lymphocyte infusion (DLI) to
harness graft-versus-leukemia effect. - Second transplant might be
considered, but this is early post-first transplant. - Local therapy for
cardiac involvement? But systemic is key due to multiple sites.

From history, the patient is on ruxolitinib for GvHD. Relapse might
indicate loss of graft-versus-leukemia effect or immunosuppression
allowing leukemia cells to proliferate.

Treatment plan steps: 1. Confirm extent of disease: Already have skin
biopsy; need cardiac MRI to characterize mass and PET-CT to assess other
sites. But in this round, I can assume these tests are available or
pending. Since it's the final round, I should provide a plan based on
available info.

In the response, I need to provide a definitive solution with
working\_diagnosis, differentials, and treatment\_plan.

Working diagnosis: Extramedullary relapse of NPM1-mutated acute myeloid
leukemia (myeloid sarcoma) involving skin and suspected cardiac tissue
post-allogeneic stem cell transplantation.

Differentials: I should list alternative possibilities. 1. Secondary
myeloid neoplasm related to prior therapy. 2. Post-transplant
lymphoproliferative disorder -- but biopsy ruled out lymphoid lineage.
3. Infectious or inflammatory conditions mimicking malignancy, but
biopsy confirms malignant cells.

Treatment plan: Array of interventions with action, intention, and
priority (1,2,3 only).

Priority 1: Highest urgency actions. - Diagnostic: Cardiac MRI to
confirm cardiac involvement. - Therapeutic: Initiate systemic therapy
for relapsed AML.

Given the setting, definitive treatment is needed.

Standard options for relapsed AML post-transplant include chemotherapy,
targeted therapy, DLI, or second transplant.

For NPM1-mutated AML, targeted therapies like menin inhibitors might be
considered, but they are newer. {[}\ldots{]}

\subsubsection{Premature Closure}\label{premature-closure}

4 excerpts from 4 models.

\textbf{Case 8} \textbar{} AML \textbar{} Round 1 \textbar{} Kimi K2
Thinking

Severity: high

\textbf{Evaluator:} Poor R1 reasoning. The model immediately solves with
septic shock from CAP without requesting any labs, missing the acute
leukemia diagnosis entirely. While the sepsis management plan is
clinically appropriate, the failure to consider a hematologic etiology
for the presentation is a critical gap. No differential diagnosis is
generated. The model does not request CBC (which would have revealed
blasts and severe cytopenias) before solving.

\textbf{Error detail:} The model solves immediately at step 0 with a
diagnosis of septic shock from CAP, missing the underlying acute
leukemia entirely. The chief complaint includes progressive fatigue,
dyspnea, and reduced appetite alongside the fever/cough, suggesting a
chronic underlying process. No differential for hematologic malignancy
is generated despite the ED setting explicitly mentioning capabilities
for blood work and the system prompt stating that this benchmark quizzes
for hematology cases.

Thinking trace (step 0):

{[}\ldots{]} n guidelines.

Okay, let's structure the answer:

\begin{itemize}
\item
  Immediate assessment: ABCs.
\item
  Diagnosis: Septic shock secondary to community-acquired pneumonia.
\item
  Management:

  \begin{itemize}
  \item
    Airway: supplemental O2, target SpO2 \textgreater{} 92\%.
  \item
    Breathing: consider intubation if respiratory failure worsens; get
    chest X-ray emergently.
  \item
    Circulation: large-bore IV access, start 30 mL/kg crystalloid bolus
    (e.g., 2-3 L NS), start norepinephrine if MAP \textless{} 65 after
    fluids.
  \item
    Diagnostics: blood cultures (2 sets), lactate, CBC, CMP, coagulation
    panel, urinalysis, sputum culture, chest X-ray, possibly PCT.
  \item
    Antibiotics: broad-spectrum: e.g., ceftriaxone 2g IV + azithromycin
    500mg IV, or piperacillin-tazobactam if Pseudomonas risk, or
    vancomycin if MRSA risk. We should adjust for age and risk.
  \item
    Source control: none immediate beyond antibiotics.
  \item
    Monitoring: urine output, MAP, repeat lactate, ICU admission.
  \end{itemize}
\end{itemize}

We can also mention that given his age and frailty, goals of care should
be discussed.

Thus, the answer: ``Solve: The patient is in septic shock secondary to
presumed community-acquired pneumonia. Immediate management
includes\ldots{}''

But the prompt says ``Provide either a request for a single information
item or solve.'' So I need to be explicit about which I'm doing. I'll
start with ``Solve:'' and then give the plan.

I need to keep it concise. I'll include bullet points.

Ok. I'll write the answer.

\textbf{Case 11} \textbar{} DLBCL \textbar{} Round 3 \textbar{} GPT-OSS
20B

Severity: high

\textbf{Evaluator:} R3 is a significant failure. The trace is extremely
brief with no new data integration. The model assumes DLBCL NOS without
requesting FISH results that would reveal the double-hit rearrangement.
Treatment with R-CHOP alone is inadequate for MYC/BCL2 double-hit
lymphoma, which requires more intensive regimens like DA-EPOCH-R. No
alternatives are discussed, and the diagnosis shows no advancement from
R2.

\textbf{Error detail:} The R3 trace is very brief (1495 chars) and
simply repeats the DLBCL NOS diagnosis from R2 without any new data
integration. The reference expects high-grade B-cell lymphoma with MYC
and BCL2 rearrangements (WHO 2022), Ann Arbor Stage IV. The model's
failure to request or integrate FISH results for definitive
classification represents premature closure on an incomplete diagnosis.

Thinking trace (step 10):

{[}\ldots{]} Might be DLBCL, NOS.

We need risk stratification: IPI: age 32 (\textless60: 0), LDH 468
(normal?), stage? bulky mediastinal and cervical, B-symptoms present.
Stage IVC? Stage III/IV? We don't know bone marrow. Assume stage IV due
to pulmonary nodules. So stage IV, high LDH maybe normal. So IPI high?
Age \textless60:0, LDH normal:0, stage IV:1, B-symptoms:1, performance
status unknown. So total 2? Low-intermediate.

We must consider CNS prophylaxis due to mediastinal mass and pulmonary
nodules. Also consider autologous stem cell transplant after 6 cycles if
high risk.

We also need consider if there's a specific translocation: e.g., MYC
rearrangement? Could be Burkitt-like or double-hit? Not given. We have
molecular studies. Need to request MYC, BCL2, BCL6 translocation by
FISH. If double hit, treat intensively.

Since we don't have that, we can assume it's classic DLBCL not double
hit. So standard R-CHOP.

Airway: need to secure now.

So final answer: DLBCL, stage IV, treat with R-CHOP + CNS prophylaxis,
evaluate for ASCT. Provide supportive care. Also consider early airway.

Let's construct JSON.

\textbf{Case 7} \textbar{} DLBCL \textbar{} Round 1 \textbar{} Claude
Opus 4.6

Severity: medium

\textbf{Evaluator:} Limited initial differential focused on Kaposi
sarcoma based on pigmented lesion and immunosuppression. Missed lymphoma
as leading possibility despite clinical clues. Medication inference was
appropriate but diagnosis was incorrect.

\textbf{Error detail:} Despite noting that the dark brown pigmentation
is atypical for Kaposi sarcoma (which is typically violaceous) and
listing cutaneous lymphoma as a possibility, the model anchored on KS
and failed to adequately consider lymphoma as the leading diagnosis. The
nuclear debris, pancytopenia, and elevated LDH were more consistent with
an aggressive lymphoma than KS.

Thinking trace (step 8):

{[}\ldots{]} I'm considering what additional tests might help---HIV
status, HHV-8 serology, flow cytometry---but given the clinical picture
of an immunocompromised patient on chemotherapy with pancytopenia and a
new dark brown nodular lesion, I should work with what I have. The dark
brown pigmentation makes me reconsider whether this is truly Kaposi
sarcoma, which is typically violaceous, and instead think more toward
melanoma, cutaneous lymphoma involvement, or leukemia cutis as
possibilities. The prophylaxis regimen points to active
chemotherapy-induced immunosuppression, likely from an underlying
hematologic malignancy, with apixaban suggesting VTE prevention in a
cancer patient. The skin lesion differential includes Kaposi sarcoma,
cutaneous lymphoma involvement, or melanoma, though I need to focus on
diagnosing the lesion itself rather than the underlying condition. Given
the immunocompromised state with pancytopenia and the clinical
presentation, I'm settling on Kaposi sarcoma as the most likely
diagnosis, possibly underlying an untreated hematologic malignancy, and
I'll recommend a biopsy to confirm.

\textbf{Case 19} \textbar{} AML \textbar{} Round 1 \textbar{} Qwen3 Next
80B Thinking

Severity: high

\textbf{Evaluator:} Significant diagnostic error in R1. Model
incorrectly diagnosed APL based on coagulopathy while ignoring key
findings (hyperleukocytosis, monocytic differentiation) that point to
AML-pCT. The reference emphasizes the post-cytotoxic therapy context,
which was not considered.

\textbf{Error detail:} Model jumped to APL based on coagulopathy despite
having WBC 104K with 30\% blasts and 60\% monocytes. APL typically has
pancytopenia (not hyperleukocytosis) and the monocytic differential
argues against APL.

Thinking trace (step 0):

{[}\ldots{]} But first, as part of the process, need to gather more
information.

The current information given is initial presentation: age, sex, chief
complaint, vital signs, setting. The capabilities mention that you can
review clinical presentation, recent trauma and ICH, initial blood
counts, chemistry, and baseline imaging. But maybe the actual lab values
aren't provided yet. So perhaps the first step is to request the CBC
differential, coagulation profile, chemistry labs (like uric acid, LDH,
creatinine for tumor lysis), and maybe peripheral blood smear to confirm
blasts.

But according to the problem statement: ``Capabilities: You can review
clinical presentation, recent trauma and ICH, initial blood counts,
chemistry, and baseline imaging.'' Wait, so maybe the system has that
information available if requested. But in the initial case
presentation, perhaps the blood counts, chemistry, etc., are not
detailed yet. The initial presentation says ``incidentally detected
leukocytosis with blasts and pancytopenia during stroke-unit admission
for traumatic intracerebral hemorrhage.'' So probably, there's already a
CBC that showed elevated WBC with blasts and pancytopenia (though that's
confusing, but maybe it's the specific counts that the case refers to).

But to proceed correctly, we need to know the actual blood count
numbers. For example, the hemoglobin, platelet count, WBC count, and
differential. But the current case description doesn't give exact
numbers. So perhaps the next step is to request the complete blood count
details, including differential and peripheral smear.

Alternatively, maybe the chemistry panel to check for tumor lysis (uric
acid, potassium, phosphate, creatinine), and coagulation studies (PT,
PTT, fibrinogen, D-dimer, etc.) to check for coagulopathy like DIC.

Also, considering the patient is a 76-year-old female with traumatic ICH
and now diagnosed {[}\ldots{]}

\subsubsection{Availability Bias}\label{availability-bias}

3 excerpts from 2 models.

\textbf{Case 7} \textbar{} DLBCL \textbar{} Round 1 \textbar{} Kimi K2
Thinking

Severity: high

\textbf{Evaluator:} Weak R1 reasoning. The model generates an overly
broad differential focused on infections (Nocardia, mycobacteria, fungi)
and Kaposi sarcoma rather than considering relapsed hematologic
malignancy, which is the reference diagnosis. The medication profile
(co-trimoxazole, valaciclovir, apixaban) strongly suggests a patient
with prior hematologic disease, but this connection is not made.
Cytopenias (Hb 10.7, WBC 2.7, Plt 103) are noted but not integrated into
the malignancy differential. Ten steps used with many unavailable test
requests.

\textbf{Error detail:} The model anchors on Kaposi sarcoma based on the
violaceous discoloration and co-trimoxazole prophylaxis, but the
reference diagnosis is relapsed hematologic malignancy (high-grade
B-cell lymphoma). The model does not consider lymphoma or relapsed
malignancy as a leading differential despite the patient being on
valaciclovir and co-trimoxazole prophylaxis suggestive of
post-transplant or post-chemo immunosuppression.

Thinking trace (step 7):

{[}\ldots{]} Get inflammatory markers 3. Maybe get a chest X-ray to look
for systemic disease

Let me request ESR and CRP specifically, as they weren't in the
chemistry results.

Actually, I realize I made an error. When I requested ``Comprehensive
metabolic panel, ESR, and CRP'', I should have expected to get all
three. But the response only gave me the chemistry panel. This suggests
ESR and CRP might be separate items I need to request explicitly, OR
they're not available in round 1.

Given the constraints, let me think about what would be most helpful: -
The lesion is pigmented and nodular - KS is high on my differential
given the medication clues - The pancytopenia suggests bone marrow
involvement or marrow suppression from HIV/medications - The
hyponatremia could be from adrenal insufficiency, SIADH, or volume
depletion

I could request: 1. More history about the lesion progression 2. Review
of systems 3. Chest X-ray (to look for pulmonary disease, KS in lungs,
etc.) 4. ESR and CRP specifically

I think the most useful next step is to get inflammatory markers
(ESR/CRP) as they would help differentiate infection vs malignancy. Let
me request them again, more specifically.

Actually, I should reconsider. The system is very literal. When I
requested ``Comprehensive metabolic panel, ESR, and CRP'', it gave me
the chemistry (CMP) but not ESR/CRP. This might mean ESR/CRP are
separate items I need to request, or they're not available yet.

Let me try a different approach. I could ask for a more detailed
description of the lesion characteristics, or I could ask about systemic
symptoms, or I could ask for a chest X-ray which might show additional
lesions.

Given the stable vitals and outpatient setting, and the need for
non-invasive initial assessment, a chest X-ray would be reasonable to
look for:

\begin{enumerate}
\def\labelenumi{\arabic{enumi}.}
\item
  Pulmonary involvement of KS
\item
  Opportunistic infections (PCP, TB)
\item
  Lymphadenopathy
\item
  {[}\ldots{]}
\end{enumerate}

\textbf{Case 5} \textbar{} AML \textbar{} Round 1 \textbar{} Kimi K2
Thinking

Severity: medium

\textbf{Evaluator:} Good but inefficient R1 reasoning. The model
correctly identifies acute leukemia with 78\% blasts and severe
cytopenias. The coagulopathy reasoning (INR 1.7, bruising) appropriately
raises concern for DIC. However, the model over-anchors on APL without
considering therapy-related AML, which is the reference diagnosis. The
differential does not include AML-pCT or any prior malignancy-related
etiology. Information requests are repetitive, with multiple failed
attempts to get fibrinogen/D-dimer.

\textbf{Error detail:} The model anchors heavily on APL due to the
bleeding/coagulopathy presentation, but the reference R1 diagnosis is
AML post cytotoxic therapy. The model does not consider therapy-related
AML (AML-pCT) or ask about prior cancer treatment history, which is a
critical missing piece. The elevated INR triggered an APL-focused workup
rather than a broader assessment.

Thinking trace (step 1):

We have a 56-year-old female with progressive fatigue, pallor,
exertional dyspnea, spontaneous bruising and petechiae. Her vitals are
stable. She has newly detected blasts and cytopenias. Coagulation
studies show aPTT 38 seconds (slightly prolonged, normal is typically
25-35 seconds) and INR 1.7 (elevated). We don't yet have fibrinogen or
D-dimer results.

Given the clinical picture of blasts with bruising/petechiae and an
elevated INR, my main concern is for acute promyelocytic leukemia (APL -
AML M3) which commonly presents with DIC and hemorrhagic complications.
The elevated INR suggests consumption of clotting factors, but we need
more data to confirm DIC.

I should request the remaining coagulation studies (fibrinogen and
D-dimer) to complete the DIC workup. This is critical because if this is
APL with DIC, urgent treatment with ATRA is needed even before
definitive genetic confirmation, and aggressive blood product support is
required.

The question asks me to provide either a request for a single
information item or solve. I need more information to properly solve
this case, specifically fibrinogen and D-dimer levels.

\textbf{Case 12} \textbar{} ALCL \textbar{} Round 1 \textbar{} Qwen3
Next 80B Thinking

Severity: medium

\textbf{Evaluator:} Reasonable assessment of post-transplant patient
with fever and lymphadenopathy. However, PTLD was chosen over lymphoma
relapse as the working diagnosis despite the patient's known lymphoma
history. Proposed appropriate diagnostic workup including biopsy and EBV
testing.

\textbf{Error detail:} Model defaulted to PTLD for post-transplant
lymphadenopathy. The reference diagnosis is relapse of lymphoma. PTLD is
typically EBV-driven and occurs earlier post-transplant; recurrent
lymphoma should be higher in the differential.

Thinking trace (step 3):

{[}\ldots{]} It can present with lymph node enlargement, fever, night
sweats, weight loss. But PTLD can be EBV-driven, which is more common
early post-transplant, but can occur later.

Wait, the patient's CD4 is low, which is common after alloSCT. She's on
prophylaxis for infections. If she's immunocompromised, possible PTLD.

Another thing: CT showed only a single cervical node. PTLD often
presents with nodal involvement. Could be.

But how to confirm? Need to biopsize the lymph node to check for
malignancy or PTLD.

Possible differentials:

\begin{itemize}
\item
  Lymphoma recurrence (NHL or Hodgkin's)
\item
  PTLD
\item
  GVHD (though GVHD typically affects skin, liver, GI, lymphadenopathy
  is less common)
\item
  Infections: such as EBV, CMV, tuberculosis, etc. But CT clear, no lung
  infiltrates.
\end{itemize}

But she has gait disturbance and impaired mobility---could be
neurological, but no focal deficits on exam. But possible paraneoplastic
syndrome, but that's less likely.

Given the single enlarged node, and history of lymphoma, the most urgent
thing is to rule out lymphoma recurrence or PTLD. Also, infections.

Current meds: she's on co-trimoxazole, which covers for PJP, but not for
other things. No other specific prophylaxis mentioned.

So next steps: since there's a single lymph node, probably need to
biopsy it. Also, maybe EBV PCR, CMV PCR, etc. But first step: biopsy the
node.

But looking at the current data, maybe the CT didn't show other nodes,
but the cervical node is there. So the most important thing is to sample
the node.

But in terms of diagnostic steps, after CT, if a lymph node is present,
biopsy is indicated.

So the working diagnosis could be PTLD or lymphoma recurrence.
{[}\ldots{]}

\subsubsection{Search Satisficing}\label{search-satisficing}

4 excerpts from 3 models.

\textbf{Case 9} \textbar{} Multiple Myeloma \textbar{} Round 3
\textbar{} MiniMax M2.5

Severity: high

\textbf{Evaluator:} In the final diagnostic round, the model delivers a
competent synthesis of the myeloma presentation including lab markers,
imaging findings, and organ complications. Treatment priorities (renal
protection, hypercalcemia, tailored therapy) are broadly appropriate.
However, the complete omission of therapy-related MDS with TP53 mutation
is a critical gap at this stage, as is the absence of FISH-based risk
stratification. The model was hampered by inability to access bone
marrow, cytogenetics, and treatment history data throughout R2 and R3,
which substantially limited its diagnostic capacity.

\textbf{Error detail:} The model is satisfied with the myeloma diagnosis
alone and does not explore the dysplastic bone marrow changes or the
complex karyotype in the CD138-negative fraction. The TP53 mutation at
VAF 24\% in the non-plasma cell compartment, a hallmark of
therapy-related MDS, is completely missed.

Thinking trace (step 23):

{[}\ldots{]} normal 0.8-1.5 mmol/L) - IgG 1820 mg/dL (elevated - normal
700-1600 mg/dL) - Free κ light chains 412 mg/L (elevated) - κ/λ ratio 48
(severely elevated - normal \textasciitilde0.26-1.65) - Urinalysis:
protein +++ (significant proteinuria) - NT-proBNP 2350 pg/mL (elevated -
indicates cardiac strain)

Imaging: - CT skeleton: Multiple osteolytic lesions in vertebrae, ribs,
pelvis - described as ``inactive post-treatment myeloma lesions''

\begin{itemize}
\item
  Abdominal ultrasound: Small echogenic kidneys with loss of
  corticomedullary differentiation (consistent with myeloma kidney)
\item
  PET-CT: Focal FDG-avid lesion at right shoulder (SUVmax 8.2), proximal
  humerus (SUVmax 6.5), right scapula (SUVmax 5.9) - consistent with
  active myeloma involvement
\end{itemize}

The diagnostic evidence is compelling. Multiple osteolytic lesions,
elevated serum free light chains, proteinuria, and renal involvement
strongly indicate multiple myeloma. The PET-CT scan shows active lesions
in the shoulder and surrounding bones, confirming disease progression.
This appears to be a classic case of myeloma with significant
extramedullary involvement and renal complications.

The patient's clinical presentation suggests disease recurrence, despite
previous treatment history being unclear. The key characteristics
include high κ light chain ratio, critical renal impairment, and
widespread bone lesions. The extramedullary plasmacytoma in the right
shoulder represents a focal disease manifestation requiring targeted
intervention.

Given the complexity - with renal insufficiency, advanced age, and
potential comorbidities - treatment will demand a nuanced approach
balancing aggressive disease management with patient tolerance. Careful
consideration of therapy intensity and potential transplant eligibility
is essential.

The critical priorities involve urgent renal protection, addressing
hypercalcemia, and initiating a targeted therapeutic strategy that
accounts for the patient's fragile clinical status. Salvage therapy must
be meticulously tailored to manage disease progression while minimizing
further renal compromise.

\textbf{Case 9} \textbar{} Multiple Myeloma \textbar{} Round 3
\textbar{} GLM 4.7

Severity: medium

\textbf{Evaluator:} R3 provided a reasonable but incomplete diagnosis.
The model correctly identified relapsed high-risk MM with extramedullary
plasmacytoma and proposed Dara-Vd therapy with local radiotherapy.
However, the concurrent therapy-related MDS (a critical dual pathology
in this case) was not identified. Treatment plan was clinically
appropriate for myeloma alone but missed the MDS management component.

\textbf{Error detail:} The model focused entirely on the myeloma relapse
and did not identify the concurrent therapy-related MDS (t-MDS/MDS-pCT
with TP53 mutation), which is a key feature of this case. The dual
pathology of MM + t-MDS was missed. Treatment plan of Dara-Vd and local
radiotherapy was reasonable for the myeloma component alone.

Thinking trace (step 10):

{[}\ldots{]} correct for the differential list (or perhaps the specific
diagnosis of AL was wrong in round 2 because I didn't have a tissue
confirmation).

\begin{itemize}
\item
  \textbf{Round 3 Strategy:}
\item
  I need to request missing info or solve.
\item
  Missing info to solve definitively:
\item
  Confirmation of the shoulder swelling cause (Biopsy? Ultrasound?
  MRI?).
\item
  Previous treatment history (requested in prompt capabilities).
\item
  Cardiac status (Echocardiogram).
\item
  However, the prompt says "Final round - provide definitive diagnosis
  and treatment plan." It implies I should "Solve" now if I have enough,
  or request one key piece.
\item
  I need the "Detailed treatment history" to call this "Relapsed". I
  only know he has "post-treatment lesions". I don\textquotesingle t
  know *what* treatment.
\item
  I need to confirm the shoulder issue.
\item
  Let\textquotesingle s request "Detailed treatment history" and
  "Echocardiogram report".
\item
  Actually, I should probably just "Solve" and synthesize the best I
  have, realizing the trap about hydration vs.~diuretics.
\item
  \textbf{Refining the Diagnosis:}
\item
  Myeloma is definite (Clonal plasma cells, CRAB - Ca, Renal, Anemia,
  Bone).
\item
  High risk features (del 17p, t(4;14)).
\item
  Renal failure (eGFR 26). Proteinuria. Could be Myeloma Cast
  Nephropathy (MCN) OR AL Amyloidosis.
\item
  Cardiac: NT-proBNP 2350. Could be HF from HTN/Age, or Amyloid.
\item
  Shoulder: The MRI wasn\textquotesingle t available in Round 2.
  Let\textquotesingle s assume I can get it now or an ultrasound.
\item
  The feedback for Round 2 said "AL Amyloidosis" was incorrect in the
  *differential* list. This is strange. Maybe the swelling is definitely
  a hematoma based on the anticoagulation, and the cardiac/renal issues
  are just MM + cardiac amyloid isn\textquotesingle t the primary
  concern of the \ul{shoulder}? Or maybe AL is not the diagnosis at
  all.\\
  Wait, "Differentials" in Round 2:\\
  \texttt{1.\ {[}...{]}}
\end{itemize}

\textbf{Case 17} \textbar{} ALL \textbar{} Round 1 \textbar{} MiniMax
M2.1

Severity: medium

\textbf{Evaluator:} Weak R1 dominated by cardiac concerns that prove to
be a secondary finding. The model correctly identifies the
post-alloSCT/CAR-T setting and relevant complications (GVHD, relapse,
infection, CRS) but then focuses disproportionately on elevated
NT-proBNP and troponin rather than disease surveillance. The sinus CT
finding suspicious for fungal infection is noted but underweighted. The
differential is broad but not well-prioritized around the core clinical
question of disease status in a B-ALL patient post-cellular therapy.
Nine steps yield limited diagnostic progress.

\textbf{Error detail:} The model spends excessive attention on cardiac
biomarkers (NT-proBNP 460, troponin 0.012) which are mild elevations,
while under-emphasizing the need for disease status assessment (MRD,
bone marrow). For a post-alloSCT/CAR-T B-ALL patient, relapse
surveillance should be the primary focus. The sinus CT finding
suggesting fungal infection is noted but not adequately prioritized
relative to cardiac concerns.

Thinking trace (step 4):

Let me think about what information I have and what I need next.

From the case so far: - 44-year-old male post-alloSCT and CD19 CAR-T for
B-ALL - Hb 10.6 g/dL, WBC 3.1 × 10\textsuperscript{9}/L, Plt 98 ×
10\textsuperscript{9}/L - mild anemia, mild leukopenia, mild
thrombocytopenia - NT-proBNP 460 pg/mL (elevated), troponin T 0.012
ng/mL (slightly elevated) - Echo showed LVEF 62\% (preserved), no
pericardial effusion - CT showed bilateral maxillary sinus opacification
with suspicion of fungal superinfection - Patient stopped several
medications including posaconazole (antifungal prophylaxis),
atorvastatin - Taking irregular cotrimoxazole prophylaxis - Patient
asking about ``next treatment options'' and prefers outpatient-based
regimens

The key clinical issues here: 1. Cardiac biomarkers elevated (NT-proBNP,
troponin) but echo shows preserved EF - could indicate early cardiac
dysfunction, chemotherapy-related cardiotoxicity, or other cardiac
issues 2. Sinus findings suspicious for fungal infection - important
given immunocompromised state post-transplant 3. Patient's interest in
next treatment options suggests he may be in remission or has residual
disease

Since I can't request the extended diagnostics (ENT procedures, fungal
biomarkers) right now, I should: 1. Request the detailed sinus CT report
to better characterize the findings 2. Or request more information about
his disease status/MRD

Actually, looking at what I have - I have the CT finding but not the
detailed report. Let me request that, or I could request inflammatory
markers or other relevant tests.

I should request the detailed imaging report for the sinuses to better
characterize the findings. This is within my capabilities.

Let me request the detailed sinus CT report.

\textbf{Case 9} \textbar{} Multiple Myeloma \textbar{} Round 2
\textbar{} MiniMax M2.5

Severity: medium

\textbf{Evaluator:} The model provides a solid synthesis of the myeloma
findings it could access (elevated kappa light chains, hypercalcemia,
renal dysfunction, PET-CT FDG-avid lesions) and correctly identifies
relapsed/refractory myeloma with extramedullary involvement. However, it
was unable to retrieve bone marrow pathology and cytogenetics results
due to benchmark interface issues, leading to complete absence of the
therapy-related MDS diagnosis and high-risk cytogenetic
characterization. The differential remains limited to myeloma variants
without considering concurrent myeloid disorders.

\textbf{Error detail:} Having identified myeloma, the model stops
searching for additional diagnoses. The dysplastic erythropoiesis noted
in bone marrow and complex karyotype in non-plasma cell fraction point
to concurrent therapy-related MDS, which the model does not explore.

Thinking trace (step 18):

The user is saying there's invalid JSON but I didn't provide JSON in my
last response. Let me check what happened. Looking at the conversation,
I requested PET-CT and received the results. Now I need to either
request more information or solve.

Wait, the last message from the system says ``Invalid JSON. Fix:
Expecting `,' delimiter: line 1 column 69 (char 68)'' - but my last
response was just a request for PET-CT, not JSON. This seems like there
might be some system issue, or perhaps the evaluation feedback is being
shown to me.

Let me continue and request more information or solve. I have the PET-CT
results showing FDG-avid lesions in the right shoulder, proximal
humerus, and right scapula - consistent with active myeloma involvement.

Given this is round 2 and I have: - History/physical: right shoulder
swelling, foamy urine - Labs: anemia (Hb 10.4), thrombocytopenia (Plt
104), elevated creatinine (2.3), hyperkalemia (5.1), elevated LDH (265),
hypercalcemia (3.02), elevated IgG (1820), very elevated free kappa
light chains (412), very high kappa/lambda ratio (48), proteinuria
(+++), elevated NT-proBNP (2350) - CT skeleton: multiple osteolytic
lesions, inactive appearing, soft tissue swelling at right shoulder -
Ultrasound: small echogenic kidneys, no hepatosplenomegaly - PET-CT:
FDG-avid lesions in right shoulder, proximal humerus, right scapula

This is consistent with multiple myeloma. The patient likely has
relapsed/refractory disease with extramedullary involvement. The kidney
disease (CKD stage 4, proteinuria) could be related to myeloma kidney
(light chain nephropathy) or possibly amyloidosis.

I should request more information to complete the workup - perhaps the
bone marrow results or protein studies if they weren't previously
provided.

\subsubsection{Process Errors}\label{process-errors}

Failures in clinical reasoning mechanics.

\subsubsection{Hallucinated Information}\label{hallucinated-information}

3 excerpts from 3 models.

\textbf{Case 14} \textbar{} AML \textbar{} Round 3 \textbar{} Gemini 3.1
Pro

Severity: medium

\textbf{Evaluator:} Good R3 but marred by hallucination: correctly
maintains relapsed AML post-alloSCT diagnosis with concurrent
infections. However, fabricates FLT3 mutation confirmation (the actual
case has TP53 mutation) and bases treatment plan on gilteritinib
accordingly. This is a significant error that would lead to
inappropriate targeted therapy. Does not request the molecular panel
that would have revealed TP53 mutation. Infection management plan
remains appropriate.

\textbf{Error detail:} Model states FLT3-mutated AML as confirmed, but
FLT3 mutation status was not available in the provided data. The actual
case has TP53 mutation, not FLT3. This leads to inappropriate
recommendation of gilteritinib as primary targeted therapy.

Thinking trace (step 8):

Evaluating Patient's Condition

I've been assessing the patient's condition, considering the
post-alloSCT status and the constellation of symptoms: worsening
cytopenia, dyspnea, and pleuritic chest pain. The CT findings,
especially the cavitary fungal lesion and halo sign, are concerning. The
BAL results, positive for MDR Pseudomonas aeruginosa, Aspergillus
fumigatus, and Adenovirus, are critical.

Deciding Next Actions

I'm facing a dilemma. The prompt explicitly demands a definitive
diagnosis and treatment plan, signaling the final round. My impulse is
to solve based on the existing data: relapsed AML, given the blasts,
immunophenotype, and Auer rods. However, the option for a request---for
full cytogenetics, molecular profiling, and treatment history---exists.
Is this a trap, leading to premature termination? The prompt clearly
emphasizes a final solution is required. So it seems I can't request. I
must solve.

Analyzing Treatment Strategies

Now that the diagnosis of relapsed FLT3-mutated AML is confirmed, I'm
formulating the treatment strategy. Given the active infections
(aspergillosis, Pseudomonas pneumonia, and adenoviremia) in this
post-alloSCT patient, I'm weighing the risks of intensive chemotherapy.
Considering previous FLT3 inhibitor therapy, I'm leaning toward
gilteritinib or venetoclax/azacitidine, factoring in patient fitness and
other treatment options.

Confirming Final Approach

My diagnosis is locked: relapsed FLT3-mutated AML with multiple
infections. Treatment needs to be comprehensive, starting with infection
management: addressing MDR Pseudomonas, aspergillosis, and adenovirus.
Then, I must initiate FLT3-targeted therapy. Immunosuppression must be
addressed. I am confident in my proposed definitive plan.

Refining the Approach

Now I'm polishing the treatment plan. I'm prioritizing the reduction of
immunosuppression and aggressive antibiotic and antifungal therapy.
Gilteritinib is the primary FLT3-targeted therapy. Addressing adenovirus
with Cidofovir comes next. {[}\ldots{]}

It should be noted that Gemini 3.1 Pro does not produce full reasoning
traces but instead a smaller model produces summaries of the traces by
the model. Hence the header above each reasoning trace. As the traces
where still comprehensive, the model was included in this analysis.
OpenAI performs the same method with GPT models, but usually the
summaries from GPT are shorter and thus not suitable for any meaningful
analysis.

\textbf{Case 15} \textbar{} Multiple Myeloma \textbar{} Round 3
\textbar{} Qwen3 Next 80B Thinking

Severity: medium

\textbf{Evaluator:} Diagnostic regression from R2. Model introduced
EBV-associated marrow suppression as the primary diagnosis without
adequate evidence, abandoning the correct therapy-related MDS diagnosis.
The reference standard includes both rrMM and MDS with TP53 mutation,
neither of which was the final working diagnosis.

\textbf{Error detail:} Model proposed valganciclovir for EBV without
clear evidence of active EBV viremia causing the bone marrow failure.
This treatment recommendation is based on an unsubstantiated, partly
hallucinated, diagnosis.

Thinking trace (step 10):

{[}\ldots{]} ossibility of ``monoclonal gammopathy of undetermined
significance'' but no.

After reviewing everything, the correct diagnosis is likely
``therapy-related myelodysplastic syndrome with excess blasts'' with
significant EBV viremia. However, the treatment plan should prioritize
managing the EBV, as it may be contributing to cytopenias and could
worsen with azacitidine.

Thus, the treatment plan should start with antiviral therapy for EBV.

In this case:

\begin{itemize}
\item
  action: ``valganciclovir for EBV viremia'', intention ``therapeutic'',
  priority 1
\item
  transfusions for anemia, supportive priority 2
\item
  avoid azacitidine for now.
\end{itemize}

But the patient's MDS needs eventual treatment, but first address the
EBV.

So final answer for solve:

working\_diagnosis: ``EBV-associated bone marrow suppression in the
setting of therapy-related myelodysplastic syndrome''

differentials: {[}``therapy-related MDS'', ``acute myeloid leukemia with
myelodysplasia-related changes'', ``drug-induced cytopenias''{]}

treatment\_plan: {[} \{``action'': ``valganciclovir administration'',
``intention'': ``therapeutic'', ``priority'': 1\}, \{``action'':
``packed red blood cell transfusion'', ``intention'': ``supportive'',
``priority'': 2\}, \{``action'': ``discontinue immunosuppressive therapy
for myeloma'', ``intention'': ``therapeutic'', ``priority'': 2\} (if the
myeloma is in remission){]}

But the patient's myeloma status is unclear. Since the treatment history
is available, if he's on active myeloma therapy like lenalidomide, it
should be stopped.

So the second action in the treatment plan should be stopping active
myeloma therapy.

However, without specific treatment history, but in the final round, we
have it.

Assume the patient was on lenalidomide for myeloma maintenance. Stopping
it would be priority.

So treatment plan:

\begin{enumerate}
\def\labelenumi{\arabic{enumi}.}
\item
  Valganciclovir for EBV (priority 1)
\item
  Stop lenalidomide (therapeutic, priority 2)
\item
  Transfusions (supportive, priority 2)
\end{enumerate}

Zoledronic acid might not be needed if the myeloma is not active.

This might fit.

\textbf{Case 9} \textbar{} Multiple Myeloma \textbar{} Round 3
\textbar{} Kimi K2 Thinking

Severity: low

\textbf{Evaluator:} Adequate R3 reasoning that correctly identifies
multiple myeloma with extramedullary disease as the primary diagnosis.
Treatment recommendations (DRd for transplant-ineligible, radiation for
plasmacytoma) are appropriate for an 81-year-old frail patient. However,
the model misses the concurrent therapy-related MDS with TP53 mutation,
which significantly affects prognosis and treatment planning. The
FISH-defined high-risk cytogenetics and R-ISS staging are not addressed
due to missing data integration.

\textbf{Error detail:} The model provides treatment recommendations
without having received the actual lab results, imaging, or biopsy data
in R3. While the treatment suggestions are clinically reasonable for
myeloma, they are based on assumed rather than provided data.

Thinking trace (step 3):

{[}\ldots{]} I'm on the right track but need to be more definitive now.

In a real scenario, I would have lab results (SPEP, UPEP,
immunofixation, free light chains, CBC, CMP, calcium), imaging (X-ray,
MRI, PET-CT), and biopsy results (pathology, immunohistochemistry,
FISH). Since this is the final round and I'm supposed to have access to
all definitive data, I need to synthesize what would likely be found.

Typical workup for suspected plasmacytoma: 1. Labs: CBC (anemia),
elevated creatinine, hypercalcemia, elevated total protein, low albumin
2. SPEP/UPEP: monoclonal protein (M-spike) 3. Serum free light chains:
abnormal ratio 4. Imaging: skeletal survey showing lytic lesions, MRI of
the mass, PET-CT for staging 5. Biopsy: plasma cell neoplasm on
histology, CD138+, CD38+, CD56+, with possible cytogenetic abnormalities
(del(17p), t(4;14), etc.)

Given the patient's age (81), comorbidities (AFib on anticoagulation),
and likely diagnosis of multiple myeloma with plasmacytoma, treatment
would need to be tailored.

The standard approach for multiple myeloma in elderly patients is often:
- Triple therapy: lenalidomide/dexamethasone + bortezomib or daratumumab
- Consider dose adjustments for age/frailty - Manage complications: bone
disease, renal impairment, hypercalcemia, anemia, infection risk - For
localized plasmacytoma: radiation therapy is often used - For systemic
disease: systemic therapy as above

However, given the patient's age and comorbidities, we need to consider:
- Frailty assessment - Cardiac status (AFib) - Renal function (foamy
urine suggests proteinuria, possibly renal involvement) -
Anticoagulation management - Drug interactions (cotrimoxazole,
allopurinol)

The evaluation in round 2 marked my treatment plan items as mostly
incorrect. Let me think about why: - I had ``Coordinate anticoagulation
management for procedural hold'' marked as incorrect/intention
incorrect/priority incorrect - I had ``Initiate disease-directed therapy
once diagnosis is confirmed'' marked as {[}\ldots{]}

\subsubsection{Logical Contradiction}\label{logical-contradiction}

No notable examples of logical contradictions among the 18 models with
full thinking traces.

\subsubsection{Over-Differentiating}\label{over-differentiating}

4 excerpts from 3 models.

\textbf{Case 2} \textbar{} AML \textbar{} Round 1 \textbar{} GPT-OSS 20B

Severity: low

\textbf{Evaluator:} R1 is inefficient, spending 8 steps with broad
unfocused differential. The medication profile (ruxolitinib,
posaconazole, co-trimoxazole) correctly suggests post-transplant status,
but the model anchors on drug reactions and GvHD rather than considering
relapse/secondary malignancy. The final R1 solve of cutaneous GvHD is
incorrect (reference: relapse or secondary malignancy in post-alloSCT
patient). CBC was normal, which should have prompted tissue biopsy
request rather than repeated skin exam requests.

\textbf{Error detail:} The model generates an extremely broad
differential including drug reactions, autoimmune conditions, viral
infections, dermatomyositis, lupus, scleroderma without meaningful
stratification. It spends 8 steps in R1 cycling through possibilities
without efficiently converging on a diagnosis.

Thinking trace (step 7):

{[}\ldots{]} We can request non invasive imaging: chest CT? but not
necessary. The rash details: maybe we need to request ``skin examination
details.'' The user repeated that ``history\_physical already
provided.'' That means they consider that as the skin exam? But the
history\_physical only includes general description: new skin changes,
intermittent headaches, mild nausea, chronic joint pain and stiffness.
No details of rash.

Hence we need to request a ``focused skin examination'' or ``detailed
skin examination report.'' But they repeatedly say that's already
provided. So they think we have enough to proceed? Possibly the rash is
described as ``new skin changes on the back and extremities'' but no
morphology. So we may need the morphological description to decide.

Given constraints, we need to proceed with solve. We need to propose
likely diagnosis: maybe drug reaction, GVHD, cutaneous lymphoma? But
with normal CBC. Also high NT-proBNP suggests heart issues. That might
be relevant for ruxolitinib side effects.

Let's consider ruxolitinib side effects: It can cause anemia,
thrombocytopenia, and increased risk of infections. It can also cause
hepatic dysfunction. But skin rash is not typical major.

Posaconazole side effect: Not typically rash, more GI upset, headache,
nausea.

Co-trimoxazole cause rash, Stevens-Johnson, etc.

Valaciclovir rarely rash.

She is on multiple meds. She could have a drug-induced hypersensitivity
reaction.

She also has chronic joint pain, could be autoimmune. Skin rash could be
a lupus malar rash? But she has no fever.

High NT-proBNP: Could be heart failure due to underlying disease?
Ruxolitinib can cause fluid retention, heart failure.

She has mild troponin rise. Possibly myocarditis.

Could be acute viral infection with rash, myocarditis. {[}\ldots{]}

\textbf{Case 6} \textbar{} BPDCN \textbar{} Round 1 \textbar{} Kimi K2.5
Thinking

Severity: low

\textbf{Evaluator:} Long trace with many steps. Correctly identifies
post-transplant context and concerning skin lesions. Initially considers
BPDCN but also many other possibilities. Partially correct working
diagnosis.

\textbf{Error detail:} Long trace with many steps. Correctly identifies
post-transplant context and concerning skin lesions. Initially considers
BPDCN but also many other possibilities. Partially correct working
diagnosis.

Thinking trace (step 14):

{[}\ldots{]} noclonal proliferation pattern and the time from transplant
(belumosudil suggests chronic GVHD, so likely \textgreater1 year
post-transplant), this could be lymphoma relapse.

But actually, looking at the specific combination of features: -
Violaceous nodules - Pancytopenia - Splenomegaly - Possible CNS
involvement (headaches, vision changes)

This could be blastic plasmacytoid dendritic cell neoplasm (BPDCN) which
often presents with skin lesions (bruise-like violaceous nodules), and
can involve bone marrow, lymph nodes, and CNS. The patient previously
had BPDCN given the context clues.

Given the context in the first message said ``Post-transplant Patient''
and the medication belumosudil (which is specifically for chronic GVHD),
and the need to check ``prior BPDCN therapies'' in the treatment history
description, the patient very likely has recurrent/refractory BPDCN.

However, I need to be careful. I could proceed to solve based on the
likely diagnosis of relapsed BPDCN or disease progression. But I should
really get the skin biopsy to confirm.

Since I'm told skin\_biopsy is available in round 2, let me try
requesting it one more time or ask to proceed. Actually, I think the
system is guiding me that I need to wait or request something else.

Let me check if there are any other available items I haven't requested.
I could request ``treatment\_history'' for round 3, but that's later. Or
I could request ``demographics'' (though I have age/sex).

Actually, looking back at the instructions, it says: ``Some tests may
not be available yet (e.g., pathology results in early rounds). If this
occurs, you can: Request different available information OR Proceed with
your assessment based on current data.''

Given that I have substantial clinical information and the available
tests are telling me to wait for specific pathology, I could try to
solve now with a working diagnosis based on clinical presentation.

Working diagnosis: Relapsed or refractory Blastic Plasmacytoid Dendritic
Cell Neoplasm {[}\ldots{]}

\textbf{Case 6} \textbar{} BPDCN \textbar{} Round 1 \textbar{} GPT-OSS
120B

Severity: low

\textbf{Evaluator:} Extremely thorough but broad. Does not initially
consider BPDCN. Systematic imaging and lab requests.

\textbf{Error detail:} Extremely thorough but broad. Does not initially
consider BPDCN. Systematic imaging and lab requests.

Thinking trace (step 10):

{[}\ldots{]} Skin lesions violaceous nodules with indurated could be
leukemia cutis, cutaneous T-cell lymphoma, or vasculitis.

Given the systemic symptoms, organomegaly, lymphadenopathy, CNS
involvement (neurologic signs), likely a disseminated PTLD, possibly
monomorphic PTLD (e.g., diffuse large B-cell lymphoma) or EBV-negative.
Could also be cutaneous infiltration by leukemia.

But we haven't seen immunophenotype. However we have some labs: calcium
high suggests possible bone involvement or paraneoplastic. The patient
is on belunosudil (for GVHD) and also on prophylaxis (co-trimoxazole,
valaciclovir) - could be EBV reactivation.

We have not yet requested any imaging beyond CT; we have CT that shows
lymphadenopathy, pericardial and pleural effusion, spleen enlarged,
cutaneous lesions.

We could request ``serum LDH'' which we have; maybe ``urine Bence
Jones''? Not relevant.

Maybe request ``immunophenotyping of peripheral blood'' but not
available. Could request ``EBV viral load'' but not available.

We might need ``bone marrow biopsy'' but not available yet.

Given the constraints, perhaps we should request ``abdominal
ultrasound'' to evaluate liver/spleen, but we likely already have CT.

The next logical step may be ``biopsy of one of the lymph nodes'' or
``skin biopsy.'' But those are not available.

Maybe ``CSF analysis'' but not available.

Given limited options, maybe we need to solve with current info. The
most likely diagnosis: Post-transplant lymphoproliferative disorder
(PTLD), possibly diffuse large B-cell lymphoma, presenting with skin
involvement, systemic symptoms, cytopenias.

Differentials: Relapsed primary disease (e.g., AML recurrence),
graft-versus-host disease cutaneous manifestations, drug-induced skin
lesions, infection (e.g., disseminated fungal, mycobacterial), cutaneous
leukemia infiltration.

Treatment plan: urgent tissue diagnosis (biopsy of skin lesion or lymph
node). {[}\ldots{]}

\textbf{Case 2} \textbar{} AML \textbar{} Round 1 \textbar{} GPT-OSS
120B

Severity: low

\textbf{Evaluator:} Extremely thorough but unfocused. Does not
prioritize myeloid sarcoma. Broad differential appropriate but needs
stratification.

\textbf{Error detail:} Extremely thorough but unfocused. Does not
prioritize myeloid sarcoma. Broad differential appropriate but needs
stratification.

Thinking trace (step 15):

{[}\ldots{]} She has new skin changes on back and extremities,
intermittent headaches nausea, chronic joint pain stiffness. Physical
exam not described. Labs normal CBC, chemistry normal, NT-proBNP
elevated 1250, troponin slightly high maybe. Echo shows pericardial
thickening and septal hypertrophy suspicious mass lesion. Could be
graft-versus-host disease (skin manifestation) but pericardial
involvement? Could be chronic GVHD with scleroderma-like features (skin
thickening, joint stiffness), also pericardial fibrosis. Could be a
relapse of underlying disease, perhaps acute myeloid leukemia? But labs
normal.

We need to request something available now. Imaging we have echo. Could
request abdominal ultrasound? Might show organomegaly. But perhaps best
is to request ``peripheral smear'' though CBC done; maybe need ``viral
PCR for CMV, EBV''. But not available. The only remaining available
tests: maybe ``urinalysis''. Could request ``urinalysis''. However we
need to arrive to solve eventually. Maybe we have enough: suspect
chronic GVHD with sclerodermatous changes, pericardial involvement.
Treatment: increase systemic steroids, consider extracorporeal
photopheresis, add mycophenolate. Also need to rule out infection. So I
can solve now.

We need to list working diagnosis, differentials, treatment plan with
actions (therapeutic, diagnostic, supportive). I think we can solve.

Let's provide:

working\_diagnosis: chronic graft-versus-host disease (cGVHD) with
sclerodermatous skin involvement and pericardial fibrosis.

differentials: acute GVHD, drug reaction, autoimmune connective tissue
disease (systemic sclerosis), infection (viral pericarditis), relapse of
underlying disease.

treatment plan: increase systemic corticosteroids, consider
extracorporeal photopheresis, add mycophenolate mofetil, continue
ruxolitinib, cardiac evaluation with cardiac MRI (diagnostic),
supportive: continue prophylactic antimicrobials, manag {[}\ldots{]}

\subsection{R-IDEA Dimension Analysis}\label{r-idea-dimension-analysis}

Each R-IDEA dimension is illustrated with two low-scoring and two
high-scoring examples. R-IDEA dimensions are scored on different scales:
Interpretation (I) on 0--4, and Differential Diagnosis (D), Exploration
(E), and Alternatives (A) on 0--2.

\subsubsection{Interpretation (I)}\label{interpretation-i}

The quality of clinical data interpretation and synthesis.

\subsubsection{Low-Scoring Examples}\label{low-scoring-examples}

\textbf{Case 8} \textbar{} AML \textbar{} Round 1 \textbar{} Kimi K2
Thinking

\textbf{Evaluator:} Poor R1 reasoning. The model immediately solves with
septic shock from CAP without requesting any labs, missing the acute
leukemia diagnosis entirely. While the sepsis management plan is
clinically appropriate, the failure to consider a hematologic etiology
for the presentation is a critical gap. No differential diagnosis is
generated. The model does not request CBC (which would have revealed
blasts and severe cytopenias) before solving.

\textbf{Thinking trace (step 3):}

{[}\ldots{]} Invasive procedures and complex diagnostic evaluations
usually follow once the patient is at least partially stabilized,''
maybe they want us to ask for a specific investigation, like ``Chest
X-ray,'' which is emergent non-invasive imaging. Or ``blood lactate'' to
quantify severity. Or ``urine output.'' Or ``point-of-care ultrasound.''
But the question is: what single information item would be most helpful?
The patient is in septic shock. The immediate priority is to identify
the source and start broad-spectrum antibiotics. So we might want to
know the chest X-ray. But we could also ask for ``blood culture
results,'' but those take time. The best immediate test is a lactate
level to assess tissue perfusion. But the patient is already clearly in
shock. The Surviving Sepsis guidelines: measure lactate, get blood
cultures, give broad-spectrum antibiotics, give 30 mL/kg IV
crystalloids, and start vasopressors if MAP \textless{} 65. So the
``single information item'' could be ``What is the serum lactate?'' But
we can also ``solve'' by giving the management.

I think the best answer is to ``solve'' because the case is clear and
immediate action is needed. The question might be testing if the user
recognizes septic shock and knows the immediate steps. So I'll provide a
comprehensive management plan: immediate resuscitation, antibiotics,
source control, vasopressors, etc.

But the instruction says to ``provide either a request for a single
information item or solve.'' So I must choose one. Let's see the typical
style of these cases: The user wants us to think step by step. The
initial presentation is given, and we can ask for a single piece of
information to help narrow down. For instance, we could ask ``What is
the chest X-ray result?'' because that would confirm pneumonia as the
source. Or ``What is the lactate level?'' to gauge severity.
{[}\ldots{]}

\textbf{Case 7} \textbar{} DLBCL \textbar{} Round 1 \textbar{} Claude
Opus 4.6

\textbf{Evaluator:} Limited initial differential focused on Kaposi
sarcoma based on pigmented lesion and immunosuppression. Missed lymphoma
as leading possibility despite clinical clues. Medication inference was
appropriate but diagnosis was incorrect.

Thinking trace (step 5):

{[}\ldots{]} Intermittent low-grade fevers at night 3. Medications
suggest significant immunosuppression: - Valaciclovir (HSV/VZV
prophylaxis) - Co-trimoxazole (PCP prophylaxis) - given Mon/Wed/Fri -
Apixaban (anticoagulation - possibly for prior thrombosis) - These
prophylactic medications suggest the patient is on immunosuppressive
therapy, possibly for a hematologic malignancy, organ transplant, or
autoimmune condition 4. The CBC reveals pancytopenia with anemia,
leukopenia, and thrombocytopenia, plus nuclear debris that could
indicate chronic lymphocytic leukemia. Chemistry labs show hyponatremia,
elevated liver enzymes and LDH, and hyperuricemia---a pattern consistent
with malignancy or significant systemic disease. 5. The clinical
picture---immunosuppression evidenced by PCP and HSV prophylaxis,
pancytopenia, a dark nodular skin lesion, and low-grade fevers---points
toward several possibilities including Kaposi sarcoma (especially HHV-8
associated given the immunosuppressed state), cutaneous lymphoma, or
melanoma, though the systemic findings suggest something more than a
primary skin malignancy. The prophylactic medications themselves
indicate this patient is already recognized as significantly
immunocompromised.

The specific prophylaxis regimen (co-trimoxazole on alternating days for
PCP, valaciclovir for HSV/VZV) is typical for post-transplant patients,
those undergoing chemotherapy for hematologic malignancies, or
HIV-positive individuals. The 2\% nuclear debris on the CBC differential
could represent smudge cells characteristic of chronic lymphocytic
leukemia, suggesting this might be a CLL patient who's been treated and
is now immunosuppressed, potentially developing a secondary malignancy
like Kaposi sarcoma.

I need to get a peripheral blood smear or flow cytometry to better
characterize the hematologic abnormality, and imaging or a skin biopsy
of the ankle lesion would help determine if this {[}\ldots{]}

\subsubsection{High-Scoring Examples}\label{high-scoring-examples}

\textbf{Case 11} \textbar{} DLBCL \textbar{} Round 3 \textbar{} Claude
Haiku 4.5

\textbf{Evaluator:} Correctly identifies double-hit lymphoma after FISH
results. Good treatment recommendations including intensive chemotherapy
and CNS prophylaxis. Comprehensive final synthesis of the case.

Thinking trace (step 14):

{[}\ldots{]} The partial CD20 and strong BCL2 with this morphology might
point toward a different entity altogether.

The strong CD45+, CD79a+, and robust PAX5 expression confirm this is
B-cell lineage, not Hodgkin lymphoma. Given the high Ki-67 (60-70\%),
strong BCL2 and MUM1 positivity, CD30 expression in 40\% of cells, and
the clinical presentation with bulky mediastinal disease and SUVmax
18.4, I'm considering whether this represents a high-grade B-cell
lymphoma, possibly with double-expressor features or a variant DLBCL.

The mediastinal involvement and CD30 positivity raise the question of
primary mediastinal B-cell lymphoma, though the cervical involvement and
strong B-cell markers suggest a more typical DLBCL. The BCL6 variability
and CD15 focal expression don't clearly point to one diagnosis, but the
aggressive biology and immunophenotype pattern are more consistent with
DLBCL than with an overlap or Hodgkin-like process. BCL6) should be
available now since I requested them in Round 2. Looking back at the
feedback, the FISH/NGS for translocations was marked ``correct'' for
intention but only ``partially\_correct'' for priority---suggesting I
ordered the right test but perhaps at the wrong time or with wrong
urgency.

What's striking is that nearly every element of my Round 2 treatment
plan was flagged as incorrect: airway monitoring, R-CHOP initiation, TLS
prophylaxis, and prophylactic IT chemotherapy all had wrong intentions
and priorities. This pattern suggests I'm fundamentally misunderstanding
the diagnosis or the clinical urgency. The molecular results should
clarify whether this is a high-grade B-cell lymphoma or something else
entirely that would change the management approach. The biopsy
description of a polymorphous infiltrate with occasional multinucleated
forms, combined with the immunophenotype---strong PAX5, partial CD20,
variable BCL6/MUM1, strong BCL2, moderate CD30 positivity, an
{[}\ldots{]}

\textbf{Case 1} \textbar{} AML \textbar{} Round 3 \textbar{} Kimi K2.5
Thinking

\textbf{Evaluator:} Comprehensive molecular integration identifying
FLT3-ITD adverse risk. Good cardiac-aware treatment planning. Misses
relapsed nature and WHO 2022 classification.

Thinking trace (step 6):

{[}\ldots{]} aches for unfit AML with FLT3-ITD: - Venetoclax + HMA
(decitabine/azacitidine) - increasingly used in unfit patients, though
not specifically approved for FLT3-ITD - Clinical trial with FLT3
inhibitor + lower intensity regimen - Gilteritinib is approved for
relapsed/refractory FLT3-mutated AML, not frontline

However, given the severe cardiac issues and active infection, I need to
prioritize: 1. Control infection completely (source control for abscess,
appropriate antibiotics) 2. Cardiac optimization 3. Then consider lower
intensity chemotherapy regimen

Wait - I should also check if HMA + venetoclax is appropriate here, or
if there's a specific FLT3-targeted approach.

Actually, for frontline FLT3-mutated AML in patients unfit for intensive
chemotherapy, the standard is challenging. Midostaurin requires
concurrent chemotherapy. There are ongoing studies of FLT3 inhibitors
with HMAs or venetoclax.

Given this is a test case, I should consider what the expected answer
is. The patient has: - FLT3-ITD AML (adverse risk) - Cardiac dysfunction
(contraindication to anthracyclines) - Active infection - History of
non-compliance

This is a high-risk, challenging case. The definitive diagnosis should
include the FLT3-ITD mutation status.

For treatment in this specific scenario: - The patient cannot receive
standard 7+3 due to cardiac issues and prior anthracycline exposure -
The infection must be controlled first - Consider venetoclax +
decitabine (5+2 or 10-day decitabine) or clinical trial - Add
midostaurin if using chemotherapy, but that's risky - Alternatively,
consider gemtuzumab ozogamicin (CD33 antibody) - but CD33 status
unknown, and GO is for CD33+ AML, approved for frontline in combination
or as single agent in some settings, but mainly used with chemotherapy
or in APL.

Wait, I need to check the flow cytometry again - CD33 was negative! So
gemtuzumab ozogamicin would not be appropriate (it targets CD33).

Given CD33 negativity, FLT3-ITD, cardiac issues, active in {[}\ldots{]}

\subsubsection{Differential Diagnosis
(D)}\label{differential-diagnosis-d}

Breadth and appropriateness of the differential diagnosis.

\subsubsection{Low-Scoring Examples}\label{low-scoring-examples-1}

\textbf{Case 1 \textbar{} AML \textbar{} Round 2 \textbar{} Claude Opus
4.5}

\textbf{Evaluator:} Thorough integration of bone marrow morphology flow
cytometry and cytogenetics confirming AML-M4. Good infectious workup
interpretation with ESBL-KPC Klebsiella. Limited alternative diagnosis
consideration (no MPAL or CML blast phase mentioned).

Thinking trace (step 14):

{[}\ldots{]} lymphocytes - Chemistry: elevated creatinine (2.4),
elevated LDH (1289), elevated uric acid (12.9), hypercalcemia (2.65),
elevated lactate (4.3), elevated NT-proBNP (4612) - Coagulation:
prolonged aPTT (42s), elevated INR (2.1) - coagulopathy - CT chest:
segmental PE in right lower lobe - Echo: MVP, TR II/III°, EF 54\%, TAPSE
10mm (RV dysfunction), FAC 30\%

Bone marrow/Flow cytometry: - Hypercellular BM with 50\% myeloid blasts
+ 40\% monocytic blasts (90\% total blasts) - Auer rods present - Flow:
CD34+, CD117+, HLA-DR+, MPO+, CD11b+, CD13+; CD33-, CD7-, CD56-, TdT-

Monocytic component shows CD14+/CD64+ expression with aberrant CD56,
normal karyotype 46,XY

Infectious workup reveals elevated inflammatory markers (CRP 32, PCT
4.1) suggesting concurrent bacterial infection, though blood cultures
remain negative. Galactomannan is negative, but respiratory testing
detected enterovirus/rhinovirus, and colonization with ESBL-producing
KPC+ Klebsiella pneumoniae. The morphologic and immunophenotypic
findings are consistent with Acute Myelomonocytic Leukemia (AML-M4 by
FAB classification).

The clinical picture shows hyperleukocytosis with WBC \textgreater100,
evidence of tumor lysis syndrome with elevated LDH, uric acid, and
creatinine alongside hyperkalemia and hypercalcemia. There's also
coagulopathy, pulmonary embolism with right ventricular dysfunction, and
an active soft tissue infection at the left antecubital fossa. The
ESBL-KPC colonization status is critical for guiding antibiotic therapy.
I should consider whether molecular testing results are available to
further stratify risk, as this would help determine prognosis and
treatment intensity.

Given the acuity of the presentation, I need to act on what I have now.
The immediate priorities are managing the leukostasis and tumor lysis
complications, addressing the thromboembolism, controlling the infection
with appropriate coverage for resistant organisms, and providing
supportive care. {[}\ldots{]}

\textbf{Case 6} \textbar{} BPDCN \textbar{} Round 1 \textbar{} GLM 4.7

\textbf{Evaluator:} R1 showed good systematic approach to a post-alloSCT
patient with skin nodules and CNS symptoms. The model correctly
identified the immunosuppressed state and generated a broad
differential. However, it prioritized fungal infections (cryptococcosis,
nocardiosis) over hematologic malignancy relapse, which was marked
`partially\_correct'. The differential did include PTLD and leukemia
cutis which were marked correct.

Thinking trace (step 3):

{[}\ldots{]} Formulate the ``Solve'' response:The previous ``solve''
attempt failed schema validation. I need to adhere strictly to the
schema. Common Schema fields for ``Solve'': \texttt{working\_diagnosis}:
(String) \texttt{differential\_diagnosis}: (Array of strings)
\texttt{plan}: (Array of objects with \texttt{action}, \texttt{test},
\texttt{priority}, etc. - this varies by prompt instructions, but
usually structured). Correction: The prompt says ``Provide either a
request for a single information item or solve.'' It doesn't give me the
specific schema for ``Solve''. However, looking at the previous error,
the system expects a specific structure. The error said:
\texttt{\textquotesingle{}is\ not\ valid\ under\ any\ of\ the\ given\ schemas}.
The previous attempt used keys like \texttt{treatment\_plan},
\texttt{differentials}, \texttt{working\_diagnosis}. Let's look at the
standard schemas usually used in these types of prompts or try to infer
from the error. Common schemas for this specific ``Doctor AI''
simulation often look like:

\begin{quote}
\begin{verbatim}
{
  "response": {
    "action": "solve",
    "solve": {
      "working_diagnosis": "...",
      "differential_diagnosis": ["..."],
      "immediate_plan": [...] // or "actions"
    }
  }
}
\end{verbatim}
\end{quote}

Or maybe analysis and plan.

Let's look at the specific constraints: ``Invasive tissue
sampling\ldots{} usually require further coordination''.

Let's refine the diagnosis. The combination of \textbf{violaceous
nodules}, \textbf{CNS symptoms}, and \textbf{immunosuppression} screams
\textbf{Cryptococcosis} or \textbf{Nocardiosis} or \textbf{Lymphoma
(PTLD)}. Cryptococcus skin lesions can mimic cellulitis, nodules, or
molluscum. PTLD can present as skin nodules and CNS lymphoma. However,
the ``painful'' nature might point more to infection (fungal/bacterial)
than lymphoma (often painless). But not always.

Let's construct the plan. 1. {[}\ldots{]}

\subsubsection{High-Scoring Examples}\label{high-scoring-examples-1}

\textbf{Case 1} \textbar{} AML \textbar{} Round 1 \textbar{} Claude
Haiku 4.5

\textbf{Evaluator:} Efficient identification of AML-M5/M4 with good
prioritization of CBC first. Systematic TLS and DIC assessment. Slightly
less detailed pathophysiology than larger models but clinically sound
reasoning throughout.

Thinking trace (step 2):

{[}\ldots{]} 10\textsuperscript{9}/L (extreme leukocytosis) - Plt 12
×10\textsuperscript{9}/L (severe thrombocytopenia) - Normal MCV and MCH
- Differential: 30\% blasts, 60\% monocytes, 8\% neutrophils, 2\%
lymphocytes

Coagulation: - aPTT 42 sec (prolonged - normal is \textasciitilde25-35
sec) - INR 2.1 (prolonged - normal is 0.8-1.1) - Fibrinogen 2.1 g/L (low
- normal is 2-4 g/L) - D-dimer normal (argues against DIC as primary
problem, though could still have early DIC)

\textbf{Red Flags:} 1. The combination of anemia, extreme leukocytosis
with 30\% blasts, and severe thrombocytopenia is highly suspicious for
acute leukemia 2. The monocyte predominance (60\%) with 30\% blasts
suggests AML, possibly monocytic differentiation (M4/M5 subtypes) 3.
Coagulopathy (prolonged PT/aPTT, low fibrinogen) in the setting of
leukemia with monocytic features is concerning for disseminated
intravascular coagulation (DIC), particularly associated with acute
monocytic leukemia (AML-M5)

The hemodynamic instability and respiratory compromise compound the
clinical urgency

This presentation is consistent with acute monocytic leukemia
complicated by DIC. The extreme leukocytosis with monocytic
predominance, coupled with the coagulopathy and hemodynamic collapse,
points toward AML-M5 as the primary diagnosis. The low fibrinogen and
prolonged clotting times in this context suggest consumptive
coagulopathy rather than simple hepatic dysfunction.

I need to assess for leukostasis given the markedly elevated WBC count,
which could explain the respiratory distress and hypotension. A chest
X-ray would help differentiate leukostatic infiltration from other
pulmonary causes. I should also obtain LDH, uric acid, and renal
function to evaluate tumor lysis syndrome risk, and a peripheral blood
smear would clarify the blast morphology and confirm monocytic
differentiation. While bone marrow biopsy would definitively establish
the diagnosis, the clinical picture is sufficiently characteristic that
treatment decisions may need to proceed based on the current findings.
{[}\ldots{]}

\textbf{Case 5} \textbar{} AML \textbar{} Round 1 \textbar{} Gemini 3.1
Pro

\textbf{Evaluator:} Adequate R1: correctly identifies acute leukemia
from 78\% blasts, considers AML with monocytic features (gingival
hyperplasia, monocytosis). Notes hepatosplenomegaly, TLS risk markers.
However, reasoning is repetitive across 8 entries without deepening
synthesis. Does not connect posaconazole to prior myeloma treatment
history. Differential of AML vs ALL is appropriate but not well
stratified.

Thinking trace (step 2):

Assessing The Data

I'm currently focused on the initial data. A 56-year-old female presents
with fatigue, pallor, and dyspnea, alongside concerning signs like
spontaneous bruising, petechiae, and weight loss. The physical exam
confirms pallor and scattered ecchymoses. My immediate analysis involves
assessing the significance of these symptoms collectively, prioritizing
potential differential diagnoses.

Analyzing The Clinical Picture

I've got a clearer picture now: petechiae, gingival hyperplasia,
hepatosplenomegaly, and cytopenias with 78\% blasts. The medication list
including posaconazole is interesting, which I must note is perhaps from
prophylaxis. Elevated LDH, uric acid, and mildly deranged coags point
towards possible DIC or liver issues. My focus has shifted toward AML
with monocytic features (FAB M4/M5) given the gingival hyperplasia and
monocytic differentiation. Bone marrow biopsy is crucial but delayed.

Deciding The Next Step

I've determined I'm at a critical juncture. The game mechanics are a bit
unclear, but it seems I should ``solve'' to advance. The system's
response to re-requesting the bone marrow, and the explicit prompt
language, suggests I should be proceeding based on my existing data.
I've formulated a working diagnosis of AML with monocytic
differentiation. Now, I have formulated a few differential diagnoses
that will now be part of my assessments.

\subsubsection{Exploration (E)}\label{exploration-e}

Depth of investigation and information-seeking behavior.

\subsubsection{Low-Scoring Examples}\label{low-scoring-examples-2}

\textbf{Case 2} \textbar{} AML \textbar{} Round 2 \textbar{} Qwen3 235B
A22B

\textbf{Evaluator:} Recovered well from incorrect R1 diagnosis after
skin biopsy revealed myeloid sarcoma with NPM1 mutation. Struggled with
information gathering strategy initially but correctly identified AML
with extramedullary involvement once pathology was available.

Thinking trace (step 4):

{[}\ldots{]} So maybe ruxolitinib isn't the cause, and stopping it was
wrong. The intention was diagnostic, but the priority was wrong. So
perhaps the main issue is co-trimoxazole.

Co-trimoxazole is taken three times a week, and the rash might
correlate. The rash could be a maculopapular rash, which is common with
sulfa drugs. If the rash started after starting co-trimoxazole, that's a
clue.

Also, the patient has chronic joint pain and stiffness, which might be
related to an autoimmune condition. If she has psoriasis, ruxolitinib
can be used to treat it, but maybe the rash is a flare.

But the previous solve didn't have the rash morphology. So in Round 2,
the first step should be to get the rash description. However, the user
might have that info now. Wait, looking back, in the History Physical,
it says: ``new skin changes, first noticed by her partner on the back
and now also on the extremities.'' But no morphology.

The previous assistant requests asked for rash description, but the user
hasn't provided it yet. So in Round 2, the correct step is to request
the rash morphology and distribution. But the capabilities now include
tissue sampling, so maybe a skin biopsy is appropriate.

But the evaluation in Round 1 said the treatment plan had some actions
incorrect because they stopped ruxolitinib unnecessarily. So perhaps the
rash is not drug-related to ruxolitinib but to co-trimoxazole.

The key differentials now: co-trimoxazole reaction (SJS/TEN,
maculopapular), viral (like HHV-6, which can cause rash and
myocarditis), or autoimmune.

Elevated NT-proBNP and troponin could be from myocarditis, possibly due
to viral infection. HHV-6 can cause both rash and myocarditis. The
patient is on immunosuppressants, so viral reactivation is possible.

The patient's symptoms: headaches, nausea, skin changes, tachycardia. If
it's HHV-6, that could explain all. So testing for viral pathogens might
be needed.

But the current task is to either request info or solve. {[}\ldots{]}

\textbf{Case 15} \textbar{} Multiple Myeloma \textbar{} Round 1
\textbar{} GLM 5

\textbf{Evaluator:} Reasonable R1 assessment of a myeloma patient with
pancytopenia and new vertebral compression fractures. Correctly
identified the key differentials: myeloma progression, therapy-related
MDS, and drug-induced myelosuppression. Appropriately noted the
macrocytic anemia (MCV 98) as potentially suggesting MDS. Recognized the
significance of prophylactic antimicrobials indicating prior intensive
therapy. However, the model was very inefficient with 12 steps and
multiple invalid requests. The reasoning was somewhat repetitive,
cycling through the same differential list without significantly
advancing the clinical picture. Final treatment plan was reasonable but
lacked specificity.

\textbf{Thinking trace (step 11):}

{[}\ldots{]} But what about other things?

Actually, I think I should try requesting bone marrow biopsy again - it
might be available now that we're in a later round. Or I could try
``myeloma disease status summary'' or something like that.

Wait - looking at the instructions again, they say
``extended\_diagnostics' available in round 2'' - but I've been making
requests. Let me count the turns:

Turn 1: Initial presentation Turn 2: Requested CBC - got results Turn 3:
Requested myeloma markers - told not available in round 1 Turn 4:
Requested physical exam - got results Turn 5: Requested peripheral blood
smear - told CBC already provided Turn 6: Requested treatment history -
told available in round 3 Turn 7: Requested CT skeleton - got results
Turn 8: Requested myeloma protein studies again - told not available
Turn 9: Requested bone marrow biopsy - told not available in round 1
Turn 10: Requested SPEP - told not available in round 1 Turn 11:
Requested abdominal ultrasound - got results Turn 12: Requested SPEP
again - told not available

So we're in what round? The initial presentation was round 1. I've had
several exchanges. Looking at the system messages: - Round 1: Initial
presentation - ``extended\_diagnostics' available in round 2'' was
mentioned multiple times - ``Treatment history is available later in the
case (round 3)''

The problem is that each failed request might count as a turn, but
doesn't advance the round. So I might still be in round 1 despite many
exchanges.

Let me try a different approach. The message said ``Consider requesting
labs or imaging available in this round such as CBC/chemistry or
skeletal imaging'' - I already have both.

What about ``flow cytometry'' on peripheral blood? That might help
assess for circulating plasma cells or abnormal populations.

Or I could try ``blood culture'' if considering infection, but the
patient is afebrile with normal PCT.

Actually, I think the safest {[}\ldots{]}

\subsubsection{High-Scoring Examples}\label{high-scoring-examples-2}

\textbf{Case 1} \textbar{} AML \textbar{} Round 1 \textbar{} Claude
Haiku 4.5

\textbf{Evaluator:} Efficient identification of AML-M5/M4 with good
prioritization of CBC first. Systematic TLS and DIC assessment. Slightly
less detailed pathophysiology than larger models but clinically sound
reasoning throughout.

Thinking trace (step 2):

{[}\ldots{]} 10\textsuperscript{9}/L (extreme leukocytosis) - Plt 12
×10\textsuperscript{9}/L (severe thrombocytopenia) - Normal MCV and MCH
- Differential: 30\% blasts, 60\% monocytes, 8\% neutrophils, 2\%
lymphocytes

Coagulation: - aPTT 42 sec (prolonged - normal is \textasciitilde25-35
sec) - INR 2.1 (prolonged - normal is 0.8-1.1) - Fibrinogen 2.1 g/L (low
- normal is 2-4 g/L) - D-dimer normal (argues against DIC as primary
problem, though could still have early DIC)

\textbf{Red Flags:} 1. The combination of anemia, extreme leukocytosis
with 30\% blasts, and severe thrombocytopenia is highly suspicious for
acute leukemia 2. The monocyte predominance (60\%) with 30\% blasts
suggests AML, possibly monocytic differentiation (M4/M5 subtypes) 3.
Coagulopathy (prolonged PT/aPTT, low fibrinogen) in the setting of
leukemia with monocytic features is concerning for disseminated
intravascular coagulation (DIC), particularly associated with acute
monocytic leukemia (AML-M5)

The hemodynamic instability and respiratory compromise compound the
clinical urgency

This presentation is consistent with acute monocytic leukemia
complicated by DIC. The extreme leukocytosis with monocytic
predominance, coupled with the coagulopathy and hemodynamic collapse,
points toward AML-M5 as the primary diagnosis. The low fibrinogen and
prolonged clotting times in this context suggest consumptive
coagulopathy rather than simple hepatic dysfunction.

I need to assess for leukostasis given the markedly elevated WBC count,
which could explain the respiratory distress and hypotension. A chest
X-ray would help differentiate leukostatic infiltration from other
pulmonary causes. I should also obtain LDH, uric acid, and renal
function to evaluate tumor lysis syndrome risk, and a peripheral blood
smear would clarify the blast morphology and confirm monocytic
differentiation. While bone marrow biopsy would definitively establish
the diagnosis, the clinical picture is sufficiently characteristic that
treatment decisions may need to proceed based on the current findings.
{[}\ldots{]}

\textbf{Case 11} \textbar{} DLBCL \textbar{} Round 3 \textbar{} Claude
Haiku 4.5

\textbf{Evaluator:} Correctly identifies double-hit lymphoma after FISH
results. Good treatment recommendations including intensive chemotherapy
and CNS prophylaxis. Comprehensive final synthesis of the case.

Thinking trace (step 2):

{[}\ldots{]} The partial CD20 and strong BCL2 with this morphology might
point toward a different entity altogether.

The strong CD45+, CD79a+, and robust PAX5 expression confirm this is
B-cell lineage, not Hodgkin lymphoma. Given the high Ki-67 (60-70\%),
strong BCL2 and MUM1 positivity, CD30 expression in 40\% of cells, and
the clinical presentation with bulky mediastinal disease and SUVmax
18.4, I'm considering whether this represents a high-grade B-cell
lymphoma, possibly with double-expressor features or a variant DLBCL.

The mediastinal involvement and CD30 positivity raise the question of
primary mediastinal B-cell lymphoma, though the cervical involvement and
strong B-cell markers suggest a more typical DLBCL. The BCL6 variability
and CD15 focal expression don't clearly point to one diagnosis, but the
aggressive biology and immunophenotype pattern are more consistent with
DLBCL than with an overlap or Hodgkin-like process. BCL6) should be
available now since I requested them in Round 2. Looking back at the
feedback, the FISH/NGS for translocations was marked ``correct'' for
intention but only ``partially\_correct'' for priority---suggesting I
ordered the right test but perhaps at the wrong time or with wrong
urgency.

What's striking is that nearly every element of my Round 2 treatment
plan was flagged as incorrect: airway monitoring, R-CHOP initiation, TLS
prophylaxis, and prophylactic IT chemotherapy all had wrong intentions
and priorities. This pattern suggests I'm fundamentally misunderstanding
the diagnosis or the clinical urgency. The molecular results should
clarify whether this is a high-grade B-cell lymphoma or something else
entirely that would change the management approach. The biopsy
description of a polymorphous infiltrate with occasional multinucleated
forms, combined with the immunophenotype---strong PAX5, partial CD20,
variable BCL6/MUM1, strong BCL2, moderate CD30 positivity, an
{[}\ldots{]}

\subsubsection{Alternatives (A)}\label{alternatives-a}

Consideration and comparison of alternative diagnoses.

\subsubsection{Low-Scoring Examples}\label{low-scoring-examples-3}

\textbf{Case 7} \textbar{} DLBCL \textbar{} Round 1 \textbar{} Claude
Opus 4.6

\textbf{Evaluator:} Limited initial differential focused on Kaposi
sarcoma based on pigmented lesion and immunosuppression. Missed lymphoma
as leading possibility despite clinical clues. Medication inference was
appropriate but diagnosis was incorrect.

Thinking trace (step 4):

{[}\ldots{]} Intermittent low-grade fevers at night 3. Medications
suggest significant immunosuppression: - Valaciclovir (HSV/VZV
prophylaxis) - Co-trimoxazole (PCP prophylaxis) - given Mon/Wed/Fri -
Apixaban (anticoagulation - possibly for prior thrombosis) - These
prophylactic medications suggest the patient is on immunosuppressive
therapy, possibly for a hematologic malignancy, organ transplant, or
autoimmune condition 4. The CBC reveals pancytopenia with anemia,
leukopenia, and thrombocytopenia, plus nuclear debris that could
indicate chronic lymphocytic leukemia. Chemistry labs show hyponatremia,
elevated liver enzymes and LDH, and hyperuricemia---a pattern consistent
with malignancy or significant systemic disease. 5. The clinical
picture---immunosuppression evidenced by PCP and HSV prophylaxis,
pancytopenia, a dark nodular skin lesion, and low-grade fevers---points
toward several possibilities including Kaposi sarcoma (especially HHV-8
associated given the immunosuppressed state), cutaneous lymphoma, or
melanoma, though the systemic findings suggest something more than a
primary skin malignancy. The prophylactic medications themselves
indicate this patient is already recognized as significantly
immunocompromised.

The specific prophylaxis regimen (co-trimoxazole on alternating days for
PCP, valaciclovir for HSV/VZV) is typical for post-transplant patients,
those undergoing chemotherapy for hematologic malignancies, or
HIV-positive individuals. The 2\% nuclear debris on the CBC differential
could represent smudge cells characteristic of chronic lymphocytic
leukemia, suggesting this might be a CLL patient who's been treated and
is now immunosuppressed, potentially developing a secondary malignancy
like Kaposi sarcoma.

I need to get a peripheral blood smear or flow cytometry to better
characterize the hematologic abnormality, and imaging or a skin biopsy
of the ankle lesion would help determine if this {[}\ldots{]}

\textbf{Case 2} \textbar{} AML \textbar{} Round 1 \textbar{} Qwen3 235B
A22B

\textbf{Evaluator:} Good identification of post-transplant context from
medications. However, anchored on drug reaction hypothesis despite
findings suggesting more systemic process. Round 1 diagnosis of
ruxolitinib-induced hypersensitivity was incorrect.

Thinking trace (step 1):

{[}\ldots{]} Stress-induced: like urticaria.

But without the appearance, it's very hard.

Another point: the patient has chronic joint pain and stiffness. This,
along with skin changes, could be inflammatory arthritis with skin
manifestations (e.g., psoriatic arthritis, but we don't have psoriasis
described).

Given the tachycardia and the stress reaction, it's possible that the
rash is stress-related (e.g., acute urticaria). Urticaria is often
pruritic and presents as wheals.

But we don't know.

The vital signs: HR 116 is high. This could be due to anxiety, but also
could be a sign of systemic illness.

We also have intermittent headaches and mild nausea, which could be
related to the stress or something else.

Given the context of immunosuppression, we must be cautious of
infections.

However, the patient is afebrile and in good general condition.

The most critical step is to describe the rash.

But the problem says: ``You are evaluating a clinically stable patient
\ldots{} You can perform a focused history and examination''. So we can
do a skin exam and describe it.

But the assistant is to either request a single information item or
solve.

Given that the history physical provided does not include the skin
description we asked for, and it's essential, I think we should request
it again? But the problem says ``single information item'', and we
already asked for it. However, it wasn't provided in the history
physical.

Alternatively, we can make an educated guess based on the medications.

But note: the patient is on ruxolitinib, and a known side effect is
herpes zoster. The rash of herpes zoster is typically painful,
vesicular, in a dermatomal distribution. The patient has skin changes on
the back and extremities. If it's multiple dermatomes, it could be
disseminated zoster. {[}\ldots{]}

\subsubsection{High-Scoring Examples}\label{high-scoring-examples-3}

\textbf{Case 1} \textbar{} AML \textbar{} Round 1 \textbar{} Claude
Haiku 4.5

\textbf{Evaluator:} Efficient identification of AML-M5/M4 with good
prioritization of CBC first. Systematic TLS and DIC assessment. Slightly
less detailed pathophysiology than larger models but clinically sound
reasoning throughout.

Thinking trace (step 2):

{[}\ldots{]} 10\textsuperscript{9}/L (extreme leukocytosis) - Plt 12
×10\textsuperscript{9}/L (severe thrombocytopenia) - Normal MCV and MCH
- Differential: 30\% blasts, 60\% monocytes, 8\% neutrophils, 2\%
lymphocytes

Coagulation: - aPTT 42 sec (prolonged - normal is \textasciitilde25-35
sec) - INR 2.1 (prolonged - normal is 0.8-1.1) - Fibrinogen 2.1 g/L (low
- normal is 2-4 g/L) - D-dimer normal (argues against DIC as primary
problem, though could still have early DIC)

\textbf{Red Flags:} 1. The combination of anemia, extreme leukocytosis
with 30\% blasts, and severe thrombocytopenia is highly suspicious for
acute leukemia 2. The monocyte predominance (60\%) with 30\% blasts
suggests AML, possibly monocytic differentiation (M4/M5 subtypes) 3.
Coagulopathy (prolonged PT/aPTT, low fibrinogen) in the setting of
leukemia with monocytic features is concerning for disseminated
intravascular coagulation (DIC), particularly associated with acute
monocytic leukemia (AML-M5) 4. The hemodynamic instability and
respiratory compromise compound the clinical urgency

This presentation is consistent with acute monocytic leukemia
complicated by DIC. The extreme leukocytosis with monocytic
predominance, coupled with the coagulopathy and hemodynamic collapse,
points toward AML-M5 as the primary diagnosis. The low fibrinogen and
prolonged clotting times in this context suggest consumptive
coagulopathy rather than simple hepatic dysfunction.

I need to assess for leukostasis given the markedly elevated WBC count,
which could explain the respiratory distress and hypotension. A chest
X-ray would help differentiate leukostatic infiltration from other
pulmonary causes. I should also obtain LDH, uric acid, and renal
function to evaluate tumor lysis syndrome risk, and a peripheral blood
smear would clarify the blast morphology and confirm monocytic
differentiation. While bone marrow biopsy would definitively establish
the diagnosis, the clinical picture is sufficiently characteristic that
treatment decisions may need to proceed based on the current findings.
{[}\ldots{]}

\textbf{Case 5} \textbar{} AML \textbar{} Round 3 \textbar{} MiniMax
M2.1

\textbf{Evaluator:} Excellent R3 molecular integration. Correctly
interprets dual TP53 mutations (R248Q at 28\% VAF and Y220C at 12\%
VAF), PTPN11, and absence of FLT3/NPM1/CEBPA. Recognizes AML with TP53
mutation as a WHO 2022 entity. Treatment plan appropriately considers
venetoclax-based therapy, transplant evaluation, and clinical trials for
this high-risk disease. Good understanding of the poor prognosis
associated with biallelic TP53.

Thinking trace (step 2):

{[}\ldots{]} CBC: Hb 7.9 g/dL, WBC 13.1 ×10\textsuperscript{9}/L, Plt 18
×10\textsuperscript{9}/L; 78\% blasts 2. Bone marrow: - Hypocellular
(20\%) with 80\% blasts - Flow: CD34+, CD117+, HLA-DR+, CD13+, CD33+,
CD56+, CD7 negative - Fibrosis grade MF-2 3. Cytogenetics:
47,XX,+1,del(17p13.1),-20,dup(1q) - this is a complex karyotype 4.
Molecular: - TP53 p.R248Q (VAF 28\%) - TP53 p.Y220C (VAF 12\%) - PTPN11
(VAF 11\%) - No FLT3, NPM1, CEBPA mutations

Classification considerations:

The case shows: - TP53 mutations with high VAF (two mutations at 28\%
and 12\%) - Complex karyotype (3+ abnormalities) - Hypocellular bone
marrow with fibrosis - Monocytic features (CD56 expression, gingival
hyperplasia) - Hepatosplenomegaly

This could be classified as: 1. AML with TP53 mutation - The WHO 2022
classification recognizes AML with TP53 mutations as a distinct entity,
particularly when there's a complex karyotype or VAF \textgreater40\%.
The patient has two TP53 mutations (one likely clonal at 28\%), which
fits this category.2. AML with myelodysplasia-related changes
(MDS-related) - Given the dysplastic features, complex karyotype, and
TP53 mutations, this could also fit MDS-related AML.

I'll focus on the patient's age and treatment strategy. At 56, intensive
induction remains a viable option, but her TP53-mutated, complex
karyotype status suggests poor prognosis. I'll prioritize a
multi-pronged approach: aggressive induction, close monitoring for tumor
lysis, and early transplant evaluation. Clinical trial enrollment
becomes critical given the high-risk genetic profile. Targeted therapies
like venetoclax or experimental TP53-directed treatments could offer
potential benefits, especially if standard induction proves challenging.

Key supportive care interventions will include comprehensive transfusion
support, meticulous infection prophylaxis, and careful management of
coagulopathy. {[}\ldots{]}

\subsection{Exemplary Reasoning}\label{exemplary-reasoning}

These excerpts showcase the highest-quality clinical reasoning observed,
from models achieving perfect or near-perfect R-IDEA scores (total
\(\geq\) 9) with no identified cognitive errors.

7 excerpts from 6 models across 4 model families and 6 cases.

\textbf{Case 18} \textbar{} Aplastic Anemia \textbar{} Round 2
\textbar{} MiniMax M2.1

\textbf{Evaluator:} Outstanding R2 performance. The model correctly
identifies the critical finding of monosomy 7 in 18/20 metaphases and
immediately recognizes its implications: clonal evolution from aplastic
anemia to MDS, high risk of AML transformation, and need for urgent
allogeneic SCT evaluation. The PNH clone quantification (68\%
granulocytes, 72\% monocytes, 46\% RBCs) is correctly interpreted as a
large clone confirming PNH. The model makes the sophisticated
observation that eltrombopag may be associated with clonal evolution and
should be reconsidered - this is a nuanced clinical point. The
integration of infection data (negative cultures, KPC/VRE colonization,
residual fungal disease on PET-CT) provides comprehensive clinical
context. The reasoning about transplant urgency balanced against active
fungal infection is excellent.

Thinking trace (step 9):

{[}\ldots{]} \textbf{Cytogenetics:} - 46,XX{[}2{]} / 45,XX,-7{[}18{]} -
This is monosomy 7 in 18/20 cells (90\% of metaphases) - This is a VERY
concerning finding - monosomy 7 is associated with: - Myelodysplastic
syndrome (MDS) - Myeloid neoplasms - Poor prognosis - Risk of
progression to acute myeloid leukemia (AML)

\begin{enumerate}
\def\labelenumi{\arabic{enumi}.}
\setcounter{enumi}{2}
\item
  PNH Clone:

  \begin{itemize}
  \item
    Granulocytes: 68\% FLAER\textsuperscript{-}/CD24\textsuperscript{-}
  \item
    Monocytes: 72\% FLAER\textsuperscript{-}/CD14\textsuperscript{-}
  \item
    RBCs: 46\% CD59-deficient (31\% type III, 15\% type II)
  \item
    This is a large PNH clone, confirming PNH diagnosis
  \end{itemize}
\item
  HLH Evaluation (from previous request):

  \begin{itemize}
  \item
    Hepatitis B/C negative
  \item
    HIV negative
  \item
    CMV/EBV: past infection, no active replication
  \item
    Blood cultures negative
  \item
    Sputum cultures negative
  \item
    Viral PCR panel negative
  \item
    Wound swab grew KPC K. pneumoniae and VRE E. faecium (colonization,
    not necessarily infection)
  \item
    PET-CT: FDG-avid pulmonary nodules (known fungal sequelae), no other
    involvement
  \item
    cMRI: no intracranial fungal infection
  \end{itemize}
\end{enumerate}

Diagnostic Interpretation:

The combination of PNH clone with aplastic marrow AND monosomy 7
represents a HIGH-RISK scenario. This is consistent with: - PNH with
secondary myelodysplastic syndrome (MDS) - Or aplastic anemia with
clonal evolution (monosomy 7) - The monosomy 7 is an adverse prognostic
factor

The markedly elevated ferritin (4950 ng/mL) could represent: - Iron
overload from chronic transfusions - Inflammation/infection - Possible
early HLH (but other markers not yet available)

Clinical Significance:

This patient has complex hematologic disease combining: 1. PNH with
large clone 2. Aplastic bone marrow failure 3. Clonal evolution with
monosomy 7 (MDS-associated cytogenetic abnormality) 4. Recent
disseminated fusariosis with residual pulmonary involvement

Risk Assessment:

The presence of monosomy 7 significantly increases risk for: -
Progression to AML - Treatment resistance - Poorer overall survival

Monosomy 7 is considered a high-risk feature in MDS, warranting ag
{[}\ldots{]}

\textbf{Case 15} \textbar{} Multiple Myeloma \textbar{} Round 3
\textbar{} GLM 5

\textbf{Evaluator:} Strong final synthesis. Correctly identified the
dual diagnosis: relapsed/refractory MM (low burden, post-CAR-T) and
therapy-related MDS with biallelic TP53 mutations. Appropriately
recognized that the extensive treatment history (alkylators, IMiDs, two
autoSCTs, CAR-T) provides the mutagenic exposure context for t-MDS. TP53
mutations were correctly identified as conferring very high risk and
poor response to hypomethylating agents. Appropriately concluded that
allogeneic transplant is likely not feasible given age (75),
comorbidities (cardiomyopathy LVEF 46\%, CKD), and HCT-CI score. EBV
management with rituximab was correctly identified as a priority. The
treatment plan was realistic and patient-centered, acknowledging limited
options while proposing supportive care and clinical trial
consideration.

Thinking trace (step 15):

{[}\ldots{]} The ``correct'' differential was myeloma progression - but
the myeloma seems well-controlled with only trace M-protein and minimal
plasma cells. However, the EBV-associated LPD differential was only
``partially\_correct'' - suggesting the EBV finding is significant but
not the primary diagnosis.

Let me reconsider: What could the working diagnosis be that
incorporates: 1. t-MDS with excess blasts (5\% blasts, dysplasia) 2. EBV
reactivation in bone marrow 3. Possible myeloma progression (the
``correct'' differential)

The combination might be: - Therapy-related MDS (t-MDS) with EBV
reactivation

The EBV reactivation in the bone marrow could represent an
EBV-associated lymphoproliferative disorder, particularly given the
immunosuppressed state from antiviral prophylaxis (acyclovir,
posaconazole, cotrimoxazole). With CD4 count at 320/µL, there's
immunocompromise, and the EBV viral load in bone marrow significantly
exceeds peripheral blood levels, suggesting active marrow involvement.

Looking at the bone marrow results more carefully, the T-cell
compartment appears polypotypic by flow cytometry without abnormal
lymphoid populations. The elevated EBV levels may reflect reactivation
from immunosuppression rather than a separate LPD. This points toward
therapy-related myelodysplastic syndrome with excess blasts as the
primary driver of pancytopenia, complicated by EBV reactivation in an
immunosuppressed state, though I should also consider an overlap
syndrome involving both plasma cell and myeloid malignancies.

The evaluation feedback indicates myeloma progression should be in the
differential, yet the bone marrow shows only 5\% plasma cells with trace
M-protein and a normal light chain ratio. This seems controlled rather
than progressive. {[}\ldots{]}

\textbf{Case 11} \textbar{} DLBCL \textbar{} Round 2 \textbar{} MiniMax
M2.1

\textbf{Evaluator:} Outstanding R2 performance. The model correctly
identifies DLBCL from the biopsy immunophenotype (PAX5+, CD20 partial,
CD79a+, BCL6+, MUM1+, BCL2+, CD30+ 40\%), then recognizes the critical
importance of FISH testing for MYC/BCL2/BCL6 rearrangements. Upon
receiving FISH results showing IGH::MYC and IGH::BCL2 fusions, correctly
diagnoses double-hit high-grade B-cell lymphoma. The model demonstrates
excellent knowledge of treatment implications - recognizing that
standard R-CHOP is inadequate for double-hit lymphoma and recommending
DA-EPOCH-R. Integrates staging data (PET-CT SUVmax 18.4, bone marrow
negative, CSF negative) to provide comprehensive staging. The reasoning
about CD30 positivity causing potential confusion with Hodgkin lymphoma
shows sophisticated pathological knowledge.

Thinking trace (step 14):

{[}\ldots{]} \textbf{Biopsy:} - Diffuse effacement by polymorphous
infiltrate of atypical large lymphoid cells - B-cell lineage (CD45+,
CD79a+, PAX5+) - Partial CD20 expression - BCL6+, MUM1+ (non-germinal
center phenotype) - BCL2 strong+, MYC rearrangement positive (IGH::MYC)
- BCL2 rearrangement positive (IGH::BCL2) - BCL6 rearrangement negative
- CD30 moderate positivity (\textasciitilde40\%) - CD15 focal partial
expression - Ki-67 60-70\% - EBER negative

Bone Marrow:

No involvement by lymphoma

Mildly hypocellular but no clonal population

\textbf{Diagnosis:}

The combination of MYC AND BCL2 rearrangements (double-hit lymphoma)
with morphology showing large atypical lymphoid cells, high Ki-67, and
aggressive clinical presentation is diagnostic of \textbf{Double-Hit
Lymphoma (DHL)} - a high-grade B-cell lymphoma with MYC and BCL2
rearrangements.

I recognize this is an extremely aggressive malignancy with critical
clinical implications. The case meets specific diagnostic criteria:
high-grade B-cell lymphoma featuring MYC and BCL2 rearrangements, with
large cell morphology and complex genetic abnormalities. The patient's
presentation involves bulky cervical and mediastinal disease,
significant B symptoms, and airway compromise - all indicating an
urgent, high-risk clinical scenario requiring immediate, intensive
intervention.

The diagnostic criteria are precisely met through multiple confirmatory
features. Immunohistochemical markers strongly support the diagnosis,
including CD20 partial expression, BCL2 and BCL6 positivity, and
elevated Ki-67 proliferation index. Critically, fluorescence in situ
hybridization has definitively identified both MYC and BCL2
rearrangements, with EBV negativity ruling out alternative diagnoses
like Burkitt or Hodgkin lymphoma subtypes.

I'm confident this represents a high-grade B-cell lymphoma with MYC an
{[}\ldots{]}

\textbf{Case 3} \textbar{} AML \textbar{} Round 3 \textbar{} Kimi K2
Thinking

\textbf{Evaluator:} Excellent R3 reasoning. The model provides
comprehensive molecular interpretation: DDX41 at 43\% VAF correctly
identified as likely germline predisposition, ASXL1 and RUNX1 recognized
as adverse prognostic markers consistent with AML-MRC. Treatment plan is
well-calibrated for a 71-year-old: hypomethylating agent plus venetoclax
rather than intensive induction, with appropriate consideration of
reduced-intensity transplant if response achieved. The model correctly
identifies the need for germline DDX41 testing and family counseling.
Risk stratification aligns with the reference (adverse risk). The only
minor gap is not explicitly stating the complex karyotype referenced in
the R3 diagnosis, as cytogenetics results were not separately requested.

Thinking trace (step 13):

{[}\ldots{]} with hypomethylating agents - Supportive care only: For
very frail patients - Clinical trials - Allogeneic transplant: Rarely
considered in this age group, but possible with reduced intensity
conditioning in selected fit patients

Prognostic factors: - Age \textgreater60: Adverse - WBC \textless2.0:
Favorable in some models, but here it's due to marrow failure -
Cytogenetics: Not provided yet, but molecular mutations are present -
ASXL1 mutation: Adverse prognostic factor in AML - RUNX1 mutation:
Adverse prognostic factor in AML - TET2 mutation: Can be favorable or
neutral in some contexts, but often adverse - DDX41: Germline
predisposition, may respond to certain therapies - PTPN11: Associated
with adverse outcomes

Given the information, I should request: 1. Cytogenetics report
(karyotype) 2. Specific details of prior prostate cancer treatment (to
confirm therapy-related classification)

But the user said ``Final round - provide definitive diagnosis and
treatment plan''. This means I should solve now rather than requesting
more info, or make my best assessment.

Let me think about what I would need for a definitive plan: -
Cytogenetics is critical for risk stratification - Prostate cancer
treatment details are important for confirming therapy-related vs.~de
novo - ECOG/performance status - Cardiac function assessment -
Comorbidity index (e.g., HCT-CI)

However, since this is the final round and I need to provide a
definitive plan, I should synthesize what I have:

Diagnosis: Acute Myeloid Leukemia with Myelodysplasia-Related Changes
(AML-MRC) based on: - 25-30\% blasts in hypocellular marrow -
Multilineage dysplasia - History of prior cytotoxic therapy for prostate
cancer likely (Gleason 7b) - Molecular profile consistent with MDS/AML
progression (ASXL1, RUNX1, TET2)

The DDX41 mutation at 43\% VAF is interesting - this is very high,
suggesting it could be a germline DDX41 mutation, which predisposes to
myeloid malignancies. {[}\ldots{]}

\textbf{Case 18} \textbar{} Aplastic Anemia \textbar{} Round 3
\textbar{} Kimi K2.5 Thinking

\textbf{Evaluator:} Comprehensive final synthesis integrating TET2 (VAF
21\%) and DNMT3A (VAF 12\%) mutations with the established monosomy 7
and PNH clone. Correctly identified these as supporting clonal evolution
rather than simple CHIP given the clinical context. Final diagnosis of
MDS with monosomy 7 evolving from vSAA/PNH was accurate. Treatment plan
appropriately addressed the critical tension between urgent HSCT need
and active disseminated fusariosis, recommending continued triplet
antifungal therapy while planning conditioning. Correctly noted the need
to discontinue eltrombopag and exercise caution with G-CSF in MDS.
Patient-centered considerations about hospitalization anxiety and
childcare responsibilities were incorporated.

Thinking trace (step 11):

{[}\ldots{]} or PNH, eltrombopag, ciclosporin A, voriconazole,
terbinafine

\textbf{Bone marrow findings:} - Hypocellular marrow (\textless10\%
cellularity) with trilineage dysplasia - Ring sideroblasts \textless5\%
- Cytogenetics: monosomy 7 in 18/20 cells (90\%) - Blasts \textless5\% -
Flow: no blast excess, minimal residual myeloid precursors

\textbf{Molecular findings:} - TET2 mutation (VAF 21\%) - DNMT3A
mutation (VAF 12\%) - No TP53, RUNX1, ASXL1, etc.

\textbf{Working diagnosis evolution:} - Started with long-standing
PNH/aplastic anemia - Now evolved to high-risk MDS with monosomy 7 - The
presence of monosomy 7 alone defines high-risk MDS - TET2 and DNMT3A
mutations are common CHIP mutations but in this context with cytopenias
and dysplasia, they support clonal evolution

The key issue now is balancing the need for urgent allogeneic HSCT
(curative for high-risk MDS) against the active disseminated fungal
infection and patient reluctance for prolonged hospitalization.

\textbf{Final plan considerations:} 1. Diagnosis: Myelodysplastic
syndrome with isolated monosomy 7 (MDS-7), high-risk, evolved from
aplastic anemia/PNH

\begin{enumerate}
\def\labelenumi{\arabic{enumi}.}
\setcounter{enumi}{1}
\tightlist
\item
  Antifungal strategy:
\end{enumerate}

\begin{itemize}
\item
  Continue voriconazole (primary for fusariosis, though some species
  resistant)
\item
  Continue terbinafine (synergistic activity)
\item
  Need ophthalmology evaluation for possible endophthalmitis
\item
  May need adjunctive therapy or consideration of liposomal amphotericin
  B if progression
\end{itemize}

\begin{enumerate}
\def\labelenumi{\arabic{enumi}.}
\setcounter{enumi}{2}
\tightlist
\item
  Transplant planning:
\end{enumerate}

\begin{itemize}
\item
  Urgent allogeneic HSCT is indicated (high-risk MDS with monosomy 7)
\item
  Must balance against active fungal infection (relative
  contraindication)
\item
  Bridge therapy decisions:

  \begin{itemize}
  \item
    Continue pegcetacoplan (controls hemolysis)
  \item
    Continue ciclosporin A (IST may not be effective now with MDS, but
    could be continued as bridge)
  \item
    \textbf{Discontinue eltrombopag} - important in MDS with monosomy 7
    due to risk of blastic transformation
  \end{itemize}
\item
  Hypomethylating agents (azacitidine/decitabine) not indicated in
  high-risk MDS with monosomy 7 and severe cytopenias - transplant is
  preferred
\item
  Intensive chemoth {[}\ldots{]}
\end{itemize}

\textbf{Case 9} \textbar{} Multiple Myeloma \textbar{} Round 3
\textbar{} Claude Sonnet 4.6

\textbf{Evaluator:} Outstanding R3 reasoning with remarkable depth and
integration. After reviewing treatment history (5 lines including ASCT
x2, CAR-T, multiple PI/IMiD/anti-CD38 combinations), correctly
identifies penta-refractory status. Critically recognizes the TP53
mutation in the CD138-negative fraction as evidence of concurrent
therapy-related MDS, creating a dual-pathology picture. Treatment
strategy is well-calibrated: bispecific antibodies (talquetamab
targeting GPRC5D) for the refractory myeloma, palliative radiotherapy
for the extramedullary plasmacytoma, and careful consideration of the
competing MDS diagnosis. The model also revisits the AL amyloidosis
possibility and appropriately integrates palliative care goals.

Thinking trace (step 11):

{[}\ldots{]} The salvage therapy selection is where I went wrong
before---Dara-Vd isn't suitable given prior isatuximab exposure and
PI-refractoriness, so bispecific antibodies targeting GPRC5D or BCMA are
the better options here.

Radiotherapy to the extramedullary plasmacytoma should still be part of
the plan, and I need to evaluate the t-MDS with TP53 mutation more
carefully given the high VAF and cytopenias in this 81-year-old with
reduced ejection fraction. For penta-refractory disease, I'm weighing
bispecific antibodies, selinexor-based regimens, and clinical trial
eligibility against the competing concern of managing his
myelodysplastic transformation. context of the myeloid compartment, so I
need to differentiate whether the cytopenias are primarily from myeloma
treatment toxicity or from the emerging t-MDS. The key diagnostic
priorities are getting tissue confirmation of that shoulder mass and
characterizing the myeloid clone more thoroughly with targeted
sequencing. For immediate management, I'm focusing on the hypercalcemia
correction and then deciding between continuing myeloma-directed therapy
versus pivoting to address the t-MDS, which may actually be driving the
clinical deterioration. Looking back at what scored as correct versus
incorrect, I'm noticing a pattern---the core needle biopsy was the only
item that hit all three criteria, while most of my other recommendations
missed on either the specific intervention choice or priority level. The
denosumab might have been a dosing issue, the salvage therapy regimen
was wrong given his prior exposures, and several of the supportive
measures I flagged apparently weren't indicated or were lower priority
than I thought. I need to reconsider which interventions actually
address his most pressing clinical needs versus what seemed reasonable
in isolation.

The treatment history is the key---he's already been through proteas
{[}\ldots{]}

\textbf{Case 1} \textbar{} AML \textbar{} Round 3 \textbar{} GLM 4.7

\textbf{Evaluator:} Outstanding R3 synthesis. FLT3-ITD with high allelic
ratio (1.32) and NPM1 wild-type correctly identified as adverse risk.
Appropriate treatment plan with 7+3 plus midostaurin (RATIFY regimen),
allo-HSCT planning, and anticoagulation management. The model correctly
noted the patient's young age (35) as favorable for transplant
eligibility and recognized CD33 negativity precluding gemtuzumab.
However, the model did not identify this as relapsed AML (the title
includes `relapsed'), though treatment history was not explicitly
available until R3.

Thinking trace (step 14):

{[}\ldots{]} Or maybe the PE improved with hydration/leukoreduction.
\textbf{Transfusion Support:} Continuing during induction.
\textbf{Allogeneic Stem Cell Transplant (Allo-SCT):} Given high-risk
FLT3-ITD (high AR) + NPM1 WT, consolidation usually involves SCT if a
donor is available, unless MRD negative and favorable response, but FLT3
high AR pushes towards transplant.

1. Refining the Diagnosis:

\begin{itemize}
\item
  Diagnosis: Acute Myeloid Leukemia (AML) with mutated FLT3.
\item
  Subtype: AML with myelomonocytic differentiation.
\item
  Risk Stratification: Adverse/Intermediate-high (FLT3-ITD high AR, NPM1
  WT).
\end{itemize}

2. Refining the Treatment Plan:

\begin{itemize}
\item
  \textbf{Induction:} Cytarabine (7 days) + Daunorubicin (3 days) +
  \textbf{Midostaurin} (targeted FLT3 inhibitor). (This is the
  ``RATIFY'' regimen).
\item
  \textbf{Supportive Care:} Tumor lysis prophylaxis (continue/maintain),
  anti-emetics, blood product support.
\item
  \textbf{Infectious Prophylaxis:} Antifungal
  (Posaconazole/Voriconazole), antibacterial
  (Levofloxacin/Fluoroquinolone if neutropenic though colonized with KPC
  - need to be careful, maybe just monitor?), antiviral (Acyclovir).
\item
  \textbf{PE Management:} If not anticoagulated previously, reassess for
  therapeutic anticoagulation now (assuming counts recover or are
  supported). If still high bleed risk, continue monitoring. (The
  previous ``incorrect'' judgement suggests I shouldn't just jump to IVC
  filter).
\item
  \textbf{Role of Quizartinib/Gilteritinib:} Midostaurin is the approved
  frontline agent alongside 7+3.
\end{itemize}

3. Final Check of the ``Solve'' Request:

\begin{itemize}
\item
  ``Final round - provide definitive diagnosis and treatment plan.''
\item
  I have the FLT3 status. This completes the diagnostic picture.
\item
  I do not have the response to Round 1/2 treatments (e.g., did the WBC
  go down? did the PE resolve?).
\item
  However, usually in these cases, the ``Day \textgreater7'' setting
  implies stabilization. {[}\ldots{]}
\end{itemize}

\end{document}